\documentclass[twocolumn]{svjour3}         
\smartqed  
\usepackage{times}
\usepackage{epsfig}
\usepackage{graphicx}
\usepackage{amsmath}
\usepackage{amssymb}
\usepackage{color}
\usepackage{threeparttable}
\usepackage{multirow}
\usepackage{booktabs}
\usepackage{comment}
\usepackage{pbox}
\usepackage{bm}
\usepackage{multirow}
\usepackage{multicol}
\usepackage{adjustbox}
\usepackage{algorithm}
\usepackage{algpseudocode}
\usepackage[]{natbib}
\usepackage[pagebackref=true,breaklinks=true,letterpaper=true,colorlinks,bookmarks=false]{hyperref}
\usepackage{subcaption}
\usepackage{enumitem}
\usepackage{xcolor}
\usepackage{siunitx}

\newcommand{\etal}{\textit{et al}.}
\newcommand{\ie}{\textit{i}.\textit{e}.~}
\newcommand{\eg}{\textit{e}.\textit{g}.}

\begin{document}

\title{Learning from Exemplars for Interactive Image Segmentation}

\titlerunning{Learning from Exemplars for Interactive Image Segmentation}        

\author{Kun Li{$^{1}$}       \and
        Hao Cheng{$^1$} \and
        George Vosselman{$^1$} \and
        Michael Ying Yang{$^2$}
}


 \authorrunning{Li \etal} 

\institute{Kun Li \at
              k.li@utwente.nl          
           \and
           Hao Cheng \at
              h.cheng-2@utwente.nl
           \and
           George Vosselman \at
              george.vosselman@utwente.nl
           \and
           Michael Ying Yang \at
              myy35@bath.ac.uk
            \and
        	$^1$ \quad Faculty of ITC, University of Twente, The Netherlands \at
                $^2$ \quad Visual Computing Group, University of Bath, UK \at
}

\date{Received: date / Accepted: date}

\maketitle
\begin{abstract}
\sloppy
Interactive image segmentation enables users to interact minimally with a machine, facilitating the gradual refinement of the segmentation mask for a target of interest. Previous studies have demonstrated impressive performance in extracting a single target mask through interactive segmentation. However, the information cues of previously interacted objects have been overlooked in the existing methods, which can be further explored to speed up interactive segmentation for multiple targets in the same category. To this end, we introduce novel interactive segmentation frameworks for both a single object and multiple objects in the same category. Specifically, our model leverages transformer backbones to extract interaction-focused visual features from the image and the interactions to obtain a satisfactory mask of a target as an \textit{exemplar}. For multiple objects, we propose an \textit{exemplar}-informed module to enhance the learning of similarities among the objects of the target category. To combine attended features from different modules, we incorporate cross-attention blocks followed by a feature fusion module. Experiments conducted on mainstream benchmarks demonstrate that our models achieve superior performance compared to previous methods. Particularly, our model reduces users' labor by around 15\%, \ie, requiring two fewer clicks to achieve target IoUs 85\% and 90\%. The results highlight our models' potential as a flexible and practical annotation tool.
The source code will be released after publication.

\end{abstract}

\sloppy
\section{Introduction}
\label{sec:intro}
Interactive image Segmentation (IS) is a challenging task in Computer Vision that requires the prediction of class-agnostic object segmentation masks. It differs from instance segmentation \citep{he2017maskrcnn} by incorporating interactions between a machine and users to predict a pixel-wise target mask. This progressive mechanism enables the system to serve as a reliable annotation tool for other dense visual tasks, \eg, semantic segmentation \citep{long2015fcn}, image editing \citep{ling2021imageediting}, and panoptic segmentation \citep{kirillov2019panoptic}, and downstream applications, \eg, land cover classification \citep{li2020landcover} and autonomous driving \citep{zablocki2022autonomous}.

\begin{figure}[t]
  \begin{subfigure}[t]{1\linewidth}
  \centering
  \includegraphics[width=1\linewidth]{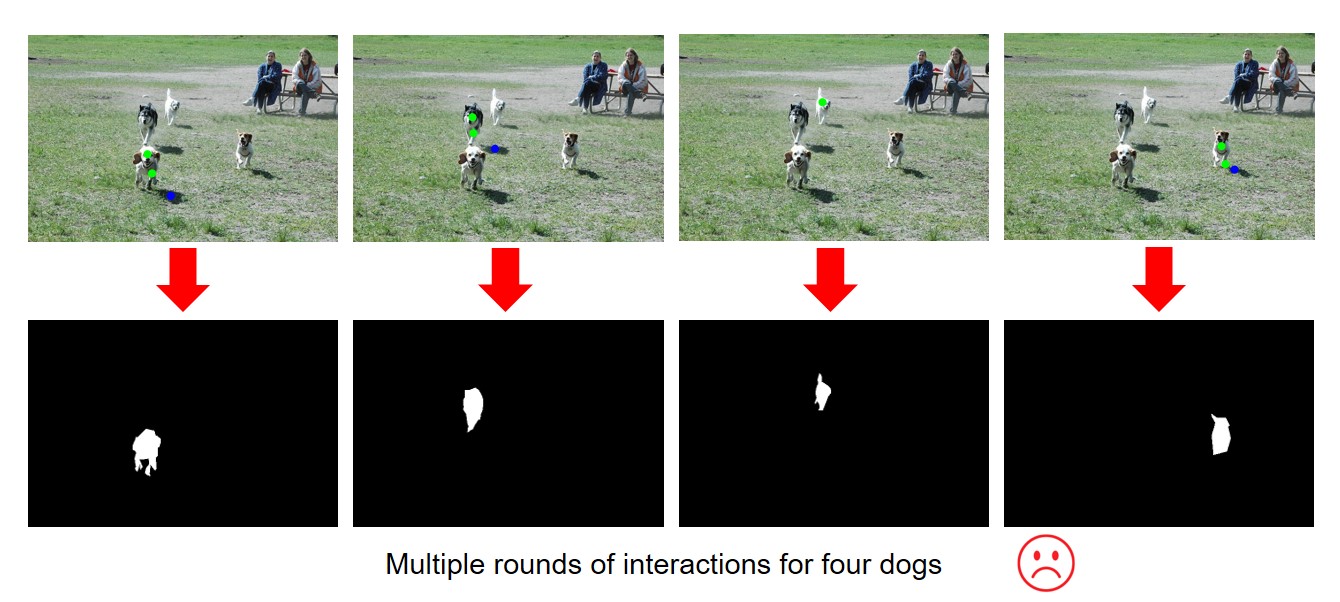}
  \subcaption{}
  \label{fig:motivation-a}
  \end{subfigure}
  \vspace{3mm}
  
  \begin{subfigure}[t]{1\linewidth}
  \centering
  \includegraphics[width=0.62\linewidth]{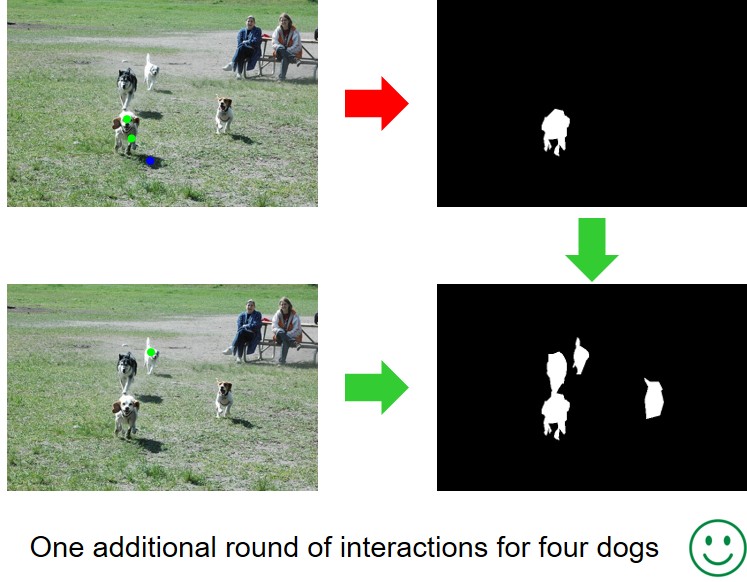}
  \subcaption{}
  \label{fig:motivation-b}
  \end{subfigure}
   \caption{(a) Current interactive segmentation methods (red arrows) designed for a single object require several rounds of interactions. In each round, users need to provide a set of positive or negative clicks to indicate the target. (b) Our interactive segmentation method (green arrows) predicts the mask of multiple objects in the same category by leveraging the previously interacted object.}
   \label{fig:motivation}
   \vspace{-4mm}
\end{figure}

In the last decade, we have witnessed the rising and fast-paced developments of deep neural networks for IS. These models support various interaction formats for users' convenience, including \textit{clicks}, \textit{scribbles}, and \textit{bounding boxes}.
In this paper, we mainly focus on click-based methods that are relatively convenient for users. In a standard setting, the first click (always a positive one) should be put into the center of a target while every other new click is placed in the regions where the model has made incorrect predictions. Whether a new click is positive or negative is decided by users based on the analysis of the current segmentation result. Hence, IS is a progressive refinement method based on a set of sequential positive and negative clicks. 

Numerous works try to improve the segmentation accuracy from different perspectives, for instance, interaction simulations \citep{mahadevan2018itis}, encoding of interactions \citep{majumder2019contentsp}, network architectures \citep{li2023icmformer, lin2024adaptiveclick}, and local-global refinement \citep{chen2022focalclick, lin2022focuscut}. However, these models typically focus on one object at a time.
In practice, we frequently encounter scenarios where batches of objects belonging to specific categories necessitate annotations, such as various applications of identifying same-category objects within one image, including crowd detection, image matting, scene understanding, and medical image analysis.
The prior works require repetitive interactions for multiple objects, which is time-consuming and labor-redundant. Taking Fig.~\ref{fig:motivation-a} as an example, previous methods treat each dog as an individual inference target. To segment all the dogs present in the image, users are required to engage in multiple rounds of interactions, refining the segmentation of each dog iteratively. Each round of interactions produces an independent segmentation result of a dog. To make a distinction, we refer to the previous interactive segmentation for a single target as Single Object Interactive Segmentation (SOIS) and the task involving multiple objects in the same category as Multiple Object Interactive Segmentation (MOIS).
As reported in Table~\ref{tab: motivation}, the quantitative results of interaction efforts using current IS methods on two datasets, COCO MVal \citep{lin2014coco} for SOIS and COCO MOIS introduced in Sec.\ref{sec:mois} for MOIS, reveals a significant disparity in performance between these two scenarios.
This performance degradation due to redundant interactions heavily restricts the practical applicability of these current methods. Therefore, there is a need to explore methods that can leverage the previously annotated information to cut short the redundant interactions for MOIS.

\begin{table}
\caption{Current interactive segmentation models' performance comparison: SOIS (on the COCO MVal \citep{lin2014coco} dataset) vs. MOIS (on the COCO MOIS dataset). The models were trained on the COCO \citep{lin2014coco} + LVIS \citep{gupta2019lvis} datasets. Number of Clicks (NoC) denotes the average number of clicks to reach target IoUs (85\% and 90\%, respectively).}
\centering
\setlength\tabcolsep{3pt}
\begin{tabular}{p{2.15cm}ccccc} 
\toprule
\multirow{2}*{Method} & \multirow{2}*{Venue}& \multicolumn{2}{c}{COCO MVal}& \multicolumn{2}{c}{COCO MOIS} \\
& & NoC85 & NoC90 & NoC85 & NoC90 \\
\hline
RITM & ICIP22 & 2.16 & 2.97 & 12.44 & 16.13 \\
SimpleClick & ICCV23 & 2.18 & 2.93 & 12.51 & 16.16 \\
AdaptiveClick & TNNLS24 & 2.07 & 2.86 & 12.44 & 16.10 \\
\bottomrule
\end{tabular}
\vspace{-2mm}
\label{tab: motivation}
\end{table}

How to effectively learn from the previously interacted objects to facilitate the segmentation of remaining objects in the same category has become a key issue in MOIS.
Some studies could be used to coarsely localize other objects belonging to the same category as the previously annotated target, such as object counting methods \citep{onoro2016objectcounting, liu2022countr}. These methods predict a density map of a certain category based on a template object, which is used to estimate the quantity of specific objects present in an image. For instance, \citet{doubinsky2024semanticcount} proposed a semantic augmentation method to enhance the diversity of synthesized images for templates and improve the quality of density maps. The outputs can be further processed to obtain a simplified prompt (\eg, points or bounding boxes) for pixel-wise predictions. However, these methods cannot guarantee that the density maps accurately or completely identify the locations of similar objects, as they do not incorporate the crucial information from previous interactions, which is essential for MOIS.
Alternatively, One-Shot/Few-Shot image segmentation methods \citep{shaban2017oneshot, ganea2021incrementaloneshot} could regard a well-annotated object from the previous interactions as a support object/mask pair for query objects within the same image.
Normally, these methods utilize a two-branch structure (\ie, a support branch and a query branch) to transfer annotation information and facilitate connections between extracted features. For instance, \citet{zhang2021fsss} integrated pixel-level support features into the query with transformers to achieve feature alignments between the two branches, and directly predicted segmentation results.
However, sharing an image of the support and query branches with multiple mask choices will inevitably confuse models in the alignments, leading to unstable model training. In addition to this limitation, the absence of users' interactions also restricts the performance.

Another key issue in transferring SOIS to MOIS is effectively fusing multiple features from different branches representing the previously annotated object and clicks, and remaining similar objects. A naive way of fusion is the channel concatenation process. For instance, FCANet \citep{lin2020fca} concatenated the features from the proposed first click attention module and the main branch before an Atrous Spatial Pyramid Pooling (ASPP) module. 
However, this naive fusion cannot effectively propagate the feedback from the first click to the remaining objects.
Even though our previously proposed iCMFormer model \citep{li2023icmformer} employs a different group of self-attention blocks on the cross-attended features, these self-attentions can lead to increased feature redundancy and computational consumption without any distinctions among the queried features. Also, the additional features from the previously obtained object make the fusion more complex.

In this paper, we propose an effective \textit{exemplar}-informed approach to smoothly transferring SOIS to MOIS.
To the best of our knowledge, we are the first to explore the connection between a previously interacted object and the multiple remaining objects sharing the same category information (as shown in Fig.~\ref{fig:motivation-b}). For brevity, we refer to them as \textit{exemplar object} and \textit{recall objects}, respectively, which are detailed in Sec.~\ref{sec: formulation}. To mitigate the above issues of linking multiple features from different branches and avoid significantly increasing feature redundancy and computational consumption, we carefully design a deep neural network (termed iCMFormer++) that makes better use of the \textit{exemplar} for MOIS, largely distinguishing our approach from the aforementioned template- or map-based approaches \citep{onoro2016objectcounting, liu2022countr, doubinsky2024semanticcount}. Concretely, in the iCMFormer++ model, we first use a two-stream transformer architecture as the backbone to represent the \textit{exemplar object} with its interactions and the original image with additional interactions for \textit{recall objects} simultaneously. Furthermore, we employ an \textit{exemplar}-informed module as the objective guidance to measure the similarities between the \textit{exemplar object} and the potential \textit{recall objects} from the image. Subsequently, a channel embedding fusion module is carefully designed to process the cross-attended features from the representations of different modules before a segmentation head. We conduct comprehensive experiments on several mainstream datasets for IS, including an extended new IS dataset for multiple objects in the same category.
Our models show superior performance for both SOIS and MOIS.
Moreover, an extensive set of ablation studies offers detailed empirical analyses of the effectiveness of our method and its individual components.

This article is an extension of our previous work (iCMFormer) on click-based IS published in \citep{li2023icmformer}. In summary, the contributions of iCMFormer are:

\begin{itemize}
    \item A simple yet effective cross-modality interactive transformer utilizes the guidance information to generate robust segmentation results.
    \item The proposed cross-modality blocks are flexibly integrated into different transformer architectures, \eg, ViT \citep{dosovitskiy2020vit} and Swin \citep{liu2021swin}.
    \item The proposed iCMFormer model outperforms other IS methods on four single object inference benchmarks.
\end{itemize}

In this article, we present the additional \textbf{main contributions} as follows:
\begin{itemize}
    \item To the best of our knowledge, this is the first work to facilitate IS models to learn from the previously interacted object for multiple objects in the same category. We also extend a dataset to assess MOIS tasks, which will be released for future research.
    
    \item We propose a new model, iCMFormer++, named after the model proposed in \citep{li2023icmformer}. As opposed to the existing methods that simply treat MOIS as an interactive combination of SOIS with redundant interactions, iCMFormer++ is equipped with the \textit{exemplar-aware} transformers to better leverage previously interacted objects for interactively predicting the masks of multiple objects in the same category.
    
    \item The iCMFormer++ model achieves state-of-the-art performance on the HIM2K \citep{sun2022humanvmultiple} dataset and the newly extended dataset COCO MOIS, highlighting its potential for real-world applications.
\end{itemize}

\section{Related Work}
\label{sec:related_work}
In Computer Vision, a large body of work deals with image segmentation across diverse scenarios. Amongst them, interactive image segmentation (IS) aims to predict the pixel-wise mask of a target with the help of users' interactions. In this section, we review the most relevant works to our IS work proposed in this paper.

\subsection{Interactive Image Segmentation}
\vspace{2mm}
\noindent\textbf{Interaction optimization and learning.} Early methods focus on the representation of interactions and the establishment of optimizer with handcrafted features, such as Graph Cut \citep{boykov2001graphcut} and Level Set \citep{liu2011levelset}. DIOS \citep{xu2016dios} is the first work to address IS using a deep neural network. This work simulated the interactions based on several strategies and fixed them during the training stage. \citet{maninis2018extremecut} proposed four extreme points (the top-most, right-most, bottom-most, and left-most) to indicate the accurate boundary of the target. However, neither of these methods offers simple or universally applicable interactions with users for labeling objects in real-world scenarios. ITIS \citep{mahadevan2018itis} introduced an iterative training scheme for simulating and saving the interactions to speed up the process. RITM \citep{sofiiuk2022ritm} further improved this scheme by adopting a new sampling strategy with less computational resources. 
GPCIS \citep{zhou2023gpcis} introduced a Gaussian process classification model, offering a probabilistic framework for IS.
BRS (Backpropagating Refinement Scheme) was first introduced in \citep{jang2019brs} as an optimization procedure that reduces the disparity between the predicted mask and the input click map with each subsequent click. Then BRS was improved in \citep{sofiiuk2020fbrs} with a reduced runtime. However, these methods only consider global information while the local representations from the provided interactions are ignored. To address this issue, \citet{chen2022focalclick} and \citet{wei2023fcf} proposed to capture the local regions to obtain a zoom-in perspective for detailed extractions. 

\vspace{2mm}
\noindent\textbf{Transformer-based methods.} Recently, several works utilize ViTs (Vision Transformers) \citep{dosovitskiy2020vit}, well known for their excellent performance in object detection and segmentation, to address the challenges of IS. For example, SimpleClick \citep{liu2023simpleclick} is the first to introduce a plain-backbone method of ViTs for IS. 
Furthermore, \citet{li2023icmformer} introduced a transformer-based framework for a better fusion of different modalities.
AdaptiveClick \citep{lin2024adaptiveclick} employed a click-aware mask-adaptive transformer framework with an adaptive focal loss to address intra-class click ambiguity problems. Although these methods have achieved significant improvements in terms of segmentation performance, they only focus on a single target from an image for IS.
\subsection{Multiple Object Segmentation}
Generally, image segmentation of multiple objects aims to identify the categories of pixels belonging to different objects. In the interactive setting with users' indications, the objects that appear in one image can belong to a) various categories, in which the task is called full image IS; or b) the same category, in which the task is called the same category IS. We call the latter MOIS if it is not otherwise stated.

 \vspace{2mm}
\noindent\textbf{General multiple object segmentation.} This task differs from MOIS as no interactions from users are needed.
Based on feature representations, methods for this task can be divided into three groups: top-down, bottom-up, and direct methods. Top-down methods \citep{bolya2019yolact, he2017maskrcnn, ke2021deeptopdown} first detect objects in bounding boxes and then segment the pixels inside those bounding boxes. Bottom-up methods \citep{fidler2013bottomup, gao2019ssap} take a contrasting approach by first learning embeddings for each pixel and subsequently grouping similar ones into different objects. Alternatively, direct methods \citep{wang2020solov2, xie2021polarmask++} predict multiple object segmentation results in a box- and group-free way.
In this paper, we follow the direct methods, but additionally, take as input the users' interactions, \ie clicks, to achieve MOIS. 
There are only a few attempts to leverage category information for class-specific object segmentation without interactions.
For example, \citet{chen2015multiseg, he2014exemplar} utilized an energy minimization framework to incorporate category-information reasoning and shape prediction. 
\citet{sun2022humanvmultiple} proposed a mutual guidance strategy and a multi-instance refinement module to interchange information among targets for human instance matting. However, these methods necessitate high-quality reference masks to link the relations between a target object and other potential candidates within one image. Moreover, they are tailored for restricted categories (such as the human category). 
This restriction prevents these methods from being extended to class-agnostic MOIS.

 \vspace{2mm}
\noindent\textbf{Full image IS.} This task considers more than one object or stuff region for IS regardless of the category information. 
\citet{agustsson2019fullinteractive} proposed a two-stage Mask-RCNN-based model to make a full image segmentation prediction for all objects and region stuff with extreme points and scribbles. Furthermore, DynaMITe \citep{rana2023dynamite} improved it with only general point indications and leveraged shared image-level features conditioned on the related interactions for multiple objects. These two methods are based on a simple principle: the positive interactions for one instance are relatively negative for others, which is not totally true for objects in the same category, \eg, similar contextual features shared by the dogs as shown in Fig~\ref{fig:motivation}. We clearly differentiate this type of clicks for objects in the same category as ``Pseudo Negative Clicks" in Sec.~\ref{sec:training_scheme}. Moreover, they only focus on the spatial locations of the interactions but overlook the acquired object-level information for the remaining objects.

\vspace{2mm}
\noindent\textbf{Multiple object IS (MOIS).}  
In this paper, we focus on MOIS which involves objects sharing the same category information with a previously interacted object, whereas different objects across images are class-agnostic.
Surprisingly, most existing methods individually detect and segment objects, meaning that the shared similarities of the objects in the same category are under-explored.
The most relevant research to class-agnostic MOIS is guided image segmentation using \textit{support guidance} for certain categories.
These methods explore the intrinsic relationships between support images and query images in terms of appearance, texture, category, spatial distribution, and other aspects in data scarcity scenarios \citep{sung2018fsl1, sun2019metafsl}. For instance, CRNet \citep{Liu_2020_crosseference} proposed a cross-reference network to make predictions for both the support and the query images. BAM \citep{lang2022newfes} proposed two complementary learners (base and meta learners) to share the class information and an ensemble module to suppress the incorrect activation from coarse predictions. However, these methods rely on a set of provided support samples with high-quality annotations. In addition, preparing the support images is also costly for each query image. 
As opposed to the image-level support guidance, in this paper, we aim to explore the category information to reduce redundant interactions of MOIS within the same image.
To achieve this goal, we regard the collection of interactive clicks from the single object as the support guidance, \ie \textit{exemplar}, for multiple objects in the same category with the target, and design an effective way to leverage this information for segmenting the remaining objects.

\section{Method}
\label{sec:method}
We first formulate the problem of interactively extracting a single object and subsequently extend it to extracting multiple objects in the same category by learning from a previously interacted object. Regarding the proposed framework, we introduce the primary components of the iCMFormer++ model: IS structure, \textit{exemplar}-informed module, and feature fusion module. We further explain the training schemes of IS for both single and multiple objects in our case.

\begin{figure*}[ht!]
\centering
\includegraphics[width=1\linewidth]{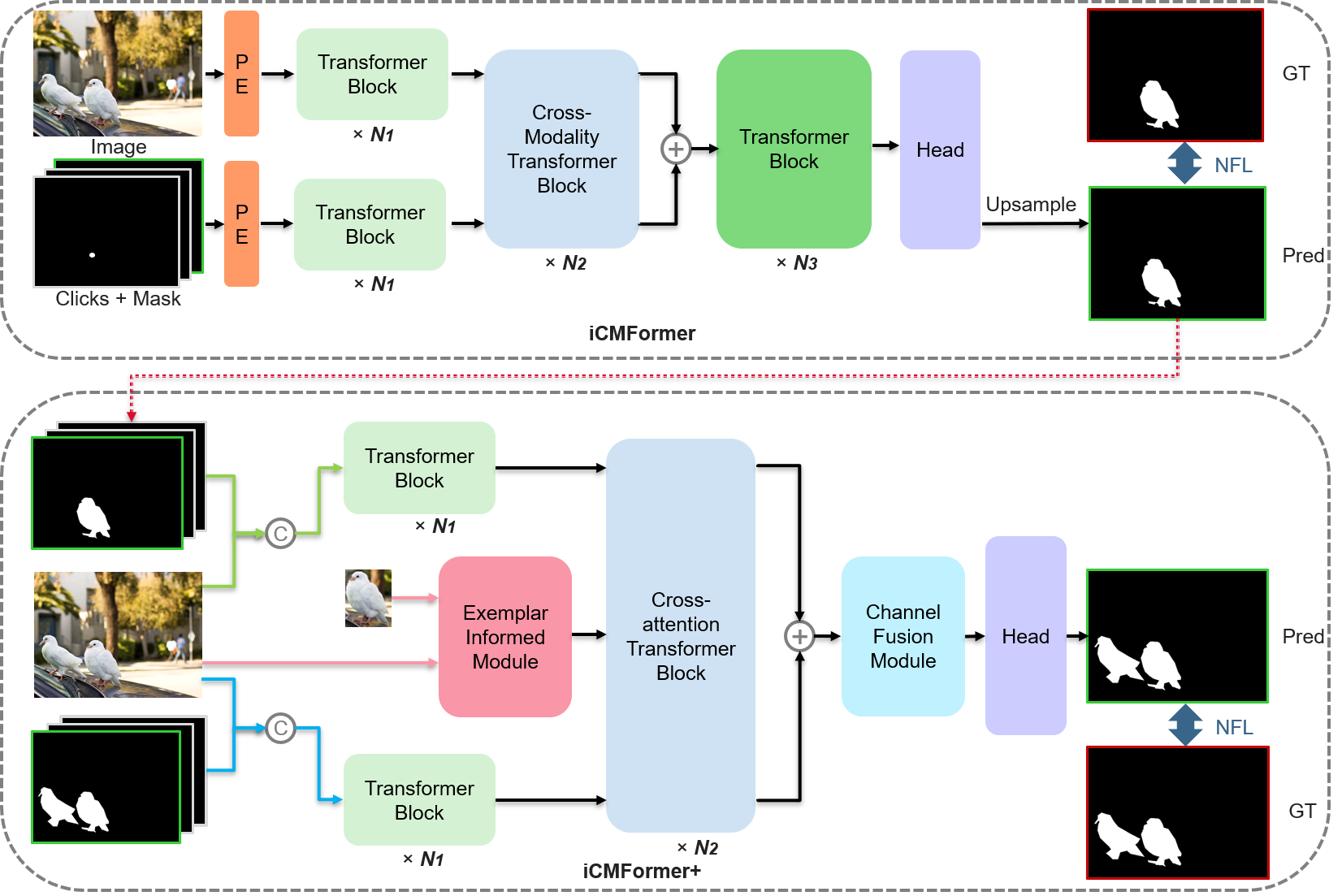}
   \vspace{-4mm}
\caption{Overall iCMFormer and iCMFormer++ frameworks. The prediction of an object by the iCMFormer serves as the \textit{exemplar} for our iCMFormer++ model, which is denoted in a red dotted line. The iCMFormer++ model is built on top of the iCMFormer model with several modifications. Specifically, We first adopt a two-stream transformer backbone to extract the features from the previously obtained \textit{exemplar} (denoted as \textit{exemplar} branch with green arrows) and the overall image with additional interactions (denoted as \textit{recall} branch with blue arrows). An \textit{exemplar}-informed module is employed to learn the similarity between the \textit{exemplar} and other potential image regions. Finally, a channel fusion module processes the cross-attended features before the final segmentation for all objects in the same category. For brevity, we remove the position embedding (PE) and the upsampling step in the illustration of iCMFormer++.}
   \label{fig:framework}
   \vspace{-4mm}
\end{figure*}

\subsection{Problem Formulation}
\label{sec: formulation}
Given an input image $I$, users need to progressively provide different interactions, including positive or negative sampled pixels, to indicate the foreground or background information until the IS system predicts a satisfactory dense mask $M$ of a target $O$.
Normally, the satisfactory mask refers to the segmentation result that reaches at least 85\% IoU compared to the ground truth.
In this paper, we focus on the click-based IS, in which the interactions are denoted as $C_p$ and $C_n$ for positive and negative clicks, respectively.

The conventional SOIS is formulated as follows:
\begin{equation}
M = f_\theta(I, \{C_p, C_n\}),
\label{eq:single object IS}
\end{equation}
where $f$ and $\theta$ represent the IS system and its parameters, respectively. Note that $C_n$ is optional because sometimes only one positive click $C_{p1}$ can already achieve a high-quality segmentation result without any background indication. However, in this formulation, SOIS treats each object independently, ignoring the connection of target objects, \eg, $O^i$ and $O^j$, in the same category during the process of different interaction rounds, consequently resulting in redundant clicks to obtain their segmentation masks. 

To improve the above formulation for MOIS, we explore a more intelligent way of leveraging the previously interacted objects. 
The key concept introduced here is to define the first target object $O^e$ obtained from an image using the same SOIS setting as \textit{exemplar target} (ET). 
It consists of the mask $M^e$ of the object $O^e$ and the corresponding clicks $\{C^e_p, C^e_n\}$, i.e., $\text{ET} = \{M^e, \{C^e_p, C^e_n\}\}$.
Consequently, we learn a function $f^{all}_\theta$ that propagates the ET to \textit{recall} objects $\{O^1, O^2,...O^d\}$ in the same category of $O^e$ from $I$. 
During the training and inference phases, the mask and clicks of the ET $\{\hat{M}^e, \{\hat{C}^e_p, \hat{C}^e_n\}\}$ do not change anymore, and the prediction of the \textit{recall} objects still follows the interactive fashion that progressively adds new clicks \{$C_p$, $C_n$\} to the input. 
This new paradigm is denoted as
\begin{equation}
M^{all}_O = f^{all}_\theta(I, \{C_p, C_n\}, \hat{M}^e, \{\hat{C}^e_p, \hat{C}^e_n\}),
\label{eq:multiple object IS}
\end{equation}
where $O^{all}=\{O^e, O^1, O^2,...O^d\}$. 
Compared to the conventional formulation, this learning paradigm is more practical and user-friendly because it requires less repetitive interaction labor for similar objects in one image.

\subsection{Interactive Segmentation Architecture}
\label{sec:IS structure}
The IS architecture includes extracting feature representations of different data (\ie, the input image and interactions), the fusion of these representations, and a prediction head.
As shown in Fig.~\ref{fig:framework}, we extend the SOIS network iCMFormer \citep{li2023icmformer} into a MOIS network iCMFormer++, using the predicted ET outputted by iCMFormer as the \textit{exemplar} to facilitate the prediction of \textit{recall} objects.

\vspace{2mm}
\noindent
\textbf{Revisit of SOIS network iCMFormer}.
Before delving into the detailed network design of iCMFormer++, we briefly revisit our iCMFormer model, including interaction encoding, feature representation, cross-modality attention and fusion, and segmentation head.

\vspace{1mm}
\noindent
\textit{Interaction Encoding.} 
We employ two disk maps with a fixed radius (five pixels) to represent the sparse positive and negative clicks, respectively, enabling the point-based clicks to be encoded as image features. 
This encoding scheme has been proven more efficient and effective than distance-based encodings, \eg, distance maps and Gaussian maps in RITM \citep{sofiiuk2022ritm}. 
Similar to the previous methods \citep{chen2021cdnet, sofiiuk2022ritm, chen2022focalclick}, we also combine the previous segmentation mask with click maps for better performance.

\vspace{1mm}
\noindent
\textit{Feature representation.}
To represent images and clicks, we employ the powerful plain vision transformer ViT \citep{dosovitskiy2020vit} as our backbone to extract the features from the input data. 
ViT is a classical self-attention network, which splits the input image into smaller patches with positional embeddings. These patches are further flattened and projected into a linear space as a vector that serves as the input for transformers. The self-attention in transformers enables each patch in a sequence to attend to all the other patches, allowing for capturing contextual relationships within the sequence.
The iCMFormer follows ViT-B which employs 12 transformer blocks but only uses the first six blocks as a group (S1) for the feature representation.

\vspace{1mm}
\noindent
\textit{Cross-modality attention and fusion.} 
Multi-head attention (MHA) is a fundamental component in transformer blocks, which utilizes query $Q$, key $K$, and value $V$ to capture various aspects of attention. It calculates weighted attention through scaled dot products between queries and keys, resulting in a summation of the values.
In iCMFormer, a cross-modality block consists of two steps of MHA: 1) self-attentions on the $Q$, $K$, and $V$ from click features $Y$; 2) cross-attentions on the $Q$ from image features $X$ with $K$ and $V$ from $Y$.
We employ three cross-modality blocks followed by the last six blocks (S2) of ViT-B for the fusion.

\vspace{1mm}
\noindent
\textit{Segmentation head.} Given that the attended features across different inputs have been obtained, we employ a hierarchical segmentation head to capture multiple objects of various sizes.
Inspired by ViT-Det \citep{li2022vitdet}, we utilize simple MLP layers to process the plain features from ViT, which involves: unifying the channel dimensions of the final output from the S2 group; upsampling to achieve a consistent resolution; fusing features via concatenation; and generating the final segmentation prediction with Sigmoid activation.
After upsampling to the original resolution, a probability map for the target prediction is transformed to a binary segmentation result with a fixed threshold (normally 0.5).

\vspace{2mm}
\noindent
\textbf{Extension to MOIS network iCMFormer++.} 
Compared to iCMFormer, our new iCMFormer++ additionally takes as input the \textit{exemplar} outputted from iCMFormer to facilitate the segmentation of \textit{recall} objects.
Here we highlight the modifications adapted to the \textit{exemplar} input. 
Compared to iCMFormer, in the iCMFormer++ model, we keep the S1 group representing the concatenated image and clicks, but before the cross-attention, we employ two-stream branches to separately handle the learning of the \textit{exemplar} and \textit{recall} objects, incorporating provided clicks.
To distinguish between the two branches, we refer to them as the \textit{exemplar} branch and the \textit{recall} branch, as illustrated at the bottom of Fig.~\ref{fig:framework}.
Subsequently, we feed the self-attended features obtained from the \textit{exemplar} branch and the \textit{recall} branch as $E$ and $R$, respectively, to the cross-attention blocks.
One head of the cross-attention is given by:
\begin{equation}
f_\text{cross} = A(Q_R,K_E,V_E) = {\rm Softmax}(\frac{Q_RK_E^T}{\sqrt{d_E}})V_E,
\label{eq:cross-a}
\end{equation}
where $Q_R$ represents the queries from $R$ while $K_E$, $V_E$ and $d_E$ denote the keys, values, and dimensionality of keys from $E$. Then it follows a feed-forward network (FFN) with ReLU activation and Dropout like a standard transformer block.
It should be noted that, compared to the cross-attention in iCMFormer, the keys and values are derived from the \textit{exemplar} instead of the click features.

In the following, we introduce an \textit{exemplar}-informed module to learn the similarity between the \textit{exemplar} and \textit{recall} objects in Sec.~\ref{sec:case module} and replace the S2 group with a new feature fusion module introduced in Sec.~\ref{sec:ff module}.

\begin{figure}[t!]
\centering
\includegraphics[width=1\linewidth]{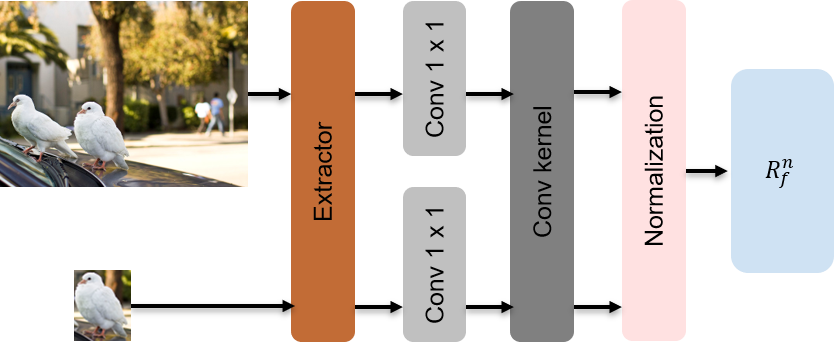}
\caption{Architecture of the proposed \textit{exemplar}-informed module (EIM). A pre-trained ResNet \citep{he2016resnet} is employed as the extractor for the visual representation. The \textit{Conv kernel} obtains the point-wise feature comparison by convolving the projected features $\text{Feat}^p_o$ with kernels from $\text{Feat}^p_e$. With a normalization layer, the EIM outputs the response activation vector denoted as $R^n_f$.}
\label{fig:case informed module}
\vspace{-2mm}
\end{figure}

\subsection{Exemplar-Informed Module}
\label{sec:case module}
For MOIS, the key insight is that within the class-agnostic framework of IS, effectively utilizing category information is crucial for accurately predicting multiple objects sharing the same category as the previously interacted object.
To solve this problem, we propose an \textit{exemplar}-informed module (EIM) by comparing an \textit{exemplar object} with the original image to obtain a similarity response, which highlights the focus in the cross-attention blocks.
Specifically, EIM leverages the prior category information from the \textit{exemplar} and helps analyze how the potential image regions are alike the \textit{exemplar object}.
The structure of EIM is shown in Fig.~\ref{fig:case informed module}.
It consists of three components: feature extraction, feature projection and comparison, and response activation.
Here, consider an \textit{exemplar object} $O^e$ and its original image $I$, which are taken as the input of EIM.
We first extract the visual features $\text{Feat}_o \in \mathbb{R}^{H_o \times W_o \times C_o}$ from the original image $I$ with a pre-trained ResNet \citep{he2016resnet} model, where $H_o$, $W_o$, and $C_o$ denote the height, width, and dimensionality of the features.
Provided with the satisfactory mask $M^e$ of the \textit{exemplar object} $O^e$, we obtain a cropped image region via a bounding box with buffers and subsequently extract the visual features for the \textit{exemplar} denoted as $\text{Feat}_e \in \mathbb{R}^{H_e \times W_e \times C_e}$.
Before directly comparing the features $\text{Feat}_e$ and $\text{Feat}_o$, we apply a projection step using $1 \times 1$ convolutional layers to extract valid information, thereby obtaining $\text{Feat}^p_e$ and $\text{Feat}^p_o$, respectively.
To this end, the features representing different visual contents are projected in a shared space. We calculate the feature comparison through a special convolution that leverages $\text{Feat}^p_e$ as kernels for $\text{Feat}^p_o$ to generate responses $R_f$ as follows:
\begin{equation}
R_f = {\rm Conv}({\rm Proj}(\text{Feat}_o), {\rm kernel} = {\rm Proj}(\text{Feat}_e)),
\label{eq:cim2}
\end{equation}
where ${\rm Conv}$ and ${\rm Proj}$ denote the convolution and projection steps above, respectively. We employ a normalization layer on $R_f$ to avoid the model being stuck in the same image region as the \textit{exemplar object}.
Finally, the response activation vector $R^n_f$ is obtained after the normalization with a ${\rm Softmax}$ function along the \textit{exemplar} dimensionality of $\text{Feat}_e$:
\begin{equation}
R^n_f = {\rm Softmax}(\frac{R_f}{\sqrt{H_eW_eC_e}}),
\label{eq:cim1}
\end{equation}
where $H_e$, $W_e$, and $C_e$ represent the shape of the \textit{exemplar} features $\text{Feat}_e$.
We leverage the response activation $R^n_f$ as a guidance signal in the cross-attention blocks.
Inspired by the learnable query proposed in \citep{goyal2021learnedquery}, we transform $R^n_f$ as a guided query merged with the query from the \textit{recall} branch to focus on similar patches.

\subsection{Feature Fusion Module}
\label{sec:ff module}
\begin{figure}[t!]
  \centering
  \includegraphics[width=1\linewidth]{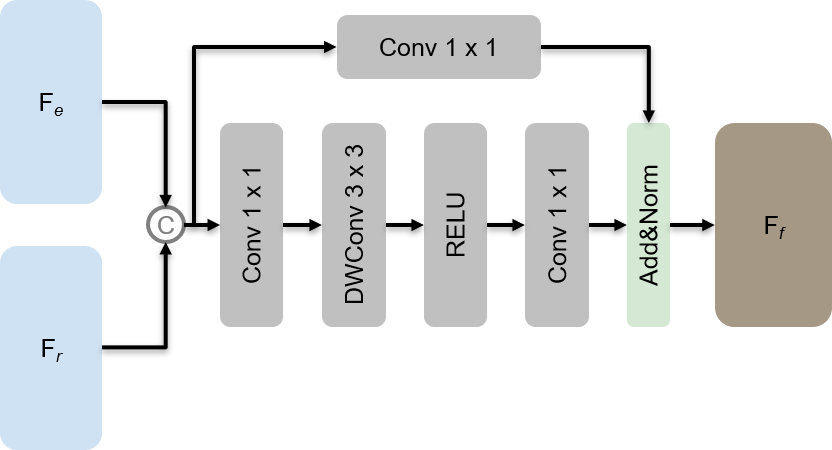}
  \caption{Illustration of the proposed channel embedding fusion module. We feed the module with cross-attended features from the \textit{exemplar} branch and the \textit{recall} branch, represented as $\text{F}_e$ and $\text{F}_r$. The \textit{DWConv} denotes a depth-wise convolution layer. The module outputs the fused feature $\text{F}_f$ before the final segmentation head.}
  \vspace{-4mm}
  \label{fig:ff}
\end{figure}

Considering the large receptive fields achieved through the self-attentions and sufficient information exchange enabled by the cross-attentions, the network necessitates feature fusion prior to reaching the segmentation head for the final prediction.
Recall the iCMFormer model, we incorporated several self-attention transformer blocks (specifically, the S2 group, as introduced in Sec.~\ref{sec:IS structure}) to further process and fuse the element-wise summed features from both the click and image branches.
If we directly adapt this fusion strategy to MOIS, the merged features from the \textit{exemplar} branch and the \textit{recall} branch will equally respond to the self-attentions of the S2 group, without maintaining the necessary distinctions.
However, the \textit{recall} refinement under the guidance of an \textit{exemplar} for the potential objects should be the focus for MOIS.
In addition, the computation complexity should also be considered for IS.
Therefore, for improved prediction quality and better \textit{exemplar} correspondence in the interactive \textit{recall} refinement, we employ an efficient channel embedding fusion module to integrate the attended features through the cross-attention blocks, as illustrated in Fig.~\ref{fig:ff}.
Specifically, we first combine the concatenated features from both the \textit{exemplar} branch and the \textit{recall} branch, followed by $1\times1$ convolutional layers to combine them. To exploit the surrounding regions, $3\times3$ depth-wise convolutional layers are adopted followed by another $1\times1$ convolutional layers. Inspired by the ResNet \citep{he2016resnet} model, we plus a skip-connected structure with $1\times1$ convolutional layers to finally obtain the fused features $\text{F}_f$. The channel fusion module minimizes computational complexity by incorporating fewer parameters (only 2.96M), accounting for only around 10\% of parameters compared to the previous self-attention blocks.

\subsection{Training Scheme}
\label{sec:training_scheme}
Different from standard image segmentation, IS requires interactions during the training stage.
However, collecting interactions for every sample is prohibitively expensive, and training a model with the substantial memory requirements imposed by these dynamic interactions is even more infeasible.
To this end, interaction simulations have become a prevalent solution for generating clicks without requiring direct user participation during model training.
For instance, RITM \citep{sofiiuk2022ritm} generates the next click in the cluster center of the largest incorrect prediction region after the morphological erosion operation and saves three iterative clicks during model training.
We employ a similar click simulation strategy as RITM and make a small change to the selection of the iterative click's position. Specifically, we combine the center point and a random point near the borders of the mislabeled regions to fit users' behaviors better.

Furthermore, we rethink the shifted role of the negative clicks $C^e_n$ of an \textit{exemplar} in MOIS.
For previous methods, SOIS has a clear distinction between positive and negative clicks, which indicate the single target and background, respectively.
When dealing with multiple objects based on a satisfactory \textit{exemplar}, the negative clicks have changed since the target shifts from a single object to multiple objects in the same category.
Specifically, in MOIS, negative clicks occurring within regions belonging to other objects of the same category as the \textit{exemplar object} should be considered positive clicks.
We term these negative clicks as \textit{Pseudo Negative Clicks}, indicating their role solely in reflecting previous interactions with the \textit{exemplar}.
To distinguish negative clicks specific to an \textit{exemplar} among multiple objects, we obtain \textit{True Negative Clicks} by excluding \textit{Pseudo Negative Clicks} from overall negative clicks.
To this end, these \textit{True Negative Clicks} of the \textit{exemplar} will not conflict with the negative clicks of the \textit{recall} objects in the same image.
It is worth mentioning that we have conducted an ablation in Sec.~\ref{sec: ab various click} to analyze the impact of various clicks.

\section{Experiments}
\label{sec:experiments}
\subsection{Datasets}
\label{sec:datasets}
To evaluate the effectiveness of the proposed IS models, we compared the performance with the prior methods on several existing datasets and our extended dataset named COCO MOIS. Here, we introduce the detailed properties of these datasets and the extension process of the COCO MOIS dataset dedicated to the MOIS task.

\subsubsection{Existing Datasets}
\label{sec:existing datasets}
Following recent works \citep{sofiiuk2022ritm, liu2023simpleclick}, the SOIS models were mainly trained on the SBD \citep{hariharan2011sbd} and COCO \citep{lin2014coco} + LVIS \citep{gupta2019lvis} datasets, and tested on various IS datasets, \ie, GrabCut \citep{rother2004grabcut}, Berkeley \citep{mcguinness2010berkeley}, and COCO Mval \citep{lin2014coco}.

\begin{itemize}
  \item \textbf{SBD.} The dataset is divided into two subsets for the object segmentation task (training: 8,498 images and 20,172 instances, validation: 2,857 images and 6,671 instances). Benefiting from its quantity and diversity, the SBD dataset serves as the main benchmark for the training and evaluation of the SOIS task.
    \item \textbf{COCO + LVIS.} Following RITM \citep{sofiiuk2022ritm}, we adopted the combined version of the COCO and LVIS datasets that contains 118K images with 1.2M instances with higher annotation quality for large-scale training of the SOIS models.
    \item \textbf{GrabCut.} The dataset contains 50 images and provides one single instance mask for each image.
    \item \textbf{Berkeley} The dataset provides 96 images and 100 instance masks, and includes common objects shown in various scenarios. 
    \item \textbf{COCO Mval.} It is a subset of the COCO dataset that contains 800 images and 10 objects from each object category, while each image only contains one object mask.
    
\end{itemize}

For the MOIS setting, in addition to the extended dataset, as we introduce in the following Sec.~\ref{sec:mois}, we utilized one existing dataset, named HIM2K \citep{sun2022humanvmultiple}, from the human editing domain. 
This dataset is composed of two subsets: synthetic images and natural images. 
We selected the natural image subset as one of our evaluation datasets for MOIS. 
It contains 320 images with 830 humans, with each image including mask annotations for two to ten humans. 
During the evaluation for a given image, we randomly selected one human and its corresponding mask as the \textit{exemplar target} for the remaining humans.
Note that the HIM2K dataset was only adopted for evaluating the MOIS models without any domain fine-tuning.

\subsubsection{Extended Dataset COCO MOIS}
\label{sec:mois}
There is a lack of large-scale datasets dedicated to evaluating the effectiveness of the MOIS task.
Hence, we introduce a new dataset, named COCO MOIS, by constructing the masks and clicks on the COCO 2017 dataset \citep{lin2014coco}. 
COCO is a comprehensive dataset that includes pixel-level annotations belonging to 80 different categories across various domains, designed for instance segmentation, semantic segmentation, and object detection tasks. 
We converted the previous annotations to a triplet consisting of an image, a randomly selected single object with the mask and simulated interactions, and all objects in the same category as $\{I, \{M^e, C^e_p, C^e_n\}, M^{all}_O\}$. 
Note that we did not employ real users' interactive exemplars as the previously interacted objects for two reasons: (1) the collection of real interactive exemplars is expensive; (2) achieving the target IoUs within 20 clicks for multiple objects from the COCO dataset is challenging, leading to unsatisfactory masks.
Hence, we automatically simulated the clicks, following the random strategy proposed in DIOS \citep{xu2016dios} based on the collected mask.
We restricted the total number of clicks for one exemplar, including both positive and negative clicks, to a random value within the range of 1 to 20.
Alg.~\ref{alg:mois} illustrates the detailed construction procedure of the COCO MOIS dataset.
For the same image, $x_i$ and $x_j$ are different samples if the objects belong to different categories. 
This results in 111,362 training samples from 76,110 images and 4,777 validation samples from 3,249 images, with each sample containing an average of 3.98 target objects. 
It is worth mentioning that we have conducted an ablation study in Sec.~\ref{sec:ablation_study} to assess the impact of the previously interacted objects (\textit{exemplar} in our case) on the segmentation performance.
The dataset will be released for future research purposes.

\begin{algorithm}
\caption{Construction of COCO MOIS}
\label{alg:mois}
\begin{algorithmic}
\State \textbf{Input:} Dataset $\bm{D}$ with mask annotations from COCO 2017
\\// {\color{gray}Find the samples with multiple instances in the same category.}
\For{$\bm{x_i} \{\bm{I_i},\bm{C_i},\bm{M_i}\} \in \bm{D}$} // {\color{gray}$\bm{I}$, $\bm{C}$, $\bm{M}$ denote the \textit{image}, \textit{category}, and \textit{mask} for a sample $\bm{x_i}$, respectively.}
\For{$\bm{x_j} \{\bm{I_j},\bm{C_j},\bm{M_j}\} \in \bm{D}$}
\If{$\bm{I_i} == \bm{I_j} \ and \ \bm{C_i} == \bm{C_j} \ and \ \bm{M_i} != \bm{M_j}$}
\State Collect unique samples $\{\bm{I}, \bm{C}\}$
\EndIf
\EndFor
\EndFor
\\// {\color{gray}Obtain mask, clicks from COCO annotations.}
\For{$\bm{y} \in \{\bm{I}, \bm{C}\}$}
\While{$\bm{y} \cap \bm{D} \ != \ \varnothing$}
\If{$\bm{I_y} == \bm{I_x} \ and \ \bm{C_y} == \bm{C_x}$} // {\color{gray}$\bm{x}$ from $\bm{D}$.}
\State Obtain valid masks with $coco.loadAnns$
\State Filter small masks with $area \ \textless \ 400$
\State Update $\bm{O^d}$ // {\color{gray}$\bm{O^d}$ denotes a valid object for $\{\bm{I}, \bm{C}\}$.}
\EndIf
\EndWhile
\State Generate mask samples $\bm{M^{all}_O = }\{\bm{M_O^1}, \bm{M_O^2}, ..., \bm{M_O^d}\}$
\State Simulate clicks with $cluster.kmedoids$ and sampling strategies
\EndFor
\end{algorithmic}
\end{algorithm}
\vspace{-4mm}

\subsection{Evaluation Protocol}
\label{sec:evaluation_metrics}
To evaluate the performance of different IS methods, we adopted both manual and automatic inferences to qualitatively assess the real results and made a quantitative comparison based on the automatically simulated interactions. The main evaluation metrics widely used are \textbf{NoC} (Number of Clicks), \textbf{NoF} (Number of Failures), and \textbf{mIoU} (mean Intersection over Union).

\vspace{2mm}
\noindent
\textbf{NoC \& NoF.}
NoC evaluates the user's efforts in achieving a certain IoU threshold within the maximum number of clicks. NoF represents the number of instances where the model fails to reach a set target IoU after the maximum rounds of clicks, indicating the method's stability. For the evaluation of MOIS, as multiple objects in the same category are from the same image, we extended NoF to \textbf{NoFI} (Number of Failed Images) that describes the overall interaction efforts from the perspective of images. We set two IoU thresholds (85\% and 90\%) and a maximum of 20 clicks, consistent with previous works \citep{xu2016dios, sofiiuk2020fbrs, sofiiuk2022ritm}. The lower the NoC, NoF, and NoFI, the better.

\vspace{2mm}
\noindent
\textbf{mIoU.}
In addition to evaluating interaction efforts, the segmentation performance is also important for IS. We adopted mIoU to measure the segmentation quality. For the automatic models without further additional clicks, we compared the predicted mask with the ground truth to calculate the mIoU. On the other hand, mIoU$\circledast k$ was calculated with provided $k$ clicks for interactive models. To distinguish MOIS from SOIS, we used mIoU$\circledast k+$ to represent the segmentation quality given $k$ additional clicks based on a previously interacted object. The higher the value of mIoU$\circledast k$ and mIoU$\circledast k+$, the better.

\subsection{Implementation Details}
\label{sec: imple details}
All the experiments were implemented on a PyTorch platform with two NVIDIA A40 GPUs. For different backbones, including ViT \citep{dosovitskiy2020vit}, Swin \citep{liu2021swin}, and ResNet \citep{he2016resnet}, we used the pre-trained models on ImageNet \citep{deng2009imagenet} from their official repositories.
We retained the original hyper-parameters of the backbones.
The images with different sizes were cropped and resized to $448 \times 448$ pixels.
For model training, we employed several data augmentation strategies: random flipping, rotation, cropping as well as random resizing with a scale from 0.75 to 1.25.
The Adam optimizer \citep{kingma2014adam} with $\beta_1=0.9$ and $\beta_2=0.99$ was adopted. We also used the Normalized Focal Loss \citep{sofiiuk2019nfl} (NFL) as the loss function for model training, following the recent works \citep{chen2022focalclick, sofiiuk2022ritm}.
We set the batch size to 24 for all compared models. For SOIS, our iCMFormer model was trained on SBD and COCO + LVIS with 55 and 85 epochs, respectively.
The initial learning rate was set to 0.00005 and decreased by 10 times after the epoch of 50.
For MOIS, we trained the iCMFormer++ model for 105 epochs on the extended COCO MOIS dataset with the same learning rate setting.
For the click sampling of \textit{exemplars}, we followed the hyperparameters in DIOS \citep{xu2016dios} such as the minimum distance (40 pixels) between two clicks, and the distance (20 pixels) from an object border.
For the image region of an \textit{exemplar} converted from its satisfactory mask, we employed multiple scales as \{0.8, 1, 1.2\} for the bounding boxes.

\subsection{Quantitative Results and Comparison}
We carry out a comprehensive comparison with the state-of-the-art IS methods on different datasets and report the results in terms of interaction efforts \ie, NoC, NoF, and NoFI, and segmentation performance measured in mIoU based on various evaluation settings.

\subsubsection{Single Object Interactive Segmentation}
\label{sec: single results}
We compared our results with previous click-based IS methods on four benchmarks for SOIS.
Note that the maximum number of clicks is set to 20 for NoC85 and NoC90 even when the results cannot reach the target IoUs, which is consistent with the previous works \citep{xu2016dios, chen2021cdnet, sofiiuk2022ritm}.

The comparison results on GrabCut, Berkeley, SBD, and COCO MVal concerning the number of clicks (NoC) are reported in Table~\ref{tab:comparison with single SOTA}. 
As some of the methods were trained on different datasets, for a fair comparison, we split the table into two sections -- one for the single dataset, \ie, PASCAL \citep{everingham2010pascal} and SBD, and the other for a collective dataset, \ie, COCO + LVIS or SAM \citep{kirillov2023sam}. 
We also report the backbones of different methods to indicate the importance of feature extraction. 
When trained on SBD, iCMFormer achieved state-of-the-art performance on three out of the four evaluation datasets. 
For instance, iCMFormer is the first to reach 90\% IoU requiring a maximum of four clicks on COCO MVal. 
Moreover, iCMFormer achieved competitive results when trained on the large-scale COCO + LVIS dataset. 
These results demonstrate iCMFormer's effectiveness for SOIS.

\begin{figure*}[ht]
  \centering
    \includegraphics[trim= 42pt 14pt 86pt 36pt, clip=True, width=0.32\linewidth]{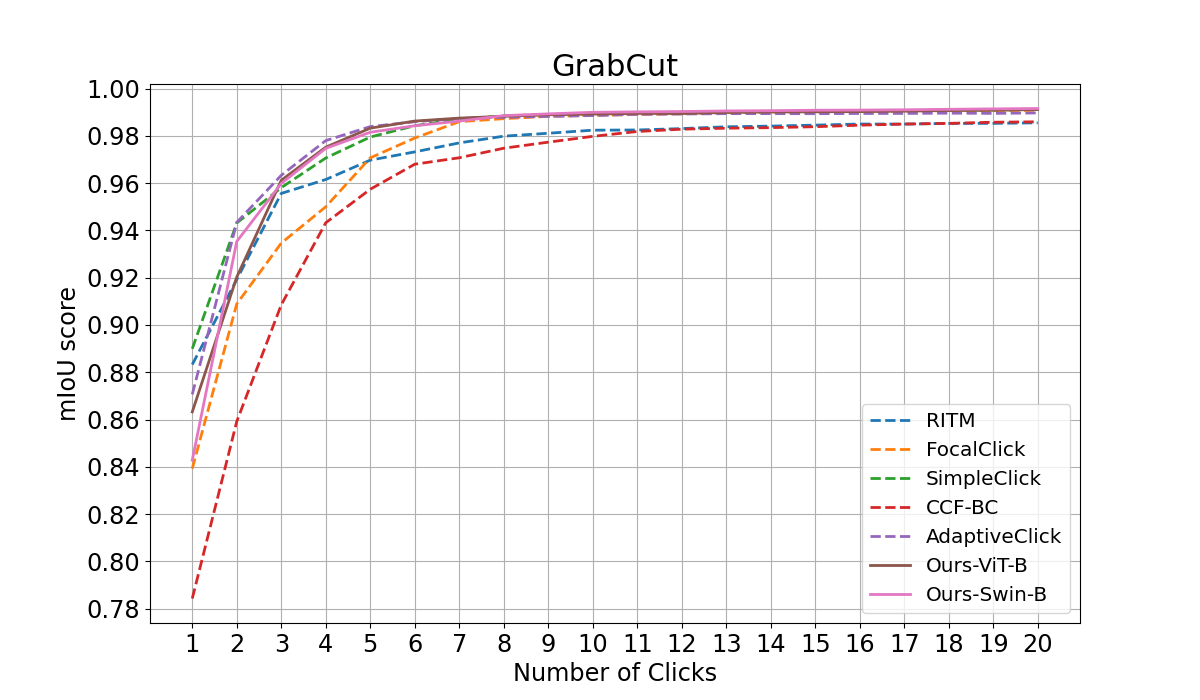}
  \hspace{2mm}
    \includegraphics[trim= 42pt 14pt 86pt 36pt, clip=True, width=0.32\linewidth]{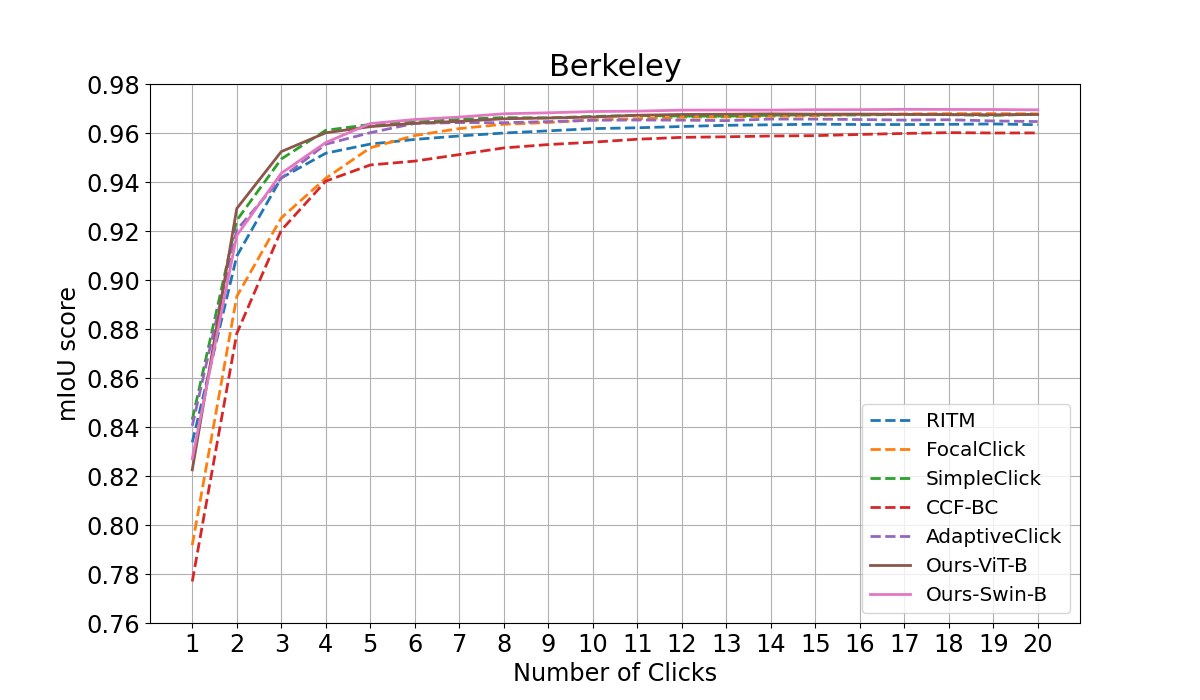}
  \hspace{2mm}
    \includegraphics[trim= 42pt 14pt 86pt 36pt, clip=True, width=0.32\linewidth]{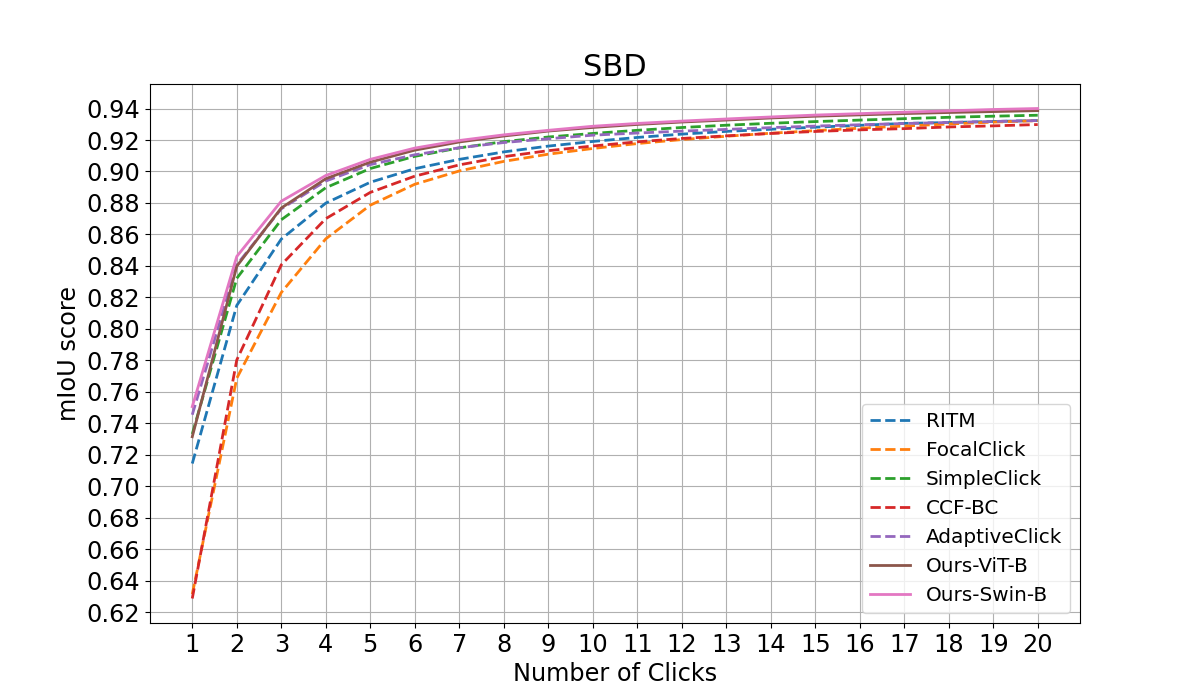}
  \vspace{1mm}
  
    \includegraphics[trim= 42pt 14pt 86pt 36pt, clip=True, width=0.32\linewidth]{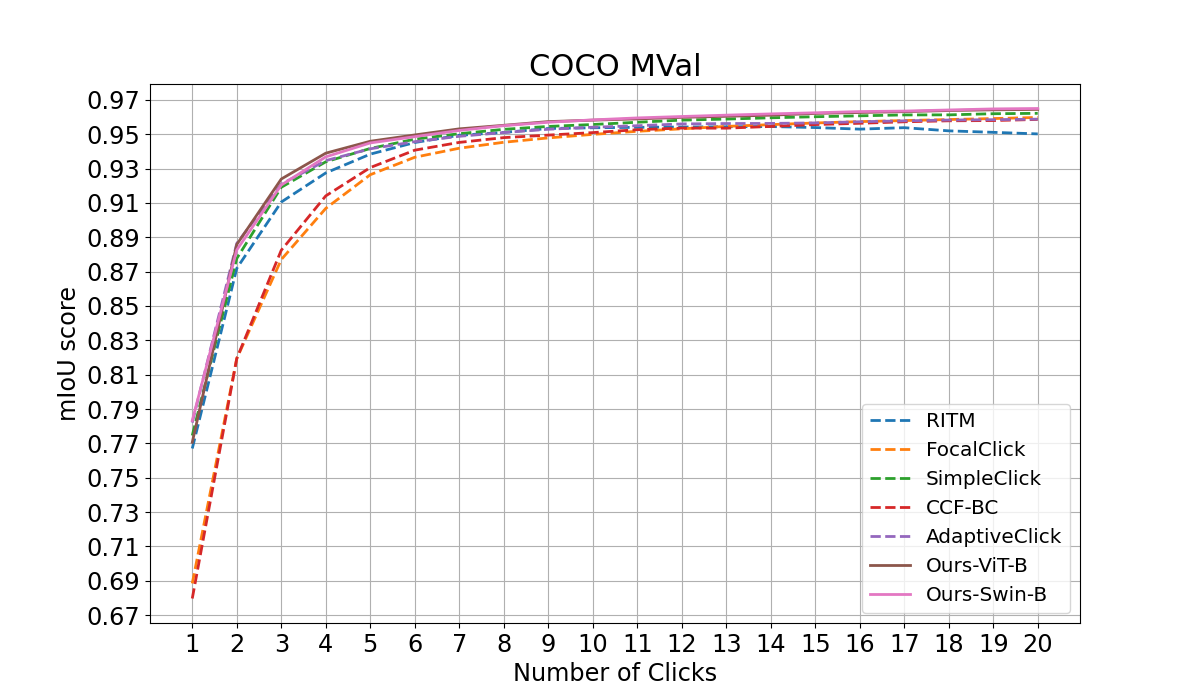}
  \hspace{2mm}
    \includegraphics[trim= 42pt 14pt 86pt 36pt, clip=True, width=0.32\linewidth]{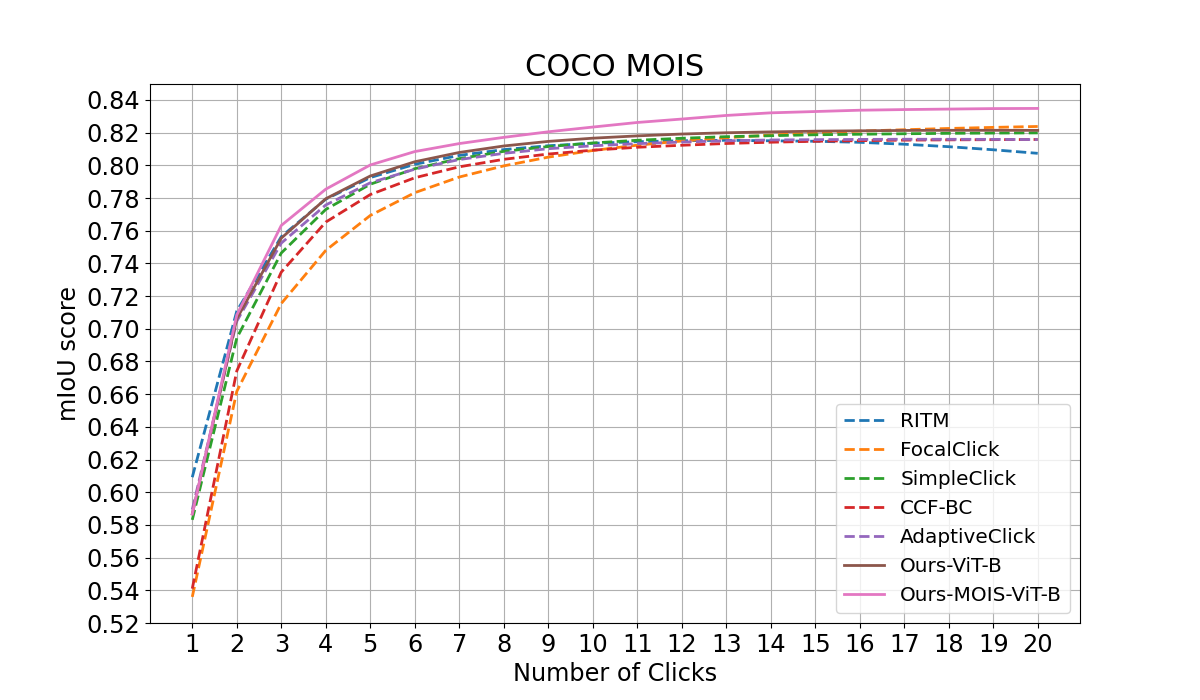}
  \hspace{2mm}
    \includegraphics[trim= 42pt 14pt 86pt 36pt, clip=True, width=0.32\linewidth]{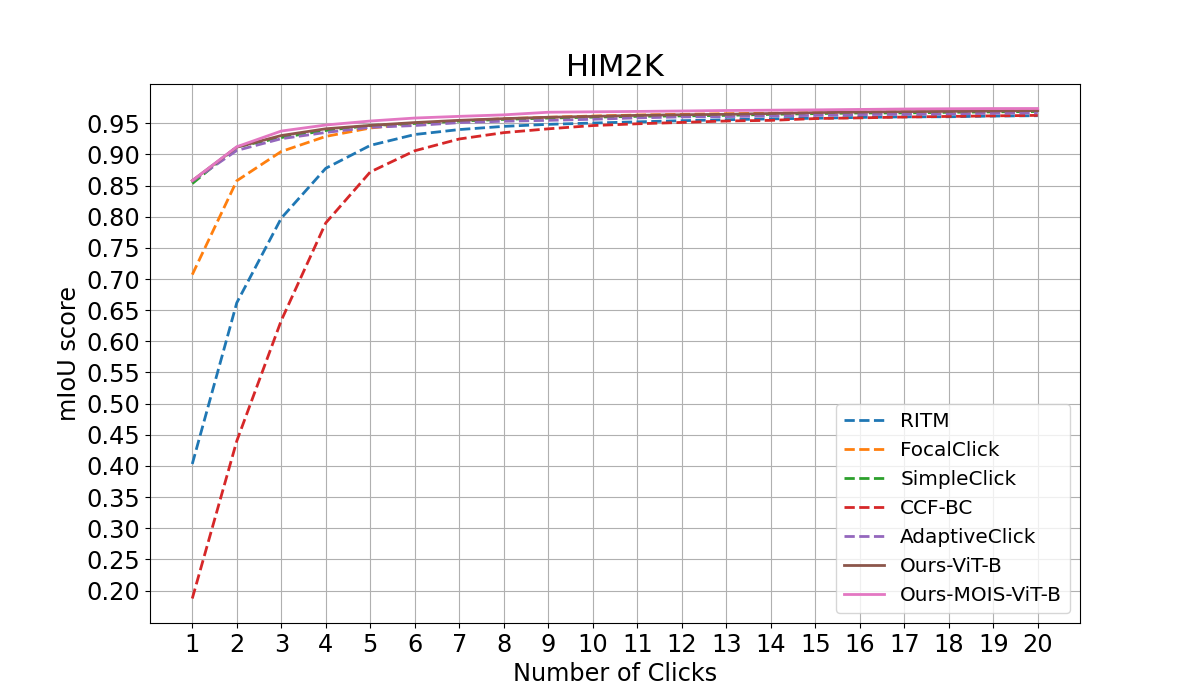}
  
  \caption{Convergence analysis of the mean IoU (mIoU$\circledast k$) curves for varying number of clicks. The evaluation results on GrabCut \citep{rother2004grabcut}, Berkeley \citep{mcguinness2010berkeley}, SBD \citep{hariharan2011sbd}, COCO MVal, COCO MOIS and HIM2K \citep{sun2022humanvmultiple} are provided. The higher starting point typically leads to better results with the first positive click. A steeper slope indicates that the method requires fewer clicks to achieve better segmentation results.}
  \label{fig:anaysis}
\end{figure*}

Figure~\ref{fig:anaysis} illustrates the segmentation performance on the evaluation datasets measured in mean IoU curves with progressively added clicks. 
We can observe that, compared to the other models, iCMFormer achieved better mean IoU scores with the same number of clicks, and required fewer clicks to reach the same target IoU. 
For instance, the iCMFormer ViT-B model improves the mIoU performance to ca. 75\% on SBD requiring only one click.
This superiority is clearly maintained in the first five clicks.

\begin{table*}
  \caption{Single object interactive segmentation evaluation results on GrabCut \citep{rother2004grabcut}, Berkeley \citep{mcguinness2010berkeley}, SBD \citep{hariharan2011sbd}, and COCO MVal. NoC85 and NoC90 denote the average number of clicks to reach target IoUs (85\% and 90\%). The best results are in \textbf{bold} and the second best are \underline{underlined}. (- indicates neither models nor numbers are provided by original works and $\dagger$ indicates the results generated by our re-implementation.)}
  \centering
  \setlength\tabcolsep{4pt}
  \begin{tabular}{p{2.35cm}ccccccccccc}
  \toprule
  \multirow{2}*{Method} & \multirow{2}*{Venue} & \multicolumn{1}{c}{Training} & \multirow{2}*{Backbone} & \multicolumn{2}{c}{GrabCut} & \multicolumn{2}{c}{Berkeley} & \multicolumn{2}{c}{SBD} & \multicolumn{2}{c}{COCO MVal}\\
 & & Data & & NoC85 &  NoC90 & NoC85 &  NoC90 & NoC85 &  NoC90 & NoC85 &  NoC90\\
 \midrule
  DIOS   & CVPR16 & Pascal & FCN & - & 6.04 & - & 8.65 & - & - & 13.99 & 16.88 \\
  LD   & CVPR18 & SBD & VGG-19 & 3.20 & 4.79 & - & -  & 7.41 & 10.78 & 7.86 & 12.45\\
  BRS  & CVPR19 & SBD & DenseNet & 2.60 & 3.60 & - & 5.08 & 6.59 & 9.78 & 5.10 & 7.92\\
  f-BRS   & CVPR20 & SBD & ResNet-101 & 2.30 & 2.72 & - & 4.57 & 4.81 & 7.73 & 4.59 & 6.47 \\
  CDNet  & ICCV21 & SBD & ResNet-50  & 2.22 & 2.64 & - & 3.69 & 4.37 & 7.87 & - & -\\
  RITM  & ICIP22 & SBD & HRNet-18  & 1.76 & 2.04 & 1.87 & 3.22 & 3.39 & 5.43 & 3.31 & 4.38\\
  FocalClick   & CVPR22 & SBD & HRNet-18s-S2 & 1.86 & 2.06 & - & 3.14 & 4.30 & 6.52 & 3.81 & 5.12 \\
  GPCIS & CVPR23 & SBD & SegF-B0-S2 & 1.60 & 1.76 & 1.84 & 2.70 & 4.16 & 6.28 & 3.18 & 4.27 \\
  FCFI & CVPR23 & SBD & ResNet-101 & 1.64 & 1.80 & - & 2.84 & 3.26 & 5.35 & 3.27 & 4.31  \\
  SimpleClick & ICCV23 & SBD & ViT-B & 1.40 & 1.54 & 1.44 & 2.46 & 3.28 & 5.24 & 3.12 & 4.07 \\
  CCF-BC & TIP23 & SBD & HRNet-18  & 1.58 & 1.78 & - & 2.88 & 3.38 & 5.33 & 3.26 & 4.50\\ 
  AdaptiveClick & TNNLS24 & SBD & ViT-B & \underline{1.38} & \underline{1.46} & \textbf{1.38} & \textbf{2.18} & \underline{3.22} & \underline{5.22} & 3.15 & 4.13 \\ 
  iCMFormer(Ours) & ICCVW23 & SBD & ViT-B & \textbf{1.36} & \textbf{1.42} & \underline{1.42} & 2.52 & 3.33 & 5.31 & \textbf{2.75} & \textbf{3.75}  \\
  iCMFormer(Ours) & ICCVW23 & SBD & Swin-B & 1.46 & 1.50 & 1.52 & \underline{2.32} & \textbf{3.21} & \textbf{5.16} & \underline{3.04} & \underline{4.00} \\
  \midrule
  RITM  & ICIP22 & COCO+LVIS & HRNet-32 & 1.46 & 1.56 & - & 2.10 & 3.59 & 5.71 & 2.16 & 2.97 \\
  FocalClick   & CVPR22 & COCO+LVIS & SegF-B0-S2 & 1.40 & 1.66 & - & 2.27 & 4.56 & 6.86 & 2.82 & 3.82  \\ 
  FCFI & CVPR23 & COCO+LVIS & HRNet-18 & \underline{1.38} & 1.46 & - & 1.96 & 3.63 & 5.83 & 2.44 & 3.19 \\
  DynaMITe & ICCV23 & COCO+LVIS & SegF-B0 & 1.58 & 1.68 & 1.61 & 2.06 & 3.89 & 6.48 & 2.47 & 3.28 \\
  SimpleClick & ICCV23 & COCO+LVIS & ViT-B & \underline{1.38} & \underline{1.48} & \underline{1.36} & 1.97 & 3.43 & 5.62 & 2.18 & 2.93 \\ 
  SAM & ICCV23 & SAM & ViT-B & 2.06$\dagger$ & 2.12$\dagger$ & 1.92$\dagger$ & 2.49$\dagger$ & 6.50 & 9.76 & 3.10$\dagger$ & 4.91$\dagger$ \\    
  SEEM & NeurIPS23 & COCO+LVIS & DaViT-B & 2.02$\dagger$ & 2.07$\dagger$ & 1.86$\dagger$ & 2.23$\dagger$ & 6.67 & 9.99 & 3.02$\dagger$ & 4.57$\dagger$ \\
  CCF-BC & TIP23 & COCO+LVIS & HRNet-32 & 1.40 & 1.49 & - & 1.90 & 3.49 & 5.53 & 2.65 & 3.61  \\ 
  AdaptiveClick & TNNLS24 & COCO+LVIS & ViT-B & \textbf{1.32} & \textbf{1.38} & \textbf{1.32} & \textbf{1.64} & \textbf{2.84} & \textbf{4.68} & \underline{2.07} & 2.86 \\ 
  iCMFormer(Ours) & ICCVW23 & COCO+LVIS & ViT-B & 1.42 & 1.52 & 1.40 & \underline{1.86} & \underline{3.29} & \underline{5.30} & 2.08 & \underline{2.85}  \\
  iCMFormer(Ours) & ICCVW23 & COCO+LVIS & Swin-B & 1.46 & 1.60 & 1.49 & 1.93 & 3.34 & 5.35 & \textbf{2.02} & \textbf{2.69}  \\
  \bottomrule
  \end{tabular}
  \label{tab:comparison with single SOTA}
  \vspace{-2mm}
\end{table*}

\begin{table*}
  \caption{Multiple object interactive segmentation evaluation results on COCO MOIS and HIM2K \citep{sun2022humanvmultiple} for collective interaction efforts. NoF85 and NoF90 denote the average number of failures that cannot reach target IoUs (85\% and 90\%) after 20 clicks. The best results within a group are in \textbf{bold}, and the overall best results are in \textcolor{red}{red}.}
  \centering
  \setlength\tabcolsep{4pt}
  \begin{tabular}{p{2.35cm}ccccccccc}
  \toprule
  \multirow{2}*{Method} & \multirow{2}*{Venue} & \multicolumn{1}{c}{Training} & \multirow{2}*{Backbone} & \multicolumn{3}{c}{COCO MOIS} & \multicolumn{3}{c}{HIM2K}\\
 & & Data & & NoC85 &  NoC90 & NoF85 & NoC85 &  NoC90 & NoF90\\
 \midrule
  LD   & CVPR18 & SBD & VGG-19 & 16.46 & 18.78 & 13747 & 6.34 & 8.15 &  47\\
  BRS  & CVPR19 & SBD & DenseNet & 14.84 & 17.73 & 12690 & 4.81 & 6.03 & 35\\
  f-BRS   & CVPR20 & SBD & ResNet-101 & 14.62 & 17.40 & 12524 & 4.66 & 5.85 & 30\\
  RITM  & ICIP22 & SBD & HRNet-18 & 13.33 & 16.68 & 10414 & 2.96 & 4.08 & 14\\
  FocalClick   & CVPR22 & SBD & HRNet-18s-S2 & 13.87 & 17.28 & 
11205 & 3.93 & 4.16 & 28\\
  GPCIS & CVPR23 & SBD & SegF-B0-S2 & 13.75 & 17.06 & 10984 & 3.63 & 4.11 & 26\\
  FCFI & CVPR23 & SBD & ResNet-101 & 13.68 & 16.94 & 10788 & 2.47 & 3.80 & 20\\
  SimpleClick & ICCV23 & SBD & ViT-B & 13.19 & 16.51 & 10306 & 1.74 & \textbf{2.59} & \textbf{11}\\
  CCF-BC & TIP23 & SBD & HRNet-18 & 13.51 & 16.84 & 10522 & 4.47 & 6.02 & 25\\
  AdaptiveClick & TNNLS24 & SBD & ViT-B & 13.21 & 16.47 & 10312 & \textbf{1.72} & 2.60 & 15\\
  iCMFormer(Ours) & ICCVW23 & SBD & ViT-B & \textbf{13.05} & \textbf{16.44} & \textbf{10252} & 1.75 & 2.63 & 15 \\
\midrule
  RITM  & ICIP22 & COCO+LVIS & HRNet-32 & 12.44 & 16.13 & 9991 & 3.47 & 4.58 & 12\\
  FocalClick   & CVPR22 & COCO+LVIS & SegF-B0-S2 & 13.08 & 16.52 & 10232 & 2.50 & 3.40 & 10 \\ 
  FCFI & CVPR23 & COCO+LVIS & HRNet-18 & 12.85 & 16.57 & 10467 & 1.98 & 2.83 & 11\\
  DynaMITe & ICCV23 & COCO+LVIS & SegF-B0 & 12.55 & 16.12  & 10013 & 1.65 & 2.50 & 10\\
  SimpleClick & ICCV23 & COCO+LVIS & ViT-B & 12.51 & 16.16  & 10006 & 1.68 & 2.49 & \textbf{8}\\ 
  CCF-BC & TIP23 & COCO+LVIS & HRNet-32 & 13.01 & 16.55 & 10260 & 4.68 & 5.77 & 14\\
  AdaptiveClick & TNNLS24 & COCO+LVIS & ViT-B & 12.44 & 16.10 & 
 9960 & 1.66 & 2.53 & 14\\
 iCMFormer(Ours) & ICCVW23 & COCO+LVIS & ViT-B  & \textbf{12.33} & \textbf{16.01} & \textbf{9869} & \textcolor{red}{1.63} & \textcolor{red}{2.41} &  9\\
\midrule
  iCMFormer++(Ours) & 2024 & COCO MOIS & ViT-B & \textcolor{red}{10.36} & \textcolor{red}{13.84} & - & 2.03 & 2.75 & -\\

  \bottomrule
  \end{tabular}
  \label{tab:comparison with multiple SOTA}
  \vspace{-2mm}
\end{table*}

\begin{table*}
  \caption{Evaluation results on COCO MOIS and HIM2K \citep{sun2022humanvmultiple} in terms of additional interaction efforts and segmentation performance for multiple objects in the same category. NoFI85 and NoFI90 denote the average numbers of failed images that cannot reach target IoUs (85\% and 90\%) for all objects after 20 clicks. The best results are \textbf{bold} while the second best are \underline{underlined}.}
  \centering
  \setlength\tabcolsep{3pt}
  \begin{tabular}{p{2.35cm}ccccc|ccccc}
  \toprule
  \multirow{2}*{Method}
  & \multicolumn{5}{c}{COCO MOIS} & \multicolumn{5}{c}{HIM2K}\\
 & NoC85$\downarrow$ &  NoC90$\downarrow$ & NoFI85$\downarrow$ 
 & mIoU$\circledast$5+$\uparrow$
 &  mIoU$\circledast$10+$\uparrow$
 & NoC85$\downarrow$ &  NoC90$\downarrow$ & NoFI90$\downarrow$
 & mIoU$\circledast$1+$\uparrow$
 & mIoU$\circledast$5+$\uparrow$\\
 \midrule
  RITM  & 10.25 & 14.41 & 1853 & 81.20\% & 84.16\% & 2.70 & 4.03 & 15 & 44.20\% & 93.55\%\\
  FocalClick  & 11.07 & 14.89 & 1946 & 78.17\% & 83.07\% & 1.29 & 1.73 & \underline{4} & \underline{88.66\%} & 95.94\% \\
  SimpleClick  & 9.95 & 14.10 & 1798 & \underline{81.60\%} & 84.52\% & 1.31 & 2.34 & 12 & 87.92\% & 95.07\%\\ 
  AdaptiveClick  & 10.36 & 14.41 & 1897 & 80.83\% & 83.69\% & \underline{1.20} & \underline{1.46} & \textbf{3} & 88.41\% & \underline{96.12\%} \\
  iCMFormer(Ours)  & \underline{9.91} & \underline{14.04} & \underline{1787} & 81.58\% & \underline{84.61\%} & 1.34 & 2.27 & 12 & 88.50\% & 95.19\%\\
  iCMFormer++(Ours)  & \textbf{8.47} & \textbf{11.92} & \textbf{1561} & \textbf{82.47\%} & \textbf{85.89\%} & \textbf{1.16} & \textbf{1.38} & \underline{4} & \textbf{88.73\%} & \textbf{96.31\%}\\
  \bottomrule
  \end{tabular}
  \label{tab:comparison with SOTA for additional clicks}
\end{table*}

\subsubsection{Multiple Object Interactive Segmentation}
\label{sec: multiple results}
We compared the proposed iCMFormer++ model with other models on two datasets for \textit{collective interaction} and \textit{additional interaction} efforts based on the introduced \textit{exemplars} for MOIS. 
Specifically, the collective interaction efforts count the average NoC and NoF to achieve a target IoU for all objects sharing the same category in one image.
The other previous models use a conventional interactive way, ignoring the category information for the \textit{recall} objects, while iCMFormer++ achieves this task based on the predicted \textit{exemplars}.
Namely, iCMFormer++ adopts iCMFormer to predict the \textit{exemplar} on a randomly selected object for each image.
Since the \textit{exemplar} may require multiple clicks to obtain a satisfactory mask, we restricted the maximum number of clicks to five and stopped adding more clicks if the improved IoU was less than 1\% in the next click.
In order to more explicitly analyze the performance gain brought by the \textit{exemplars}, the additional interaction efforts count the extra average NoC and NoFI, providing an \textit{exemplar} as previous masks to all the comparison models (similar to that introduced in Sec.~\ref{sec:IS structure}). 
In this evaluation setting, we re-trained the other models on COCO MOIS, feeding them with the so-called ``golden" \textit{exemplars} from the ground truth consisting of a mask and simulated clicks for a target object.
To ensure a fair comparison, we also used the golden \textit{exemplars} for iCMFormer++.
Note that we still set the maximum number of clicks to 20 for NoC85 and NoC90.

We report the collective interaction efforts for the models (including our iCMFormer model) in Table~\ref{tab:comparison with multiple SOTA}.
It can be seen that iCMFormer achieved the best performance on COCO MOIS among all the models trained on SBD, indicating the superiority of iCMFormer on the MOIS task even without the help of \textit{exemplars}.
iCMFormer++ achieved a significantly better performance than all the other models and outperformed iCMFormer by reducing ca. two clicks for NoC85 and NoC90 on COCO MOIS. 
Interestingly, on the relatively smaller and easier dataset HIM2K, iCMFormer++ only required 2.03 and 2.75 clicks to reach NoC85 and NoC90, respectively, requiring much fewer clicks to obtain a similar performance on COCO MOIS.
But iCMFormer++ was slightly inferior to the other models.
Recall the prediction of \textit{exemplars}, we conjecture that iCMFormer++ was overshadowed by the interactions focusing on the \textit{exemplars}, \ie, it spent most of the initial efforts to obtain a satisfactory mask for the \textit{exemplars}.
This phenomenon is more visible in the lower right subfigure of Fig.~\ref{fig:anaysis}. 
The mIoU gains from the first to the second click and from the second to the third click are both larger than 1\%. 
On the other hand, this subfigure also shows that iCMFormer++ reached higher mean IoUs than the other models in the initial clicks.

We report the additional interaction efforts in Table~\ref{tab:comparison with SOTA for additional clicks}.
iCMFormer++ set a new state-of-the-art performance on the two datasets for MOIS.
For instance, iCMFormer++ only required 11.92 clicks to reach NoC90 on COCO MOIS, which improved the previous result by 15\%. 
Additionally, it achieved higher segmentation performance with the same number of additional clicks compared to the others, \eg, 88.73\% mIoU with five additional clicks on HIM2K. 
This result on HIM2K also indirectly proves our conjecture that major initial efforts are allocated to obtain a satisfactory mask for the \textit{exemplars} in the evaluation setting of the collective interaction efforts.
Overall, the above results demonstrate the effectiveness of our proposed method for MOIS.

As a practical annotation tool, it is vital to obtain high-quality segmentation masks if provided with sufficient clicks.
So, we measured the performance of models' stability with the number of failures (NoF) for SOIS and report the numbers in Table~\ref{tab:comparison with multiple SOTA}.
iCMFormer trained on different datasets reduced the failure cases and outperformed the previous methods.
Similarly, we report the number of failed images (NoFI) for MOIS in Table~\ref{tab:comparison with SOTA for additional clicks}. 
Remarkably, iCMFormer++ reduced the failure number to 1,561 on COCO MOIS, which succeeded in segmenting 237 more images than that of the previous SimpleClick \citep{liu2023simpleclick} model.

\begin{figure*}[ht]
  \centering
  \begin{subfigure}[t]{0.22\linewidth}
    \includegraphics[width=1\linewidth]{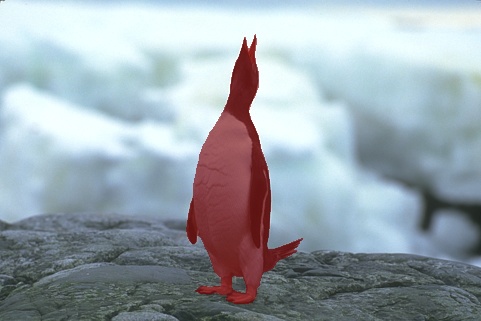}
    \subcaption*{GT}
  \end{subfigure}
  \hspace{2mm}
  \begin{subfigure}[t]{0.22\linewidth}
    \includegraphics[width=1\linewidth]{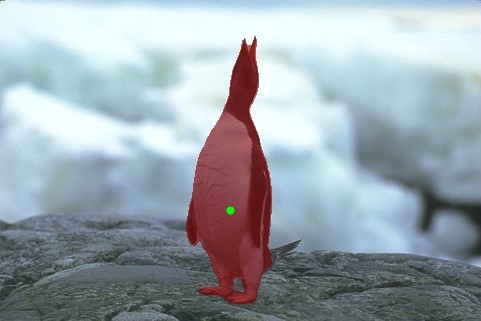}
    \subcaption*{1 click 93.2\%}
  \end{subfigure}
  \hspace{2mm}
  \begin{subfigure}[t]{0.22\linewidth}
    \includegraphics[width=1\linewidth]{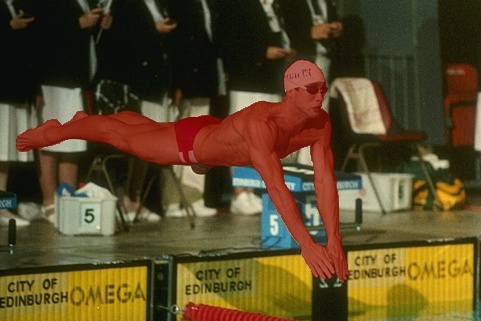}
    \subcaption*{GT}
  \end{subfigure}
  \hspace{2mm}
  \begin{subfigure}[t]{0.22\linewidth}
    \includegraphics[width=1\linewidth]{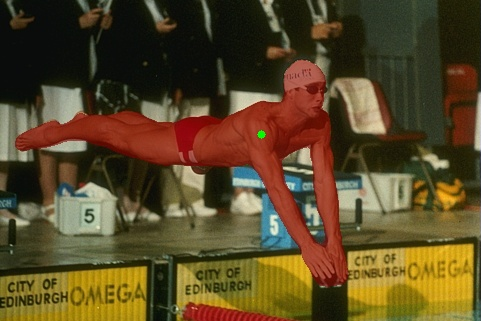}
    \subcaption*{1 click 93.4\%}
  \end{subfigure}

  \begin{subfigure}[t]{0.22\linewidth}
    \includegraphics[width=1\linewidth]{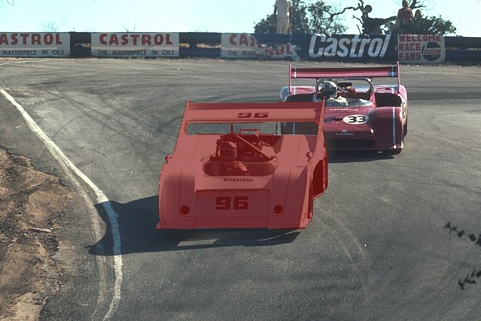}
    \subcaption*{GT}
  \end{subfigure}
  \hspace{2mm}
  \begin{subfigure}[t]{0.22\linewidth}
    \includegraphics[width=1\linewidth]{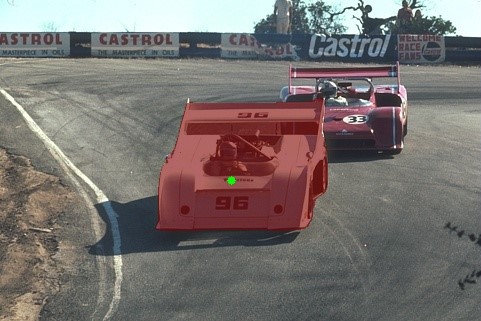}
    \subcaption*{1 click 89.9\%}
  \end{subfigure}
  \hspace{2mm}
  \begin{subfigure}[t]{0.22\linewidth}
    \includegraphics[width=1\linewidth]{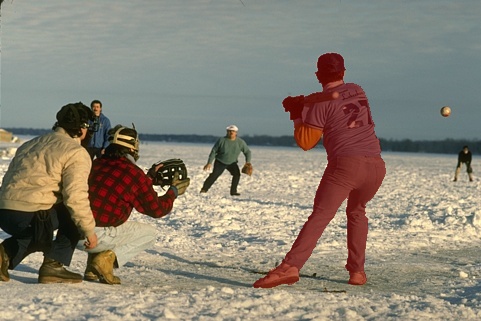}
    \subcaption*{GT}
  \end{subfigure}
  \hspace{2mm}
  \begin{subfigure}[t]{0.22\linewidth}
    \includegraphics[width=1\linewidth]{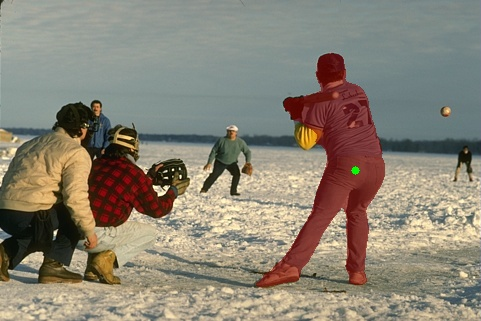}
    \subcaption*{1 click 90.6\%}
  \end{subfigure}

  \begin{subfigure}[t]{0.22\linewidth}
    \includegraphics[width=1\linewidth]{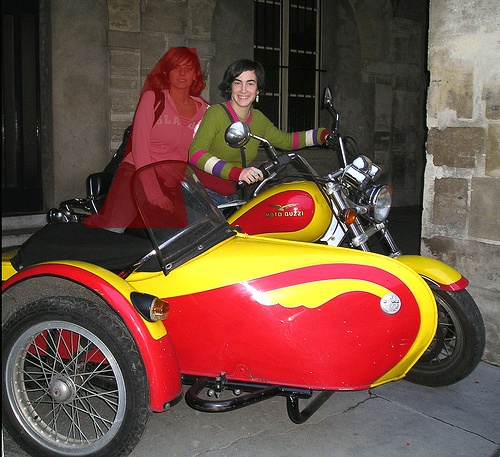}
    \subcaption*{GT}
  \end{subfigure}
  \hspace{2mm}
  \begin{subfigure}[t]{0.22\linewidth}
    \includegraphics[width=1\linewidth]{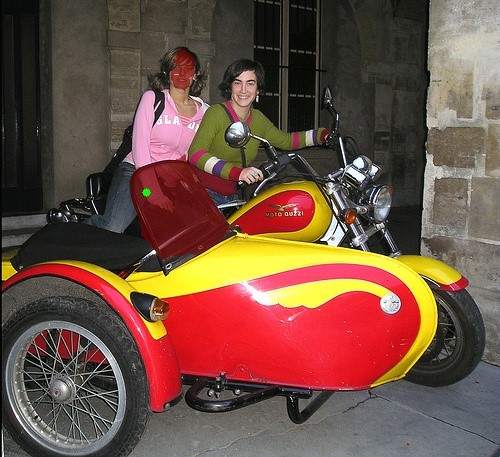}
    \subcaption*{1 click 25.6\%}
  \end{subfigure}
  \hspace{2mm}
  \begin{subfigure}[t]{0.22\linewidth}
    \includegraphics[width=1\linewidth]{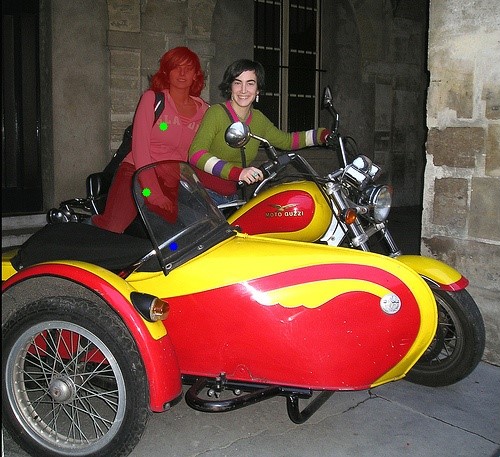}
    \subcaption*{3 click 80.5\%}
  \end{subfigure}
  \hspace{2mm}
  \begin{subfigure}[t]{0.22\linewidth}
    \includegraphics[width=1\linewidth]{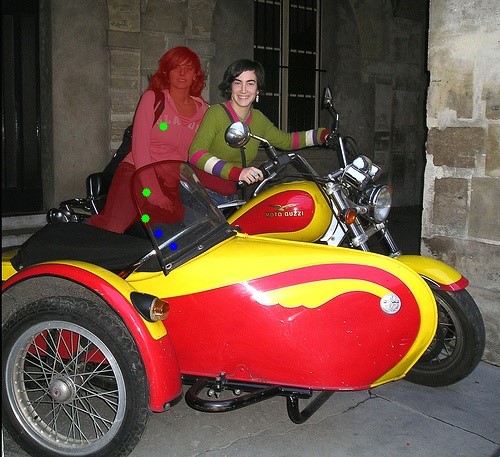}
    \subcaption*{5 clicks 90.5\%}
  \end{subfigure}

  \begin{subfigure}[t]{0.22\linewidth}
    \includegraphics[width=1\linewidth]{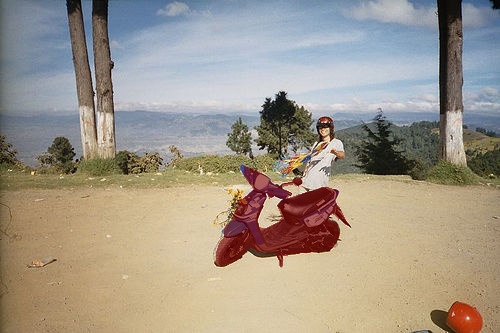}
    \subcaption*{GT}
  \end{subfigure}
  \hspace{2mm}
  \begin{subfigure}[t]{0.22\linewidth}
    \includegraphics[width=1\linewidth]{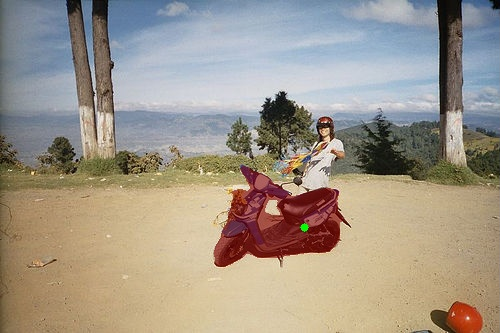}
    \subcaption*{1 click 82.3\%}
  \end{subfigure}
  \hspace{2mm}
  \begin{subfigure}[t]{0.22\linewidth}
    \includegraphics[width=1\linewidth]{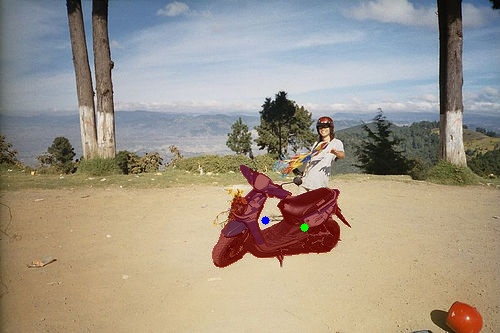}
    \subcaption*{3 clicks 87.1\%}
  \end{subfigure}
  \hspace{2mm}
  \begin{subfigure}[t]{0.22\linewidth}
    \includegraphics[width=1\linewidth]{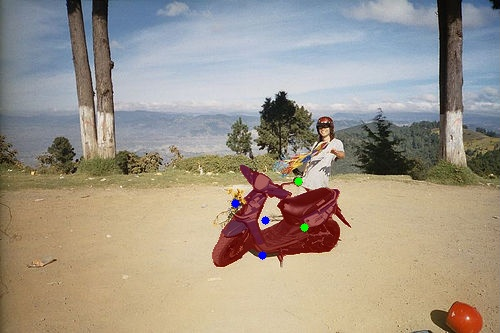}
    \subcaption*{5 clicks 90.9\%}
  \end{subfigure}

  \begin{subfigure}[t]{0.22\linewidth}
    \includegraphics[width=1\linewidth]{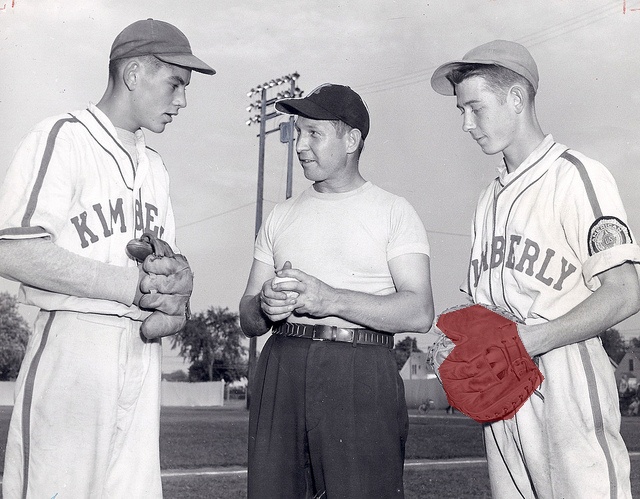}
    \subcaption*{GT}
  \end{subfigure}
  \hspace{2mm}
  \begin{subfigure}[t]{0.22\linewidth}
    \includegraphics[width=1\linewidth]{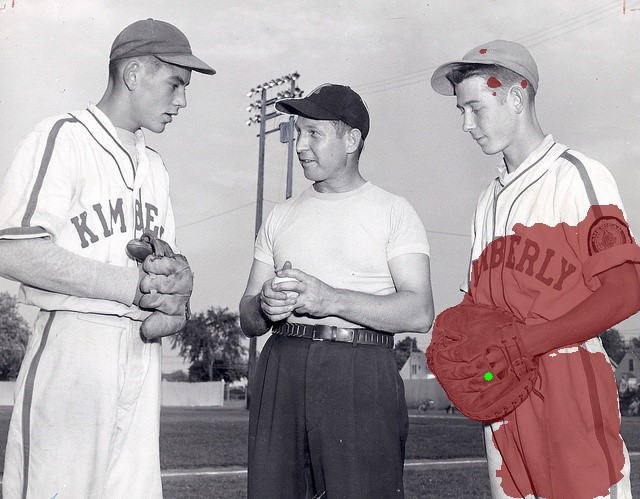}
    \subcaption*{1 click 20.2\%}
  \end{subfigure}
  \hspace{2mm}
  \begin{subfigure}[t]{0.22\linewidth}
    \includegraphics[width=1\linewidth]{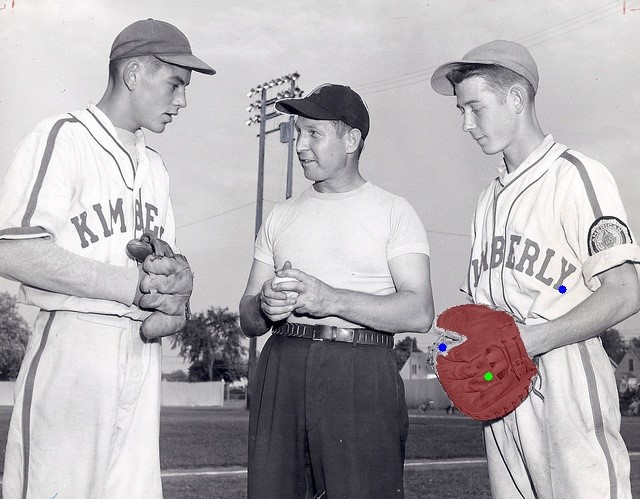}
    \subcaption*{3 clicks 88.0\%}
  \end{subfigure}
  \hspace{2mm}
  \begin{subfigure}[t]{0.22\linewidth}
    \includegraphics[width=1\linewidth]{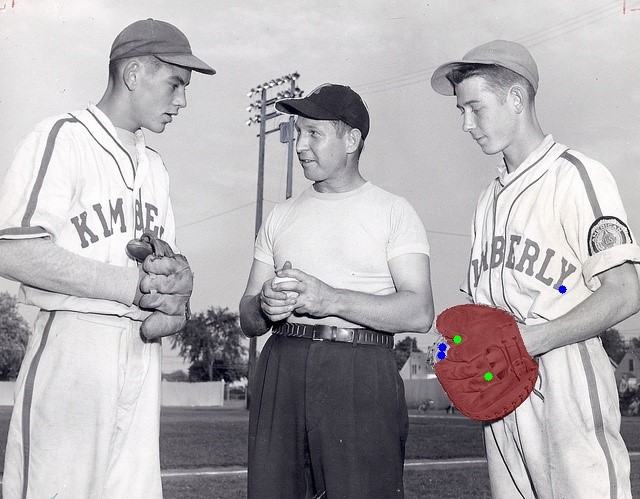}
    \subcaption*{5 clicks 90.5\%}
  \end{subfigure}
  \vspace{-2mm}
  \caption{Qualitative segmentation results from GrabCut \citep{rother2004grabcut} (Row 1), Berkeley \citep{mcguinness2010berkeley} (Row 2), SBD \citep{hariharan2011sbd} (Row 3-4), and COCO MVal (Row 5). The performance is reported with a certain number of clicks. Green and blue dots denote positive and negative clicks, respectively. The predictions and ground truths are overlaid on the original images.}
  \vspace{-2mm}
  \label{fig:segmentations single}
\end{figure*}

\subsection{Ablation Study}
\label{sec:ablation_study}
To demonstrate the effectiveness of each component of our model, we conducted several groups of ablation studies for MOIS.
We followed the settings discussed in Sec.~\ref{sec: imple details}, except for the variables adjusted for each set of ablation studies.

\begin{table}
\caption{Ablation study of each module of our iCMFormer++ model. All models were trained and evaluated on COCO MOIS. We abbreviate two-stream, cross-attention, self-attention fusion, channel fusion, and \textit{exemplar} informed module as TS, CA, SF, CF, and EIM, respectively.}
\centering
\setlength\tabcolsep{2pt}
\begin{tabular}{ccccccccc} 
\toprule
\multirow{2}*{TS} & \multirow{2}*{CA} & \multirow{2}*{SF} & \multirow{2}*{CF} & \multirow{2}*{EIM} & \multicolumn{4}{c}{COCO MOIS}\\
& & & & & NoC85$\downarrow$ & NoC90$\downarrow$ & mIoU$\circledast$5+$\uparrow$ & mIoU$\circledast$10+$\uparrow$ \\
\hline
\checkmark & & & &  & 12.84 & 16.19  & 75.27\% & 80.49\%\\
\checkmark &  & \checkmark & & & 12.51 & 15.92 & 75.95\% & 80.92\%\\
\checkmark & \checkmark & \checkmark & & & 12.14 & 15.61 & 76.89\% & 81.66\%\\
\checkmark & \checkmark & & & \checkmark & 9.06 & 12.74 & 81.89\% & 85.12\%\\
\checkmark & \checkmark & & \checkmark & \checkmark & \textbf{8.47} & \textbf{11.92} & \textbf{82.47\%} & \textbf{85.89\%}\\
\bottomrule
\end{tabular}
\vspace{-2mm}
\label{tab: ablation study on modules}
\end{table}

\subsubsection{Effectiveness of Proposed Modules}
\label{sec: ab modules}
We quantitatively evaluated the impact of each module proposed in the  iCMFormer++ model, \ie, the two-stream branches to separately handle the learning of the \textit{exemplar} and \textit{recall} objects, the cross-attention denoted in Eq.~\eqref{eq:cross-a}, the self-attention fusion in iCMFormer, the channel fusion in the feature fusion module (Sec. \ref{sec:ff module}), and the \textit{exemplar}-informed module (EIM) (Sec. \ref{sec:case module}).
The results are summarized in Table~\ref{tab: ablation study on modules}.
To establish a baseline for a progressive comparison, we used the two-stream architecture modified on the previous iCMFormer model without cross-attention blocks.
As indicated in the first row of Table~\ref{tab: ablation study on modules}, the two-stream baseline even provided with satisfactory masks and interactions could only reach a performance comparable to that of the models in Table~\ref{tab:comparison with multiple SOTA} without using the category information.
The third row shows the iCMFormer model with the cross-attentions and a self-attention fusion, which only obtained a marginal performance improvement over the baseline.
This indicates that building directly on our previous SOIS method for MOIS fails to effectively capture the guidance from the satisfactory \textit{exemplar}.
Hence, in comparison, we first added the proposed EIM to measure the similarity between the \textit{exemplar} and image regions for the potential focus.
Subsequently, we replaced the self-attention blocks with a simpler channel fusion module, which aims to focus on the channel-wise cross-attended features and reduce the redundant self-calculations before the segmentation head.
As shown in the last row, this method results in a significant click reduction, \eg, requiring 3.67 fewer clicks to reach 85\% IoU, and 4.23\% segmentation boost with ten additional clicks. These performance gains evidently highlight the effectiveness of each proposed module in reducing interaction efforts and enhancing segmentation performance.

\subsubsection{Effectiveness of Utilized Input Data}
\label{sec: ab input data}
In this section, we report the effects of using different input data for the \textit{exemplar} branch in Table~\ref{tab: ablation study on input}.
By comparing the first three rows in this table, we found that both interactions and masks are important for learning from the \textit{exemplars}.
We further evaluated the impact of the background information for MOIS.
To achieve that, we first investigated the impact of a masked image compared to the original image for the \textit{exemplar} branch. Specifically, we adopted the same bounding box extraction as that in the proposed EIM (in Sec.~\ref{sec:case module}) but only kept the biggest scale for buffers.
Then the bounding box was leveraged to obtain a masked image by overlapping the outside.
The result in row 4 shows that the masked images did not improve the performance of interaction efforts.
In addition, we analyzed the influence of the negative clicks in the \textit{exemplar} branch by comparing the results with and without them (row 3 vs. row 5).
Including the negative clicks resulted in a 0.45\% and 0.96\% mIoU gain, respectively, with five and ten additional clicks.
Based on these gains, we conclude that the background information contributes to the performance improvement of MOIS.

\begin{table}
\caption{Ablation study of different usages of the input image, satisfactory mask, and previous interactions for multiple object interactive segmentation. M denotes the masked image for the visual input of the \textit{exemplar} branch.}
\centering
\setlength\tabcolsep{2pt}
\begin{tabular}{cccccccc}
\toprule
\multirow{2}*{Image} & \multirow{2}*{Mask} & \multirow{2}*{PosC} & \multirow{2}*{NegC} & \multicolumn{4}{c}{COCO MOIS}\\
& & & & NoC85$\downarrow$ & NoC90$\downarrow$ & mIoU$\circledast$5+$\uparrow$ & mIoU$\circledast$10+$\uparrow$ \\
\hline
\checkmark & & \checkmark & \checkmark & 12.62 & 16.03 & 75.71\% & 80.74\% \\
\checkmark & \checkmark & & & 12.08 & 15.59 & 75.37\% & 80.81\%\\
\checkmark & \checkmark & \checkmark &  & 8.94 & 12.26 & 82.02\% & 84.93\%\\
M &\checkmark & \checkmark & \checkmark & 9.42 & 13.55 & 81.74\% & 84.67\% \\
\checkmark & \checkmark & \checkmark & \checkmark & \textbf{8.47} & \textbf{11.92} & \textbf{82.47\%} & \textbf{85.89\%}\\
\bottomrule
\end{tabular}
\label{tab: ablation study on input}
\end{table}

\subsubsection{Influence of Various Clicks}
\label{sec: ab various click}
In addition to the input data, we further analyzed various clicks for representing users' guidance in the \textit{exemplar} branch.
The results are reported in Table~\ref{tab: ablation study on clicks}.
For positive clicks, we provided two variants: (a) sparse representation with only the first click; and (b) complete representation with all positive clicks.
By comparing the first two rows in this table, it is evident that applying all positive clicks for the \textit{exemplar} branch resulted in fewer clicks and improved segmentation performance.
We speculate that a greater number of positive clicks indicates more accurately focused image regions.
Regarding negative clicks, as discussed in Sec.~\ref{sec:training_scheme}, we further distinguish by separating them into \textit{Pseudo Negative Clicks} and \textit{True Negative Clicks} for MOIS.
We hypothesize that {Pseudo Negative Clicks} from other objects could mislead the model when distinguishing between a target and background.
The result shown in the last row demonstrates the effectiveness of distinguishing {True Negative Clicks}, which leads to a segmentation performance improvement and a reduction in the required number of additional clicks.

\begin{table}
\caption{Ablation study of different clicks in the \textit{exemplar} branch for multiple object interactive segmentation. We abbreviate the first positive click, all positive clicks, pseudo negative clicks, and true negative clicks as FPC, APC, PNC, and TNC, respectively.}
\centering
\setlength\tabcolsep{2pt}
\begin{tabular}{cccccccc}
\toprule
\multicolumn{2}{c}{PositiveC} & \multicolumn{2}{c}{NegativeC}  & \multicolumn{4}{c}{COCO MOIS}\\
FPC & APC & PNC & TNC & NoC85$\downarrow$ & NoC90$\downarrow$ & mIoU$\circledast$5+$\uparrow$ & mIoU$\circledast$10+$\uparrow$ \\
\hline
\checkmark & & \checkmark &  & 9.85 & 13.97 & 81.62\% & 84.64\% \\
 & \checkmark & \checkmark & & 8.76 & 12.03 & 82.09\% & 84.99\%\\
 & \checkmark &  & \checkmark & \textbf{8.47} & \textbf{11.92} & \textbf{82.47\%} & \textbf{85.89\%} \\
\bottomrule
\end{tabular}
\vspace{-2mm}
\label{tab: ablation study on clicks}
\end{table}

\begin{figure*}[http]
  \centering
  \begin{subfigure}[t]{0.18\linewidth}
    \includegraphics[width=1\linewidth]{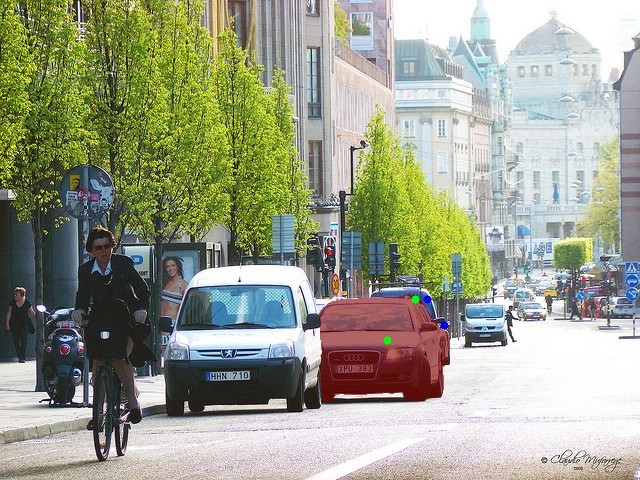}
    \subcaption*{Exemplar 86.9\%}
  \end{subfigure}
  \hspace{2mm}
  \begin{subfigure}[t]{0.18\linewidth}
    \includegraphics[width=1\linewidth]{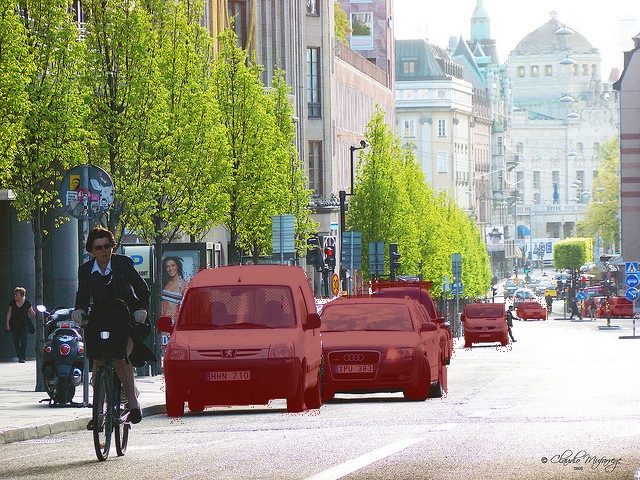}
    \subcaption*{GT}
  \end{subfigure}
  \hspace{2mm}
  \begin{subfigure}[t]{0.18\linewidth}
    \includegraphics[width=1\linewidth]{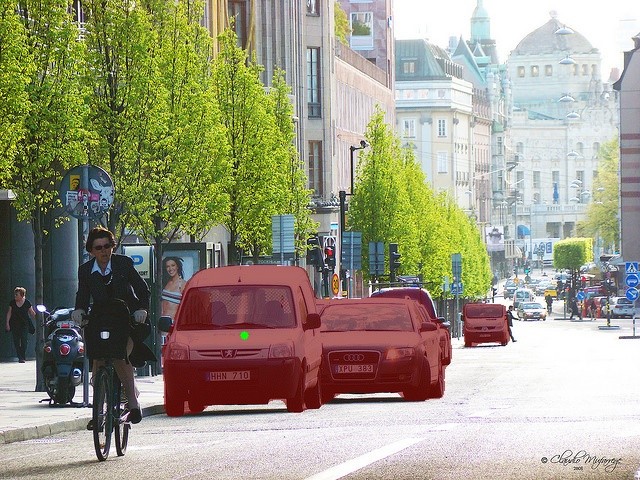}
    \subcaption*{1 click 87.7\%}
  \end{subfigure}
  \hspace{2mm}
  \begin{subfigure}[t]{0.18\linewidth}
    \includegraphics[width=1\linewidth]{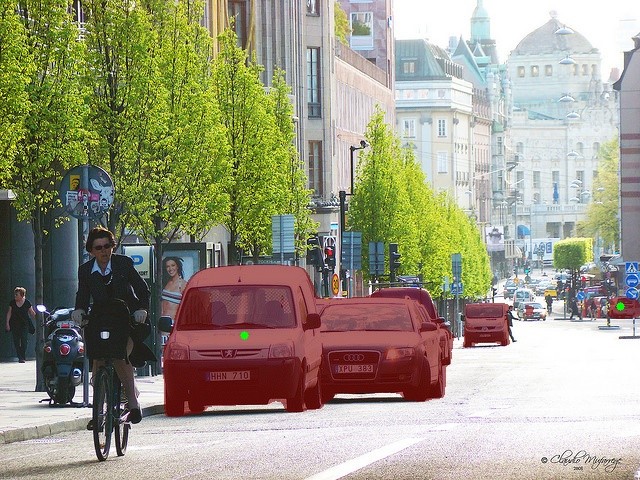}
    \subcaption*{2 clicks 89.3\%}
  \end{subfigure}
  \hspace{2mm}
  \begin{subfigure}[t]{0.18\linewidth}
    \includegraphics[width=1\linewidth]{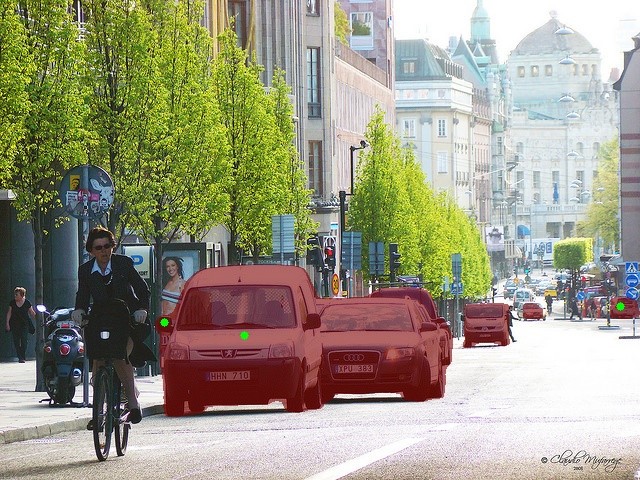}
    \subcaption*{3 clicks 90.1\%}
  \end{subfigure}

  \begin{subfigure}[t]{0.18\linewidth}
    \includegraphics[width=1\linewidth]{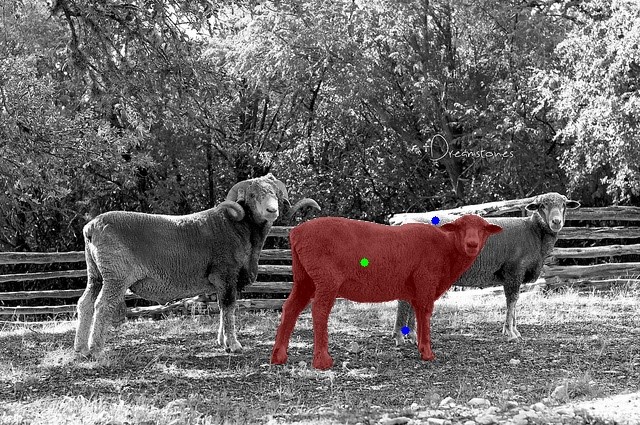}
    \subcaption*{Exemplar 89.4\%}
  \end{subfigure}
  \hspace{2mm}
  \begin{subfigure}[t]{0.18\linewidth}
    \includegraphics[width=1\linewidth]{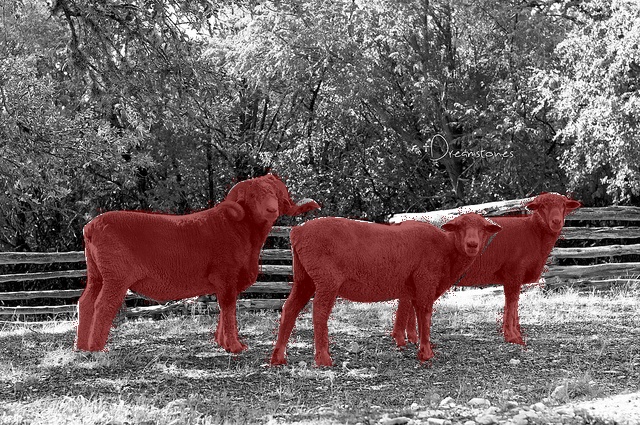}
    \subcaption*{GT}
  \end{subfigure}
  \hspace{2mm}
  \begin{subfigure}[t]{0.18\linewidth}
    \includegraphics[width=1\linewidth]{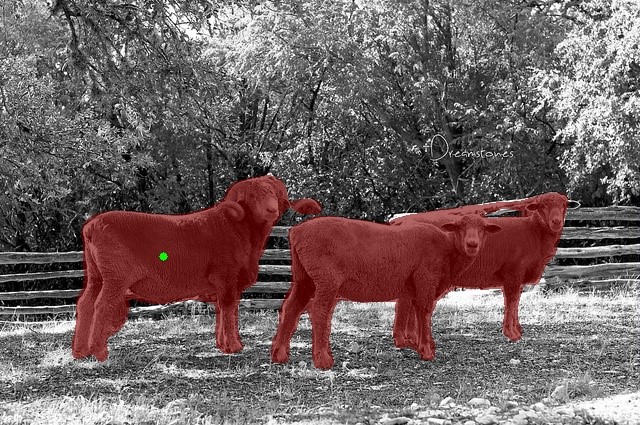}
    \subcaption*{1 click 87.0\%}
  \end{subfigure}
  \hspace{2mm}
  \begin{subfigure}[t]{0.18\linewidth}
    \includegraphics[width=1\linewidth]{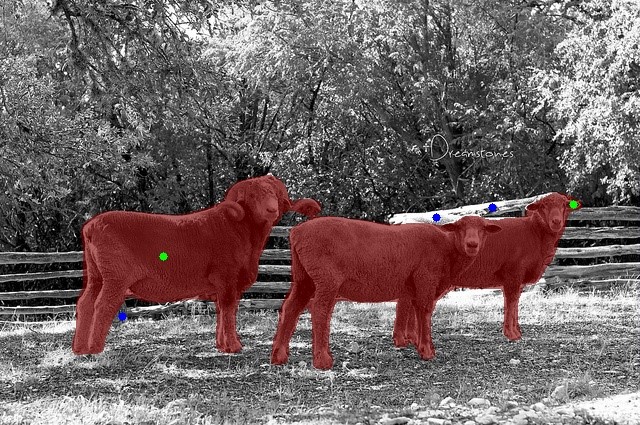}
    \subcaption*{5 clicks 89.2\%}
  \end{subfigure}
  \hspace{2mm}
  \begin{subfigure}[t]{0.18\linewidth}
    \includegraphics[width=1\linewidth]{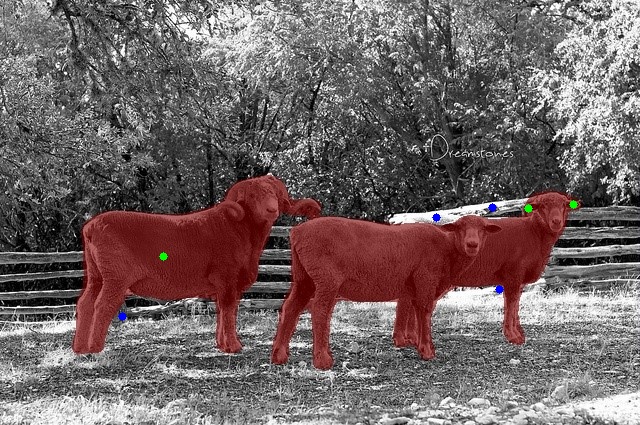}
    \subcaption*{7 clicks 90.1\%}
  \end{subfigure}

  \begin{subfigure}[t]{0.18\linewidth}
    \includegraphics[width=1\linewidth]{figures/qualitative_multiple/1136_4_873.jpg}
    \subcaption*{Exemplar 87.3\%}
  \end{subfigure}
  \hspace{2mm}
  \begin{subfigure}[t]{0.18\linewidth}
    \includegraphics[width=1\linewidth]{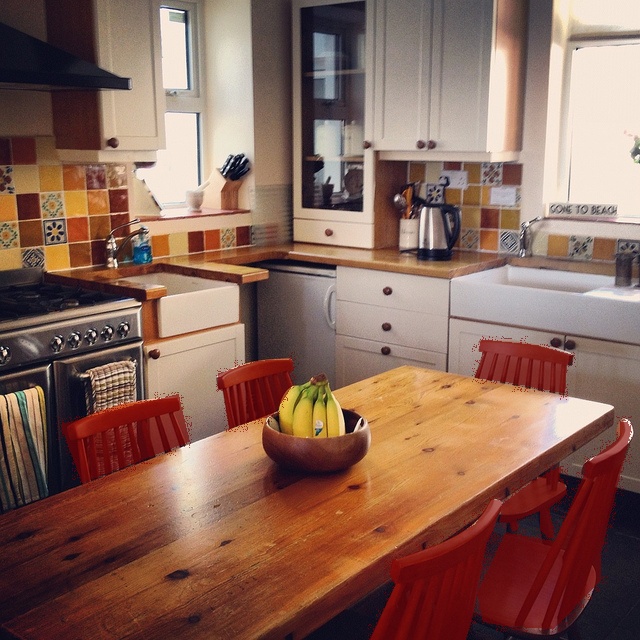}
    \subcaption*{GT}
  \end{subfigure}
  \hspace{2mm}
  \begin{subfigure}[t]{0.18\linewidth}
    \includegraphics[width=1\linewidth]{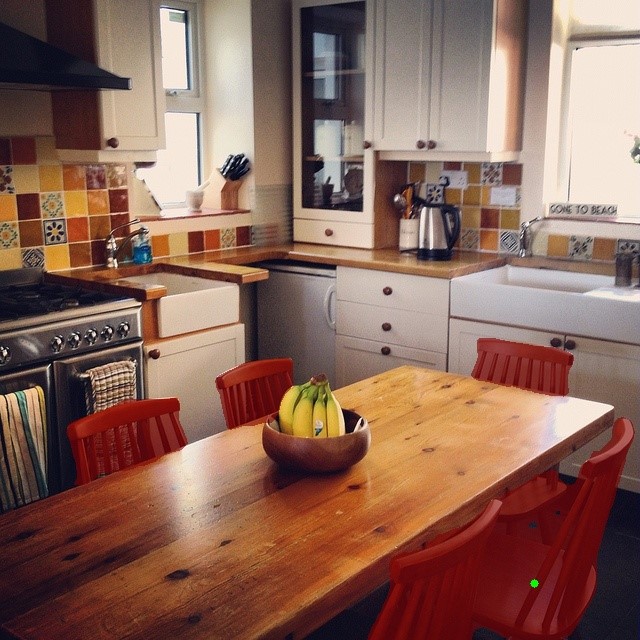}
    \subcaption*{1 click 84.2\%}
  \end{subfigure}
  \hspace{2mm}
  \begin{subfigure}[t]{0.18\linewidth}
    \includegraphics[width=1\linewidth]{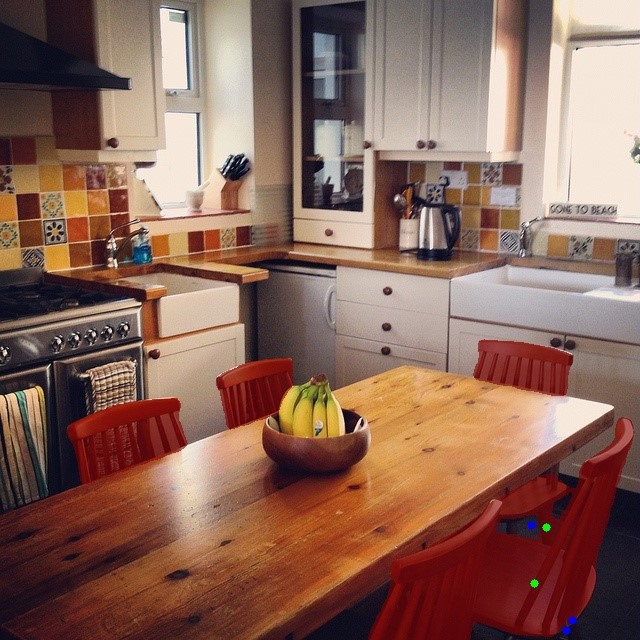}
    \subcaption*{5 clicks 86.3\%}
  \end{subfigure}
  \hspace{2mm}
  \begin{subfigure}[t]{0.18\linewidth}
    \includegraphics[width=1\linewidth]{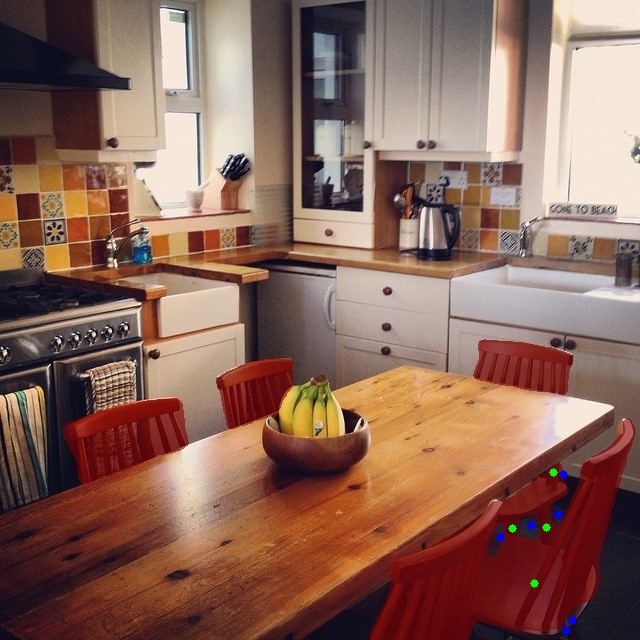}
    \subcaption*{10 clicks 90.0\%}
  \end{subfigure}

  \begin{subfigure}[t]{0.18\linewidth}
    \includegraphics[width=1\linewidth]{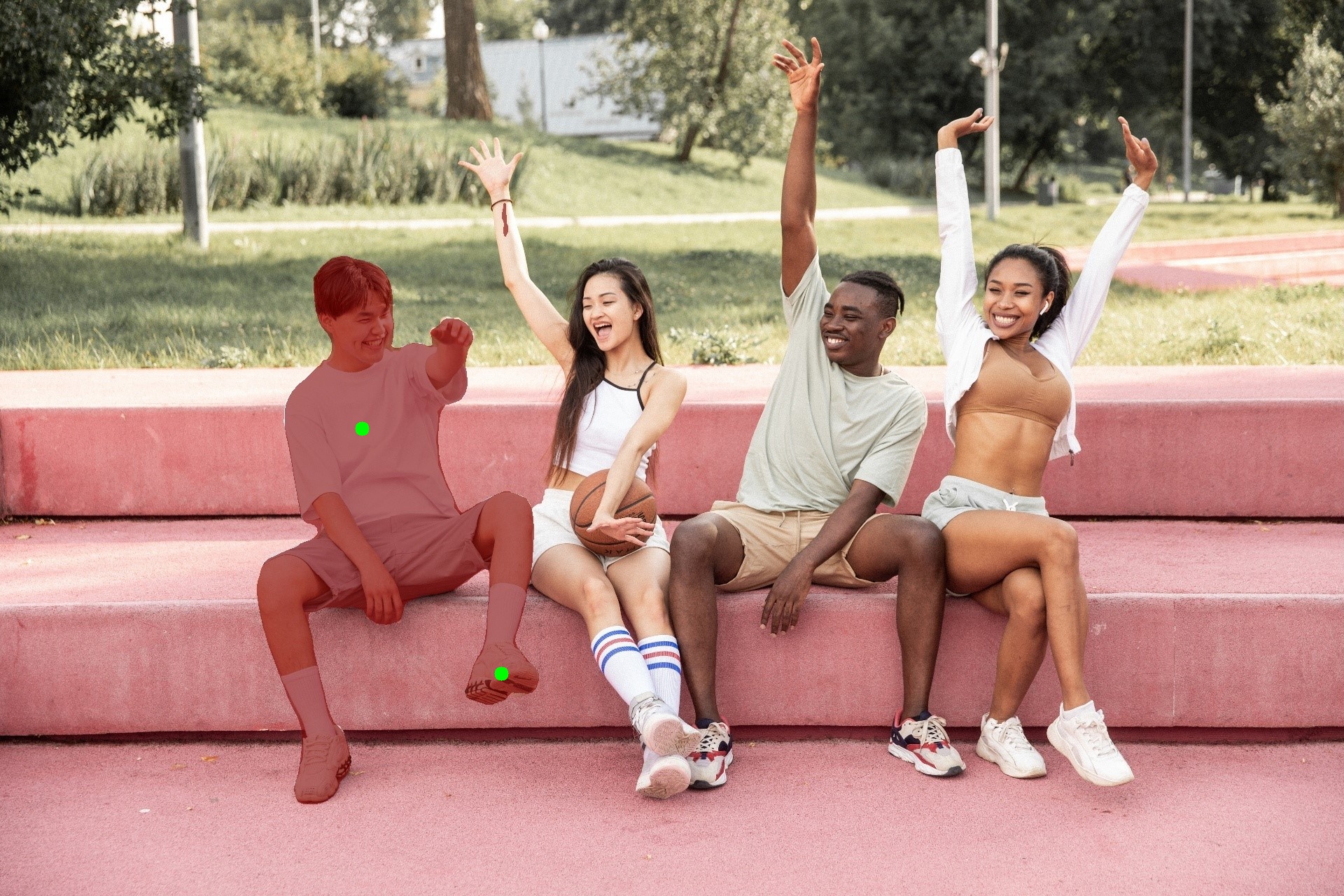}
    \subcaption*{Exemplar 93.0\%}
  \end{subfigure}
  \hspace{2mm}
  \begin{subfigure}[t]{0.18\linewidth}
    \includegraphics[width=1\linewidth]{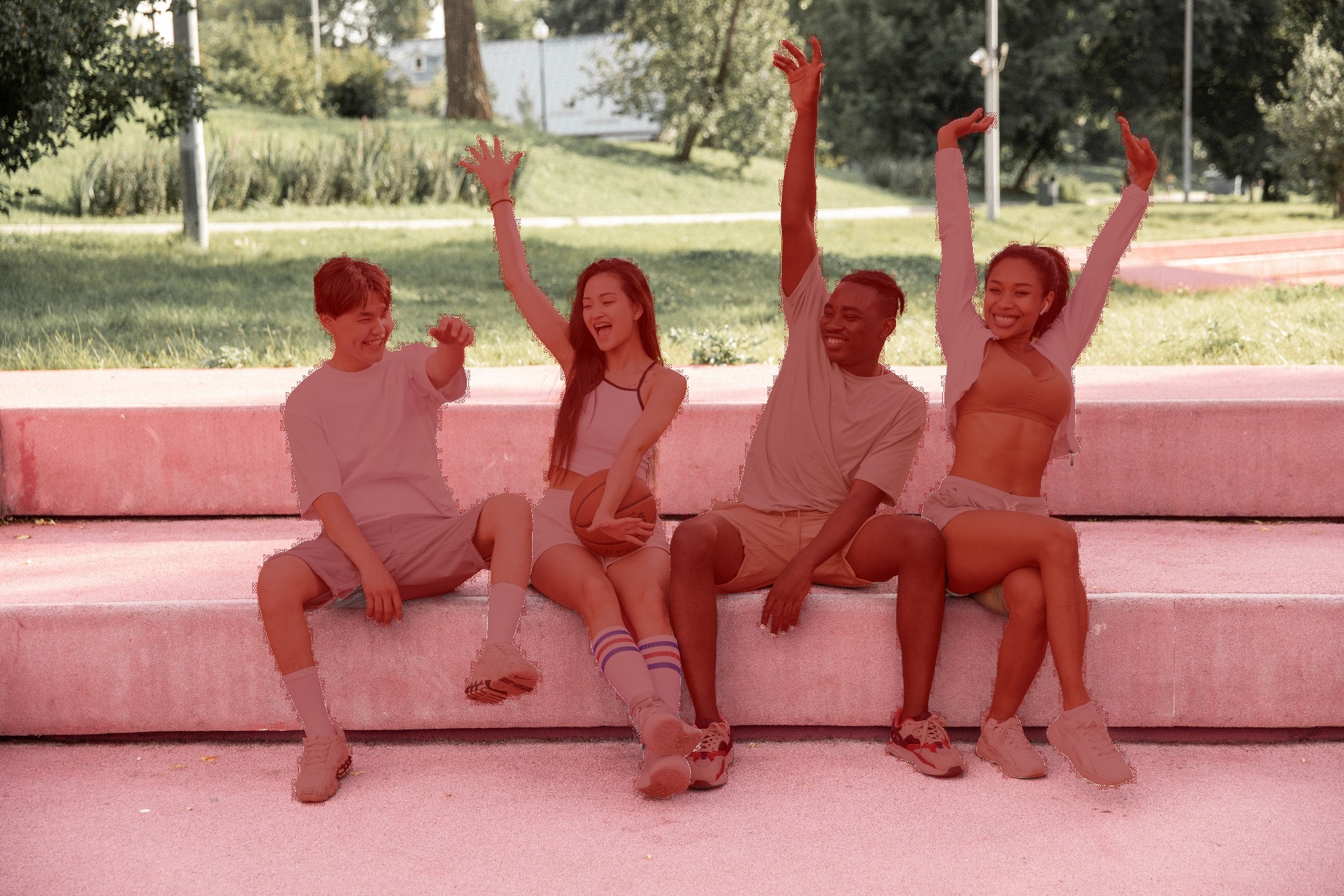}
    \subcaption*{GT}
  \end{subfigure}
  \hspace{2mm}
  \begin{subfigure}[t]{0.18\linewidth}
    \includegraphics[width=1\linewidth]{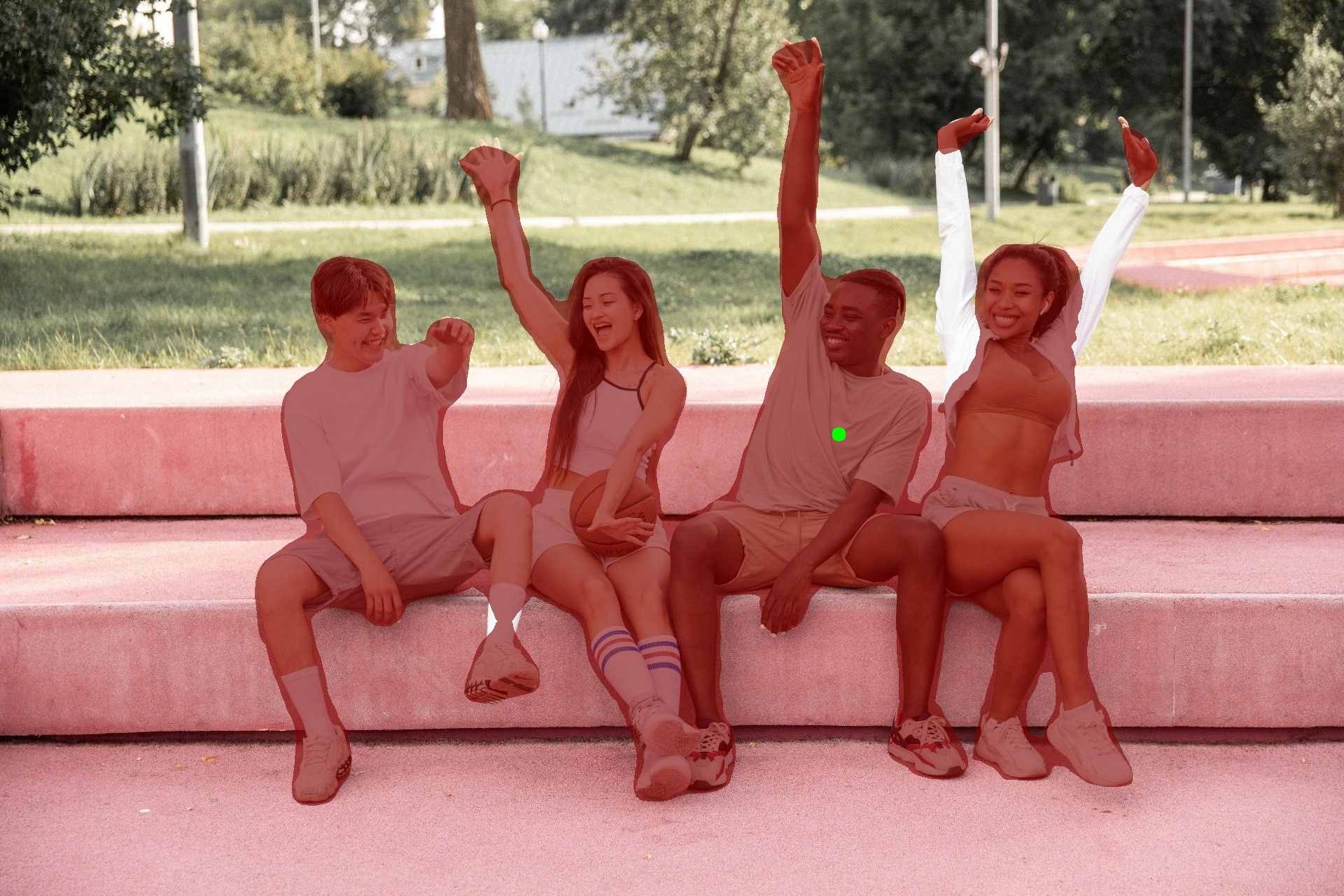}
    \subcaption*{1 click 85.4\%}
  \end{subfigure}
  \hspace{2mm}
  \begin{subfigure}[t]{0.18\linewidth}
    \includegraphics[width=1\linewidth]{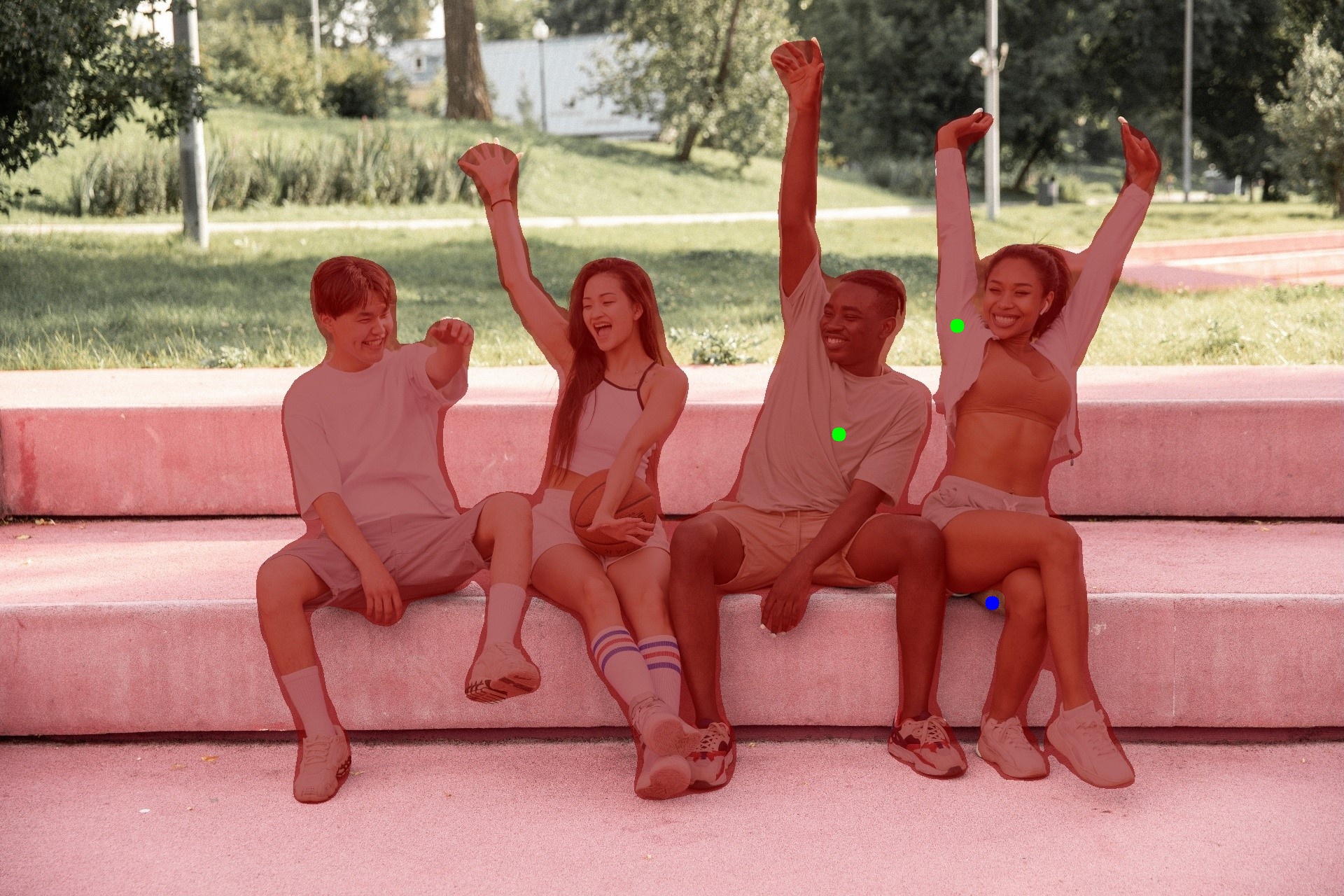}
    \subcaption*{3 clicks 89.7\%}
  \end{subfigure}
  \hspace{2mm}
  \begin{subfigure}[t]{0.18\linewidth}
    \includegraphics[width=1\linewidth]{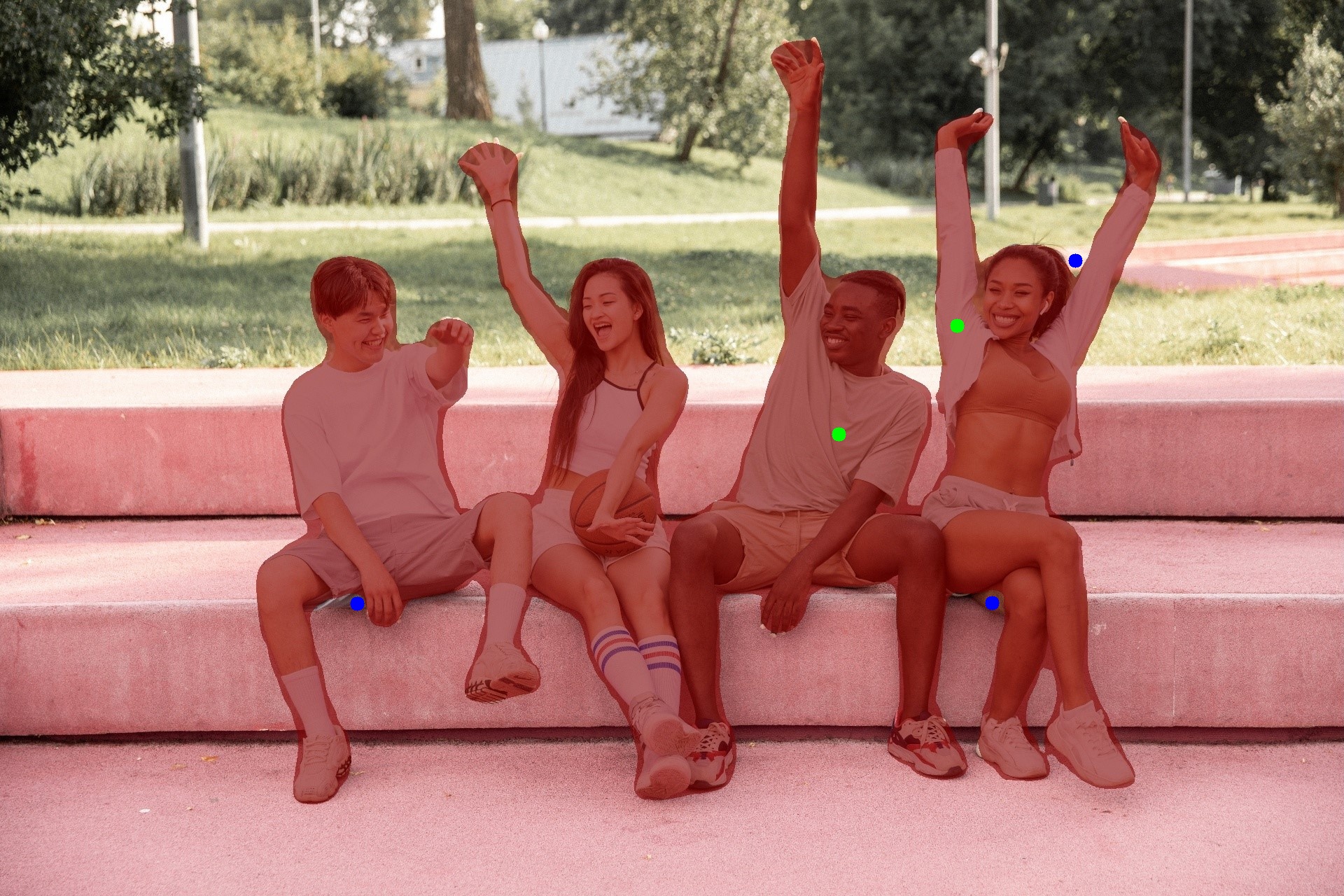}
    \subcaption*{5 clicks 90.1\%}
  \end{subfigure}
    
  \begin{subfigure}[t]{0.18\linewidth}
    \includegraphics[width=1\linewidth]{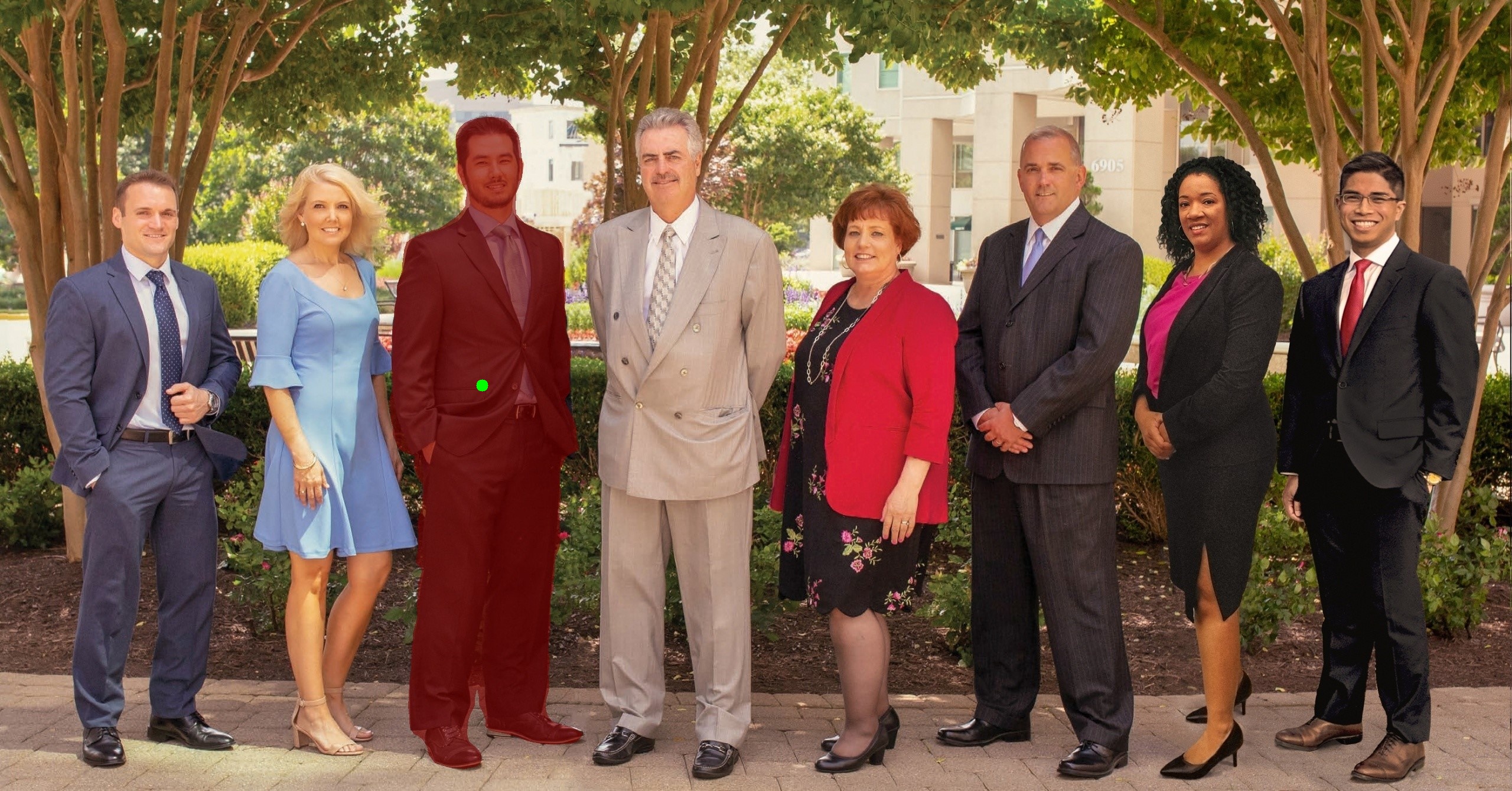}
    \subcaption*{Exemplar 94.3\%}
  \end{subfigure}
  \hspace{2mm}
  \begin{subfigure}[t]{0.18\linewidth}
    \includegraphics[width=1\linewidth]{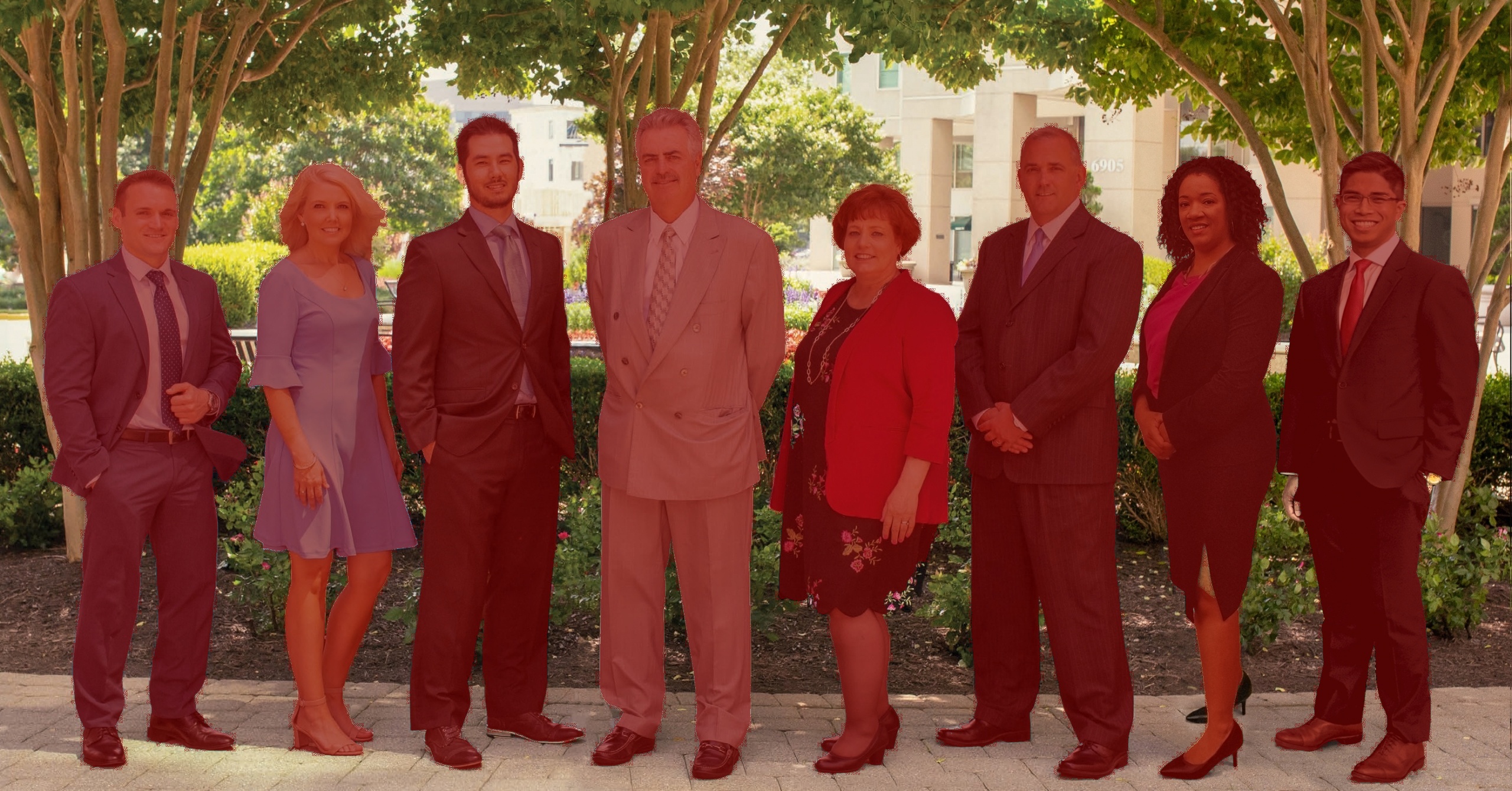}
    \subcaption*{GT}
  \end{subfigure}
  \hspace{2mm}
  \begin{subfigure}[t]{0.18\linewidth}
    \includegraphics[width=1\linewidth]{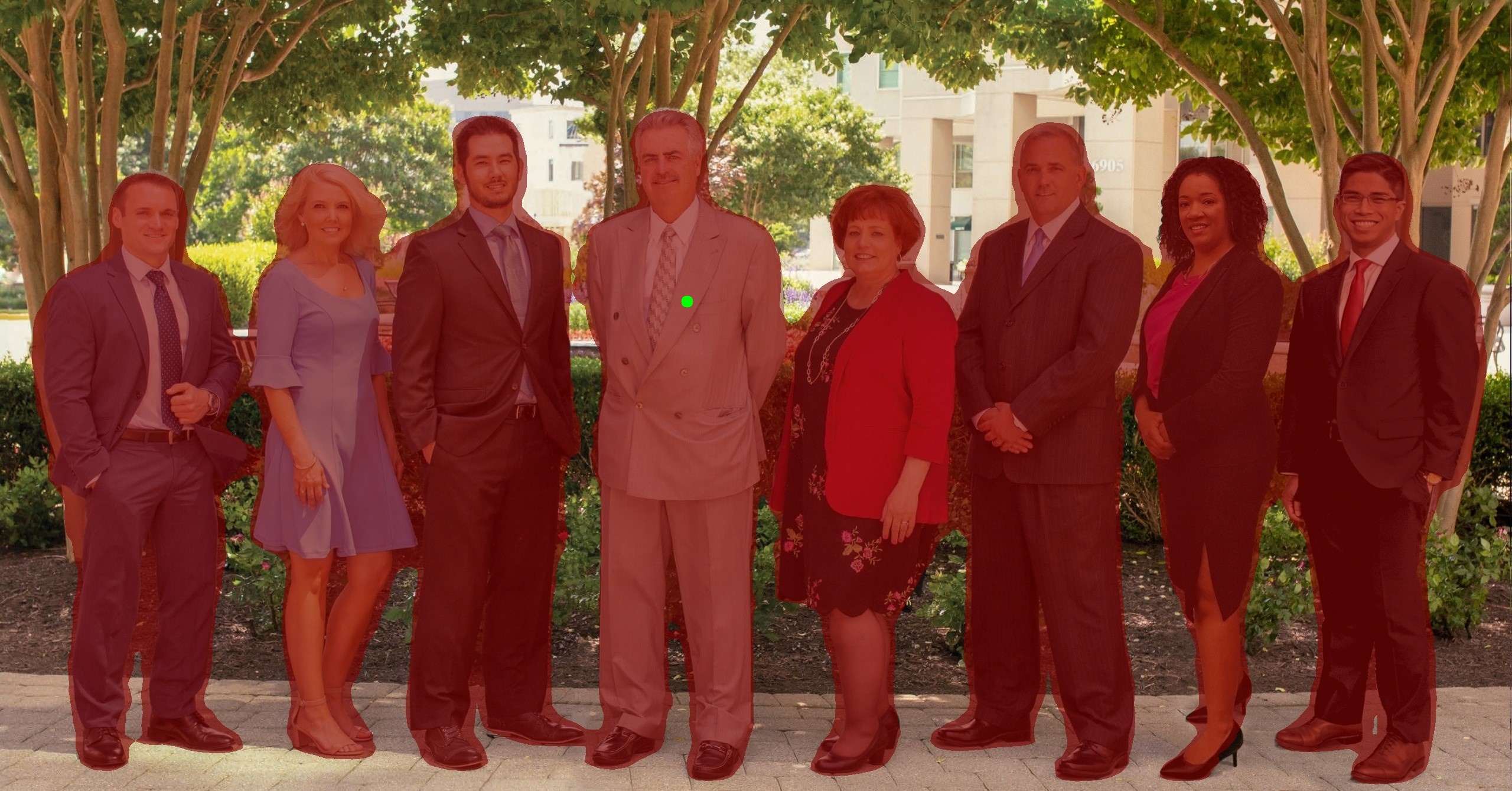}
    \subcaption*{1 click 87.8\%}
  \end{subfigure}
  \hspace{2mm}
  \begin{subfigure}[t]{0.18\linewidth}
    \includegraphics[width=1\linewidth]{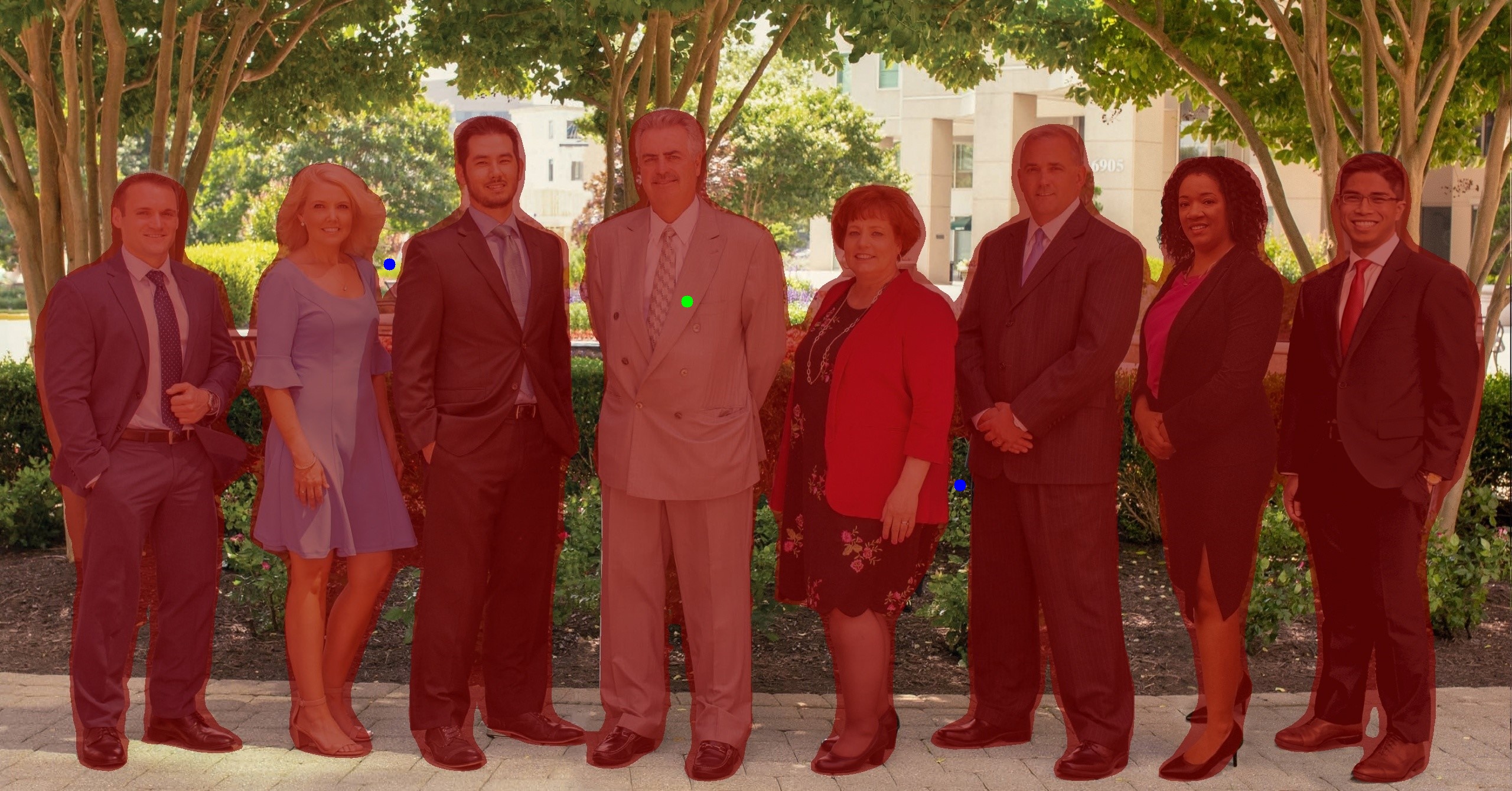}
    \subcaption*{3 clicks 89.2\%}
  \end{subfigure}
  \hspace{2mm}
  \begin{subfigure}[t]{0.18\linewidth}
    \includegraphics[width=1\linewidth]{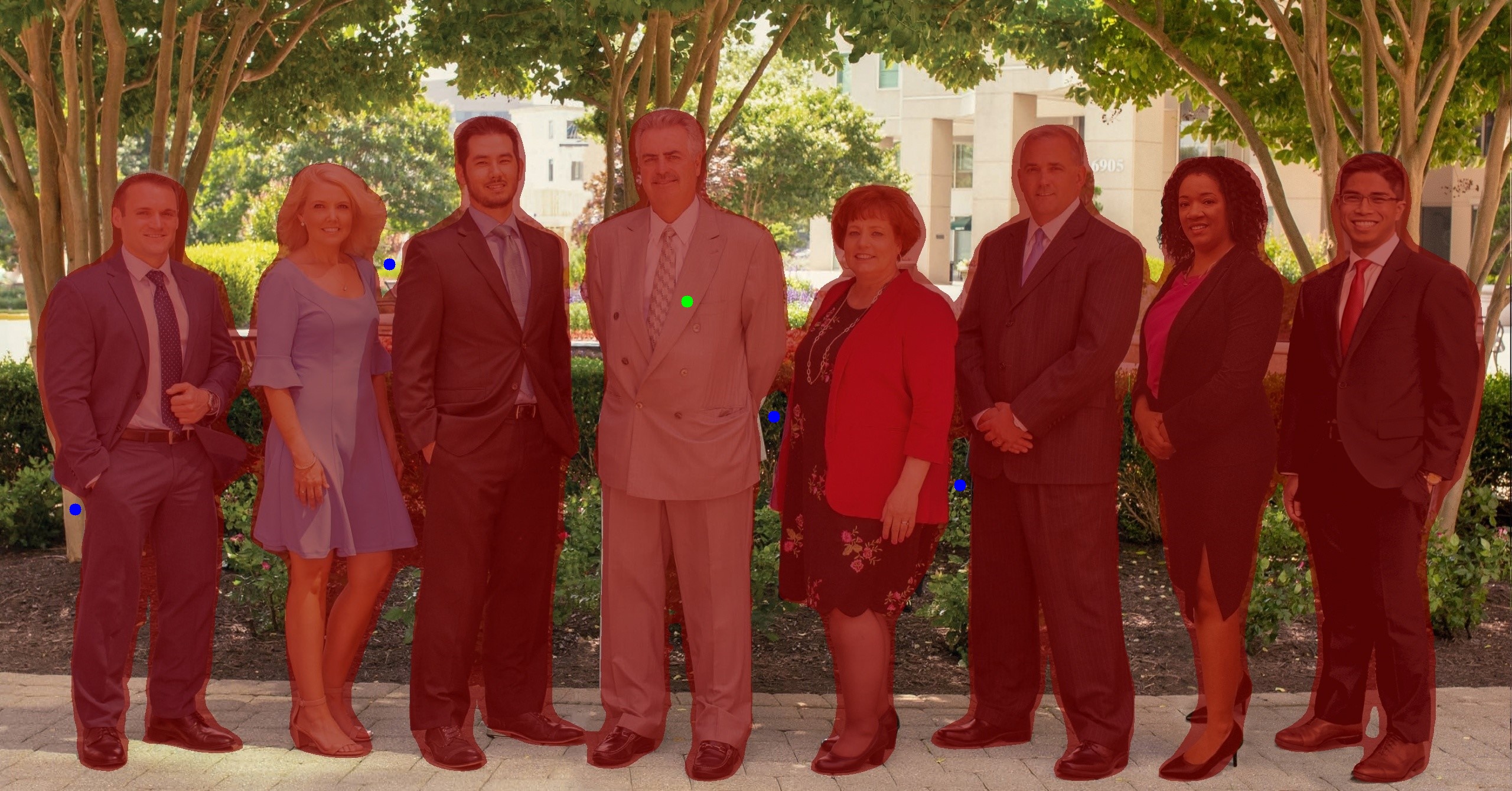}
    \subcaption*{5 clicks 89.7\%}
  \end{subfigure}
  \caption{Qualitative segmentation results from COCO MOIS (Row 1-3), and HIM2K (Row 4-5). The \textit{exemplar} shows the previously interacted object with a satisfactory mask (the IoU number reported at side). The last three columns progressively show the refinement of all objects with additional clicks.}
  \vspace{-2mm}
  \label{fig:segmentations multiple}
\end{figure*}

\begin{figure*}[ht]
  \centering
  \begin{subfigure}[t]{0.18\linewidth}
    \includegraphics[width=1\linewidth]{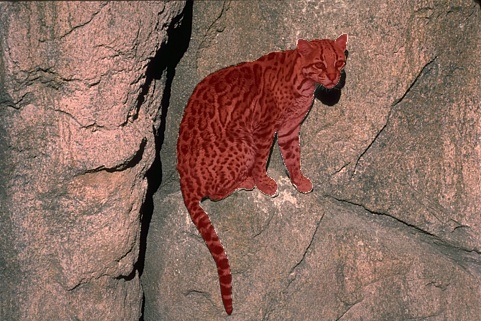}
    \subcaption*{GT}
  \end{subfigure}
  \hspace{2mm}
  \begin{subfigure}[t]{0.18\linewidth}
    \includegraphics[width=1\linewidth]{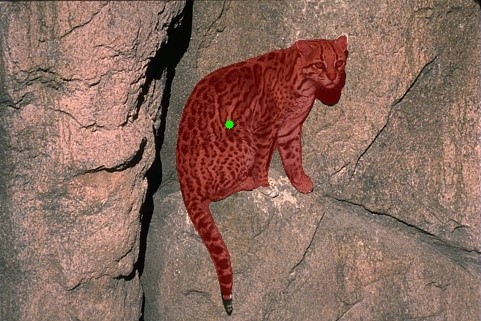}
    \subcaption*{1 click 88.0\%}
  \end{subfigure}
  \hspace{2mm}
  \begin{subfigure}[t]{0.18\linewidth}
    \includegraphics[width=1\linewidth]{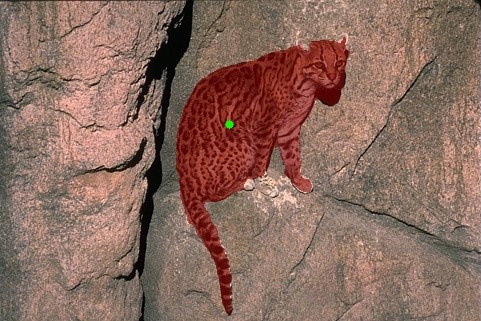}
    \subcaption*{1 click 88.3\%}
  \end{subfigure}
  \hspace{2mm}
  \begin{subfigure}[t]{0.18\linewidth}
    \includegraphics[width=1\linewidth]{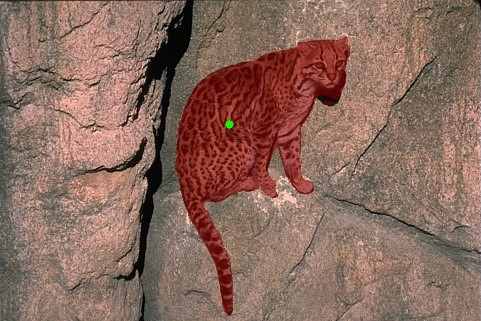}
    \subcaption*{1 click 89.6\%}
  \end{subfigure}
  \hspace{2mm}
  \begin{subfigure}[t]{0.18\linewidth}
    \includegraphics[width=1\linewidth]{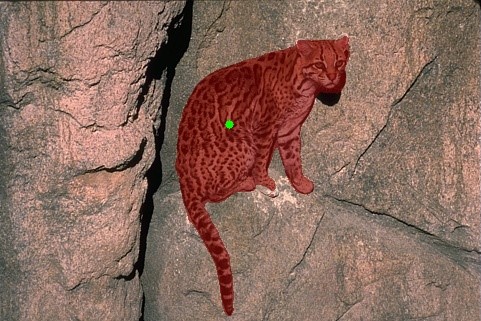}
    \subcaption*{1 click 89.5\%}
  \end{subfigure}

  \begin{subfigure}[t]{0.18\linewidth}
    \includegraphics[width=1\linewidth]{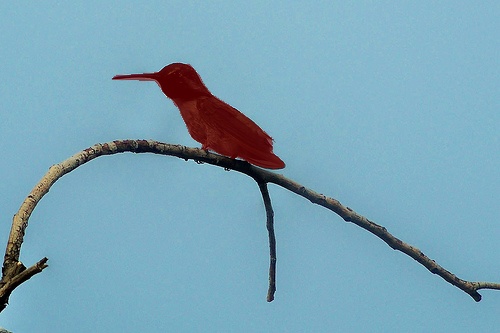}
    \subcaption*{GT}
  \end{subfigure}
  \hspace{2mm}
  \begin{subfigure}[t]{0.18\linewidth}
    \includegraphics[width=1\linewidth]{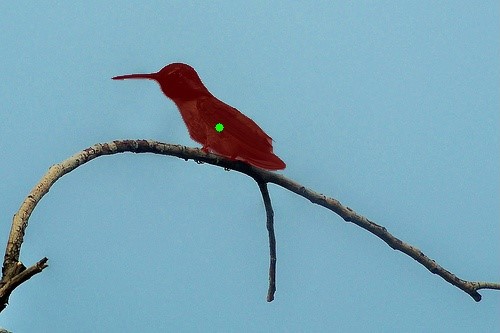}
    \subcaption*{1 click 93.7\%}
  \end{subfigure}
  \hspace{2mm}
  \begin{subfigure}[t]{0.18\linewidth}
    \includegraphics[width=1\linewidth]{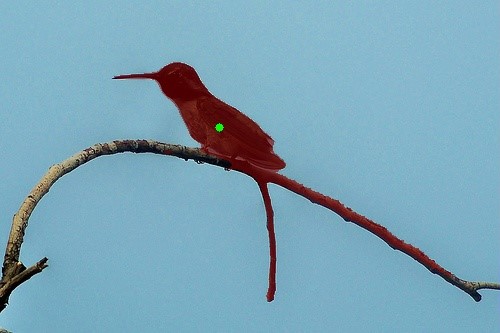}
    \subcaption*{1 click 63.6\%}
  \end{subfigure}
  \hspace{2mm}
  \begin{subfigure}[t]{0.18\linewidth}
    \includegraphics[width=1\linewidth]{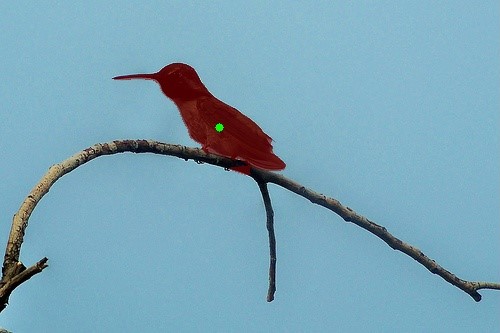}
    \subcaption*{1 click 91.4\%}
  \end{subfigure}
  \hspace{2mm}
  \begin{subfigure}[t]{0.18\linewidth}
    \includegraphics[width=1\linewidth]{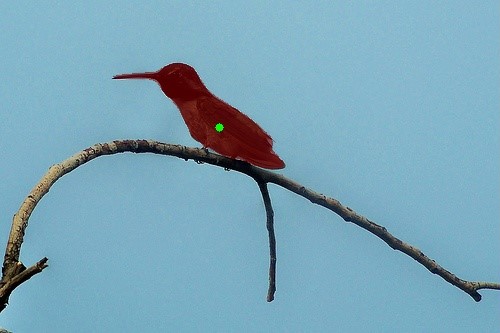}
    \subcaption*{1 click 95.1\%}
  \end{subfigure}
  
  \begin{subfigure}[t]{0.18\linewidth}
    \includegraphics[width=1\linewidth]{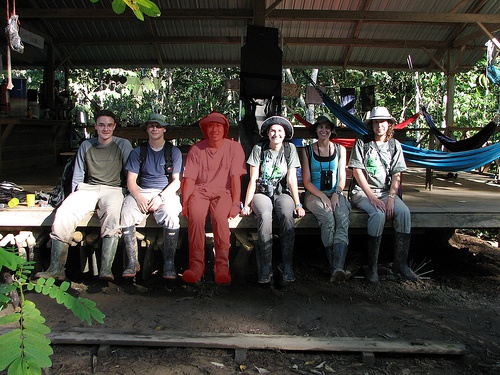}
    \subcaption*{GT}
  \end{subfigure}
  \hspace{2mm}
  \begin{subfigure}[t]{0.18\linewidth}
    \includegraphics[width=1\linewidth]{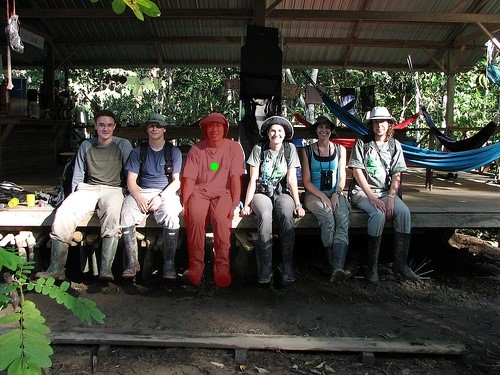}
    \subcaption*{1 click 90.0\%}
  \end{subfigure}
  \hspace{2mm}
  \begin{subfigure}[t]{0.18\linewidth}
    \includegraphics[width=1\linewidth]{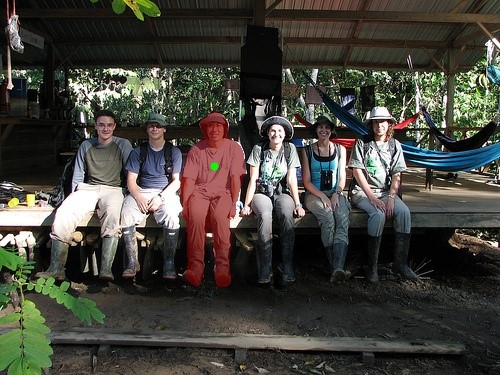}
    \subcaption*{1 click 90.0\%}
  \end{subfigure}
  \hspace{2mm}
  \begin{subfigure}[t]{0.18\linewidth}
    \includegraphics[width=1\linewidth]{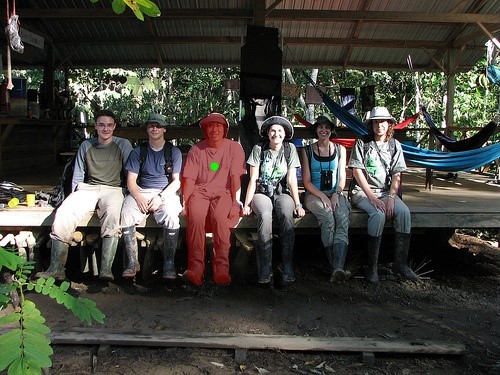}
    \subcaption*{1 click 88.1\%}
  \end{subfigure}
  \hspace{2mm}
  \begin{subfigure}[t]{0.18\linewidth}
    \includegraphics[width=1\linewidth]{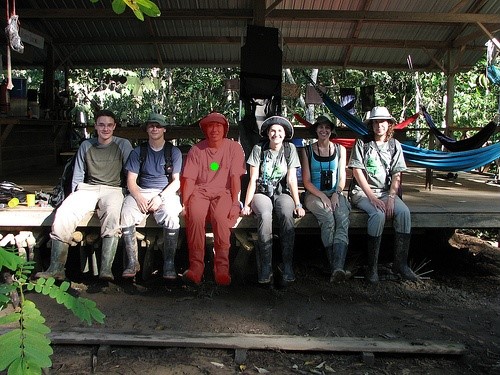}
    \subcaption*{1 click 90.7\%}
  \end{subfigure}

  \begin{subfigure}[t]{0.18\linewidth}
    \includegraphics[width=1\linewidth]{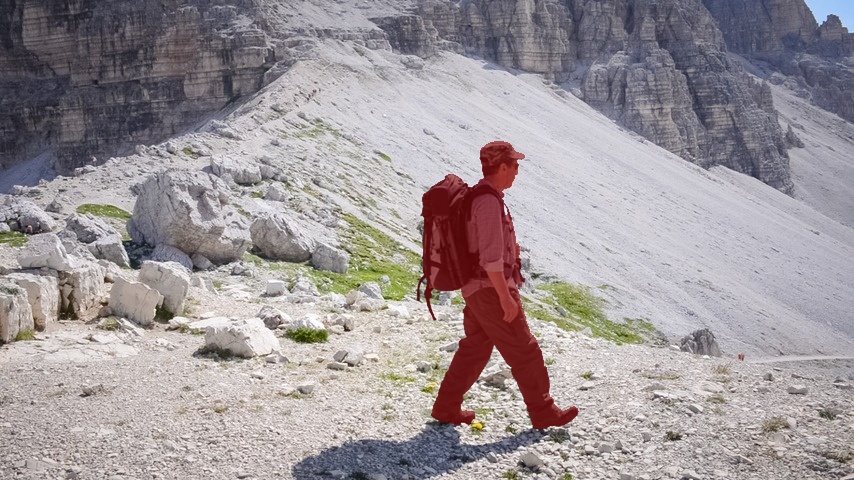}
    \subcaption*{GT}
  \end{subfigure}
  \hspace{2mm}
  \begin{subfigure}[t]{0.18\linewidth}
    \includegraphics[width=1\linewidth]{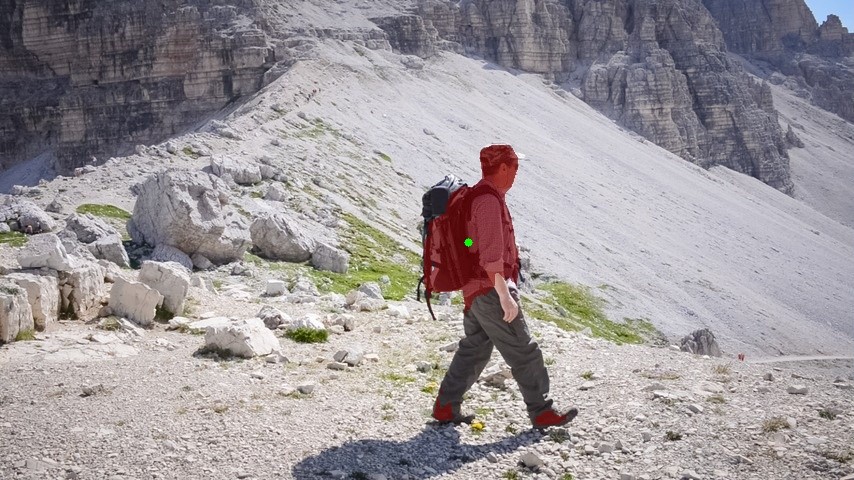}
    \subcaption*{1 click 49.8\%}
  \end{subfigure}
  \hspace{2mm}
  \begin{subfigure}[t]{0.18\linewidth}
    \includegraphics[width=1\linewidth]{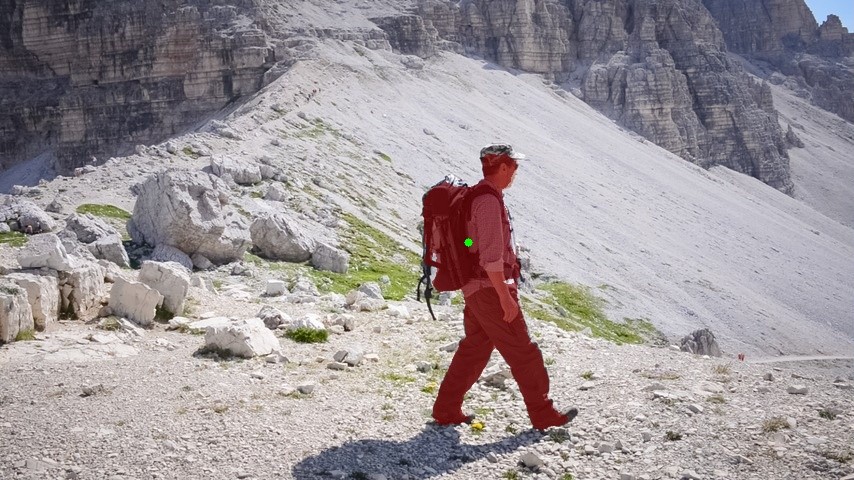}
    \subcaption*{1 click 87.2\%}
  \end{subfigure}
  \hspace{2mm}
  \begin{subfigure}[t]{0.18\linewidth}
    \includegraphics[width=1\linewidth]{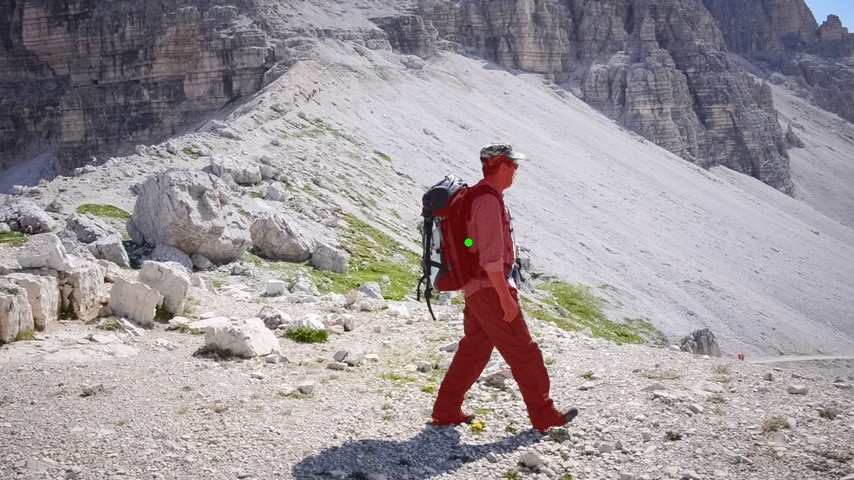}
    \subcaption*{1 click 77.7\%}
  \end{subfigure}
  \hspace{2mm}
  \begin{subfigure}[t]{0.18\linewidth}
    \includegraphics[width=1\linewidth]{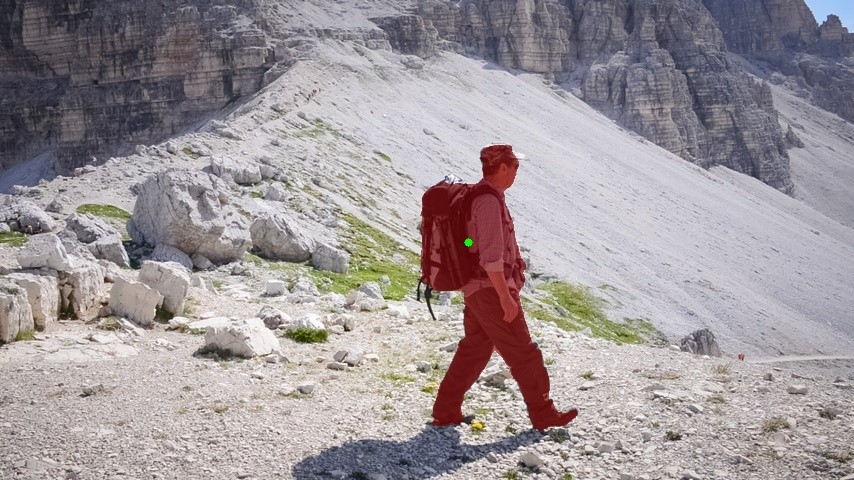}
    \subcaption*{1 click 91.8\%}
  \end{subfigure}

  \begin{subfigure}[t]{0.18\linewidth}
    \vspace{-2mm}
    \subcaption*{GT with mask}
  \end{subfigure}
  \hspace{2mm}
  \begin{subfigure}[t]{0.18\linewidth}
  \vspace{-2mm}
    \subcaption*{RITM}
  \end{subfigure}
  \hspace{2mm}
  \begin{subfigure}[t]{0.18\linewidth}
  \vspace{-2mm}
    \subcaption*{FocalClick}
  \end{subfigure}
  \hspace{2mm}
  \begin{subfigure}[t]{0.18\linewidth}
  \vspace{-2mm}
    \subcaption*{SimpleClick}
  \end{subfigure}
  \hspace{2mm}
  \begin{subfigure}[t]{0.18\linewidth}
  \vspace{-2mm}
    \subcaption*{iCMFormer (Ours)}
  \end{subfigure}

  \vspace{-2mm}
  \caption{Qualitative results compared to RITM \citep{sofiiuk2022ritm}, FocalClick \citep{chen2022focalclick}, and SimpleClick \citep{liu2023simpleclick}, within one positive click from the four SOIS datasets.}
  \vspace{-4mm}
  \label{fig:comparison with single}
\end{figure*}

\subsubsection{Influence of Exemplars for Inference}
\label{sec: ab exemplar}
As discussed in Sec.~\ref{sec:mois}, we collected the \textit{exemplars} for each image based on the ground truths from COCO \citep{lin2014coco}, and the satisfactory masks and interactions were converted and simulated automatically.
However, the simulated interactions are not always consistent with users' preferences.
To figure out the impact of different \textit{exemplars}, we conducted an ablation study on HIM2K \citep{sun2022humanvmultiple} and report the results in Table~\ref{tab: ablation study on exemplar}.
We call the simulated clicks from objects' ground truth masks \textit{Simulated-Ints}, adhering to the training settings.
There is a variant \textit{Automatic-Ints} that refers to the collection of predicted masks and sampled clicks generated from the automatic evaluation using our iCMFormer model.
In addition, we manually added clicks to correct the predictions of \textit{exemplar objects} using the iCMFormer model, referring to these clicks and predictions as \textit{Real-Ints}.
The results indicate that applying high-quality masks of \textit{exemplars} reduces users' efforts and improves segmentation performance.
For instance, both Simulated-Ints and Real-Ints achieved better results than Automatic-Ints because the IoU of a predicted mask obtained from automatic evaluation could fail to reach 90\%.
Surprisingly, the performance achieved through real interactions is barely superior to our simulated interactions.
These results demonstrate that the adopted Simulated Ints suffice to guide the \textit{exemplars} for MOIS.

\begin{table}
\caption{Ablation study of the influence of different \textit{exemplars} on predicting recall objects for multiple object interactive segmentation.}
\centering
\setlength\tabcolsep{3pt}
\begin{tabular}{p{2.05cm}cccc}
\toprule
\multirow{2}*{Exemplar} & \multicolumn{4}{c}{HIM2K}\\
& NoC85$\downarrow$ & NoC90$\downarrow$ & mIoU$\circledast$5+$\uparrow$ & mIoU$\circledast$10+$\uparrow$ \\
\hline
 Simulated-Ints & \textbf{1.16} & 1.38 & 88.73\% & 96.31\%\\
 Automatic-Ints & 1.18 & 1.42 & 88.47\% & 96.10\%\\
 Real-Ints & \textbf{1.16} & \textbf{1.36} & \textbf{88.75\%} & \textbf{96.34\%} \\
\bottomrule
\end{tabular}
\vspace{-2mm}
\label{tab: ablation study on exemplar}
\end{table}

\subsubsection{Backbone Analysis of EIM}
\label{sec: ab backbone}
We analyzed different backbones of the feature extraction in the proposed \textit{exemplar}-informed module (EIM).
As shown in Table~\ref{tab: ablation study on cim backbone}, we report the results of three backbones: (a) ResNet-18 \citep{he2016resnet}; (b) ResNet-50; (c) HRNet-18 \citep{wang2020hrnet}.
ResNet-50 achieved an almost negligible lead.
The results indicate that all three lightweight backbones can effectively extract the features from images and provide valid representations for the similarity calculation.
Due to the significantly increased computational complexity and inference time of using larger backbones such as ResNet-101 and HRNet-32, we opted against using them for the IS system.
Thus, we chose ResNet-18 as the backbone for the \textit{exemplars}-informed module.

\begin{table}
\caption{Ablation study of different backbones of the feature representation in the proposed EIM.}
\centering
\setlength\tabcolsep{3pt}
\begin{tabular}{p{1.75cm}cccc}
\toprule
\multirow{2}*{Backbone} & \multicolumn{4}{c}{COCO MOIS}\\
& NoC85$\downarrow$ & NoC90$\downarrow$ & mIoU$\circledast$5+$\uparrow$ & mIoU$\circledast$10+$\uparrow$ \\
\hline
ResNet-18 & 8.47 & \textbf{11.92} & 82.47\% & 85.89\%\\
ResNet-50 & \textbf{8.44} & 11.93 & \textbf{82.49\%} & \textbf{85.90\%} \\
HRNet-18 & 8.52 & 11.96 & 82.39\% & 85.83\%\\
\bottomrule
\end{tabular}
\vspace{-2mm}
\label{tab: ablation study on cim backbone}
\end{table}

\begin{figure*}[ht]
  \centering
  \begin{subfigure}[t]{0.18\linewidth}
    \includegraphics[width=1\linewidth]{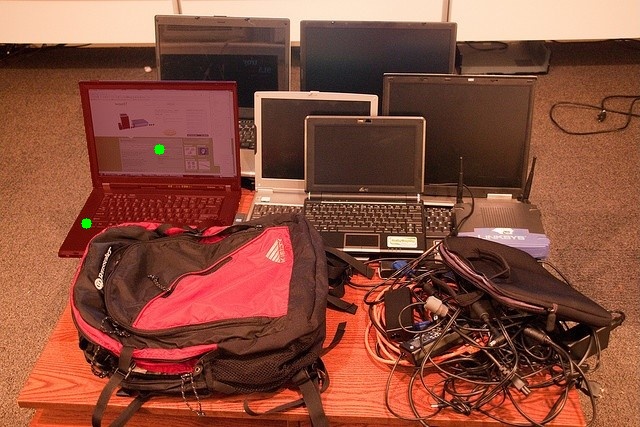}
    \subcaption*{Exemplar 95.1\%}
  \end{subfigure}
  \hspace{2mm}
  \begin{subfigure}[t]{0.18\linewidth}
    \includegraphics[width=1\linewidth]{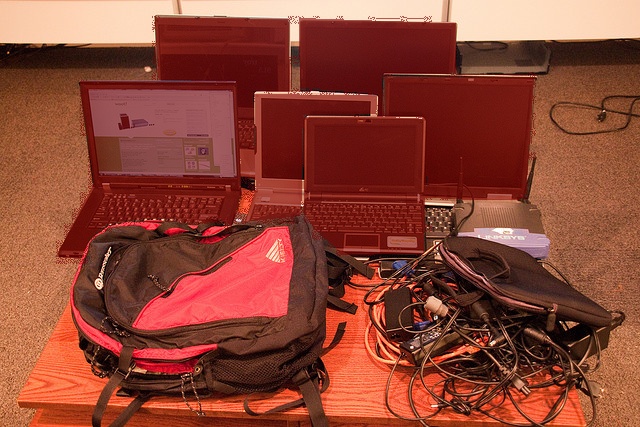}
    \subcaption*{GT}
  \end{subfigure}
  \hspace{2mm}
  \begin{subfigure}[t]{0.18\linewidth}
    \includegraphics[width=1\linewidth]{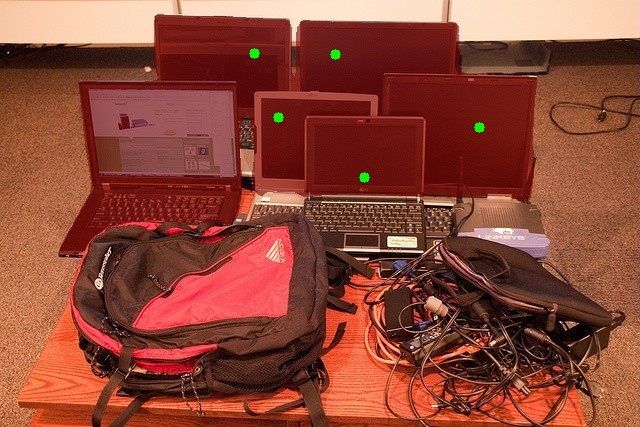}
    \subcaption*{5 clicks 84.3\%}
  \end{subfigure}
  \hspace{2mm}
  \begin{subfigure}[t]{0.18\linewidth}
    \includegraphics[width=1\linewidth]{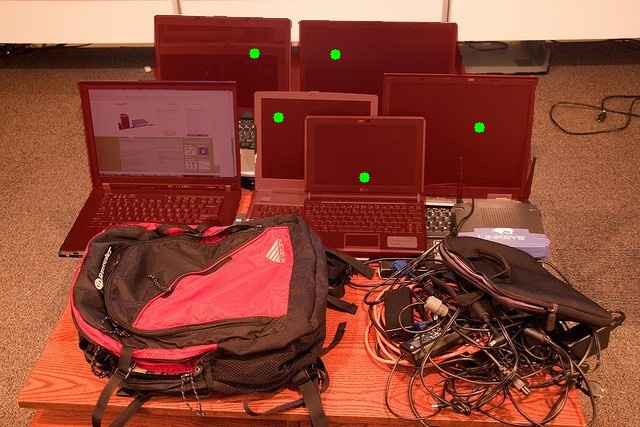}
    \subcaption*{5 clicks 92.8\%}
  \end{subfigure}
  \hspace{2mm}
  \begin{subfigure}[t]{0.18\linewidth}
    \includegraphics[width=1\linewidth]{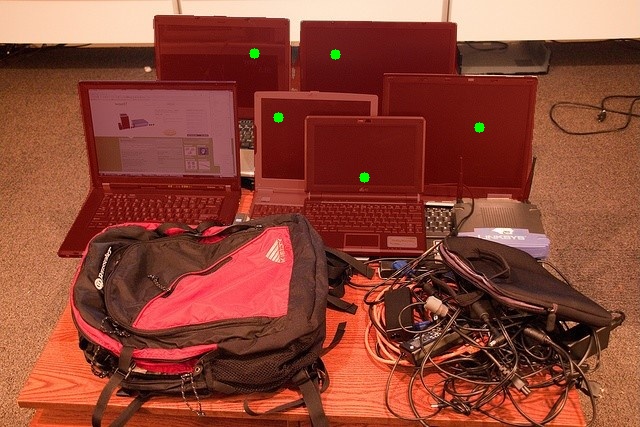}
    \subcaption*{5 clicks 93.7\%}
  \end{subfigure}

  \begin{subfigure}[t]{0.18\linewidth}
        \includegraphics[width=1\linewidth]{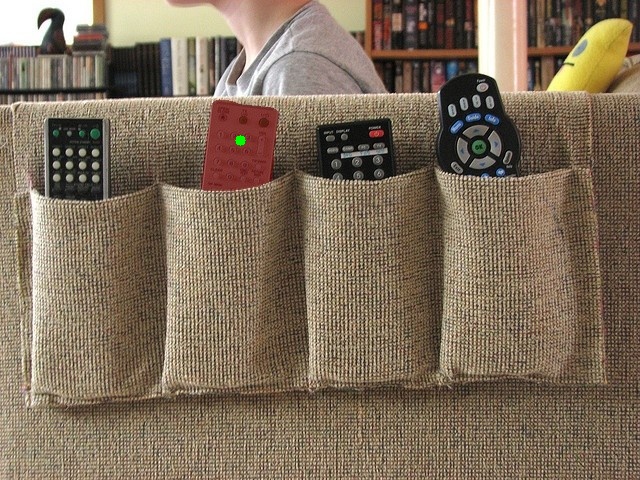}
        \subcaption*{Exemplar 91.0\%}
      \end{subfigure}
  \hspace{2mm}
  \begin{subfigure}[t]{0.18\linewidth}
    \includegraphics[width=1\linewidth]{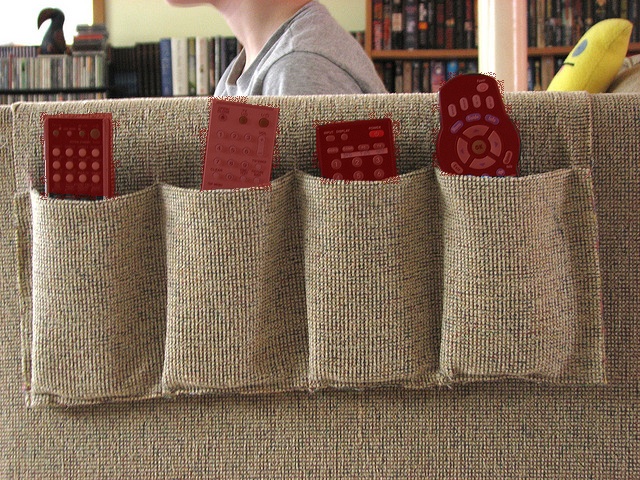}
    \subcaption*{GT}
  \end{subfigure}
  \hspace{2mm}
  \begin{subfigure}[t]{0.18\linewidth}
    \includegraphics[width=1\linewidth]{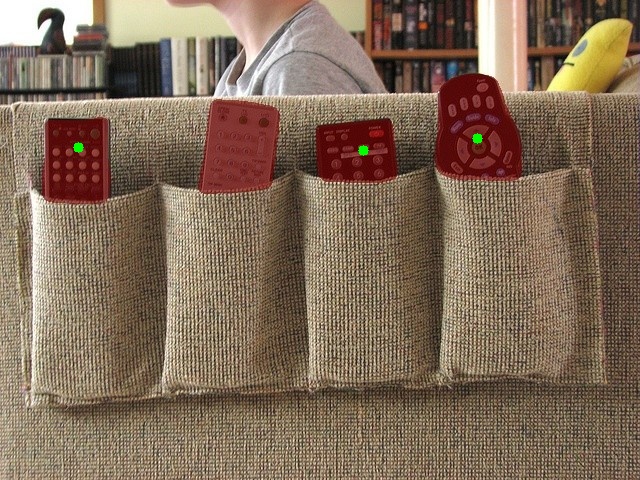}
    \subcaption*{3 clicks 85.9\%}
  \end{subfigure}
  \hspace{2mm}
  \begin{subfigure}[t]{0.18\linewidth}
    \includegraphics[width=1\linewidth]{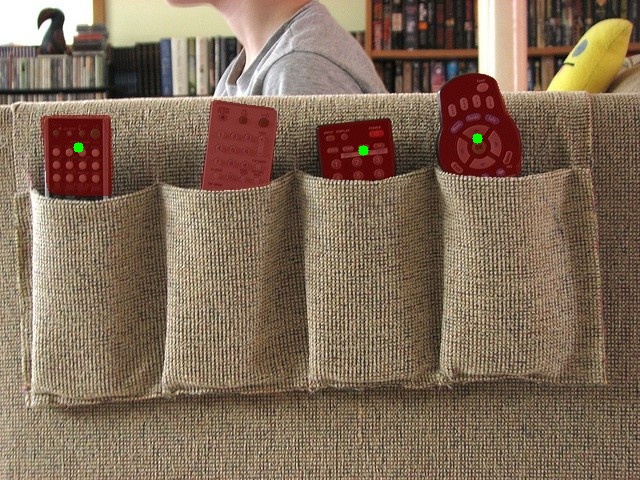}
    \subcaption*{3 clicks 90.1\%}
  \end{subfigure}
  \hspace{2mm}
  \begin{subfigure}[t]{0.18\linewidth}
    \includegraphics[width=1\linewidth]{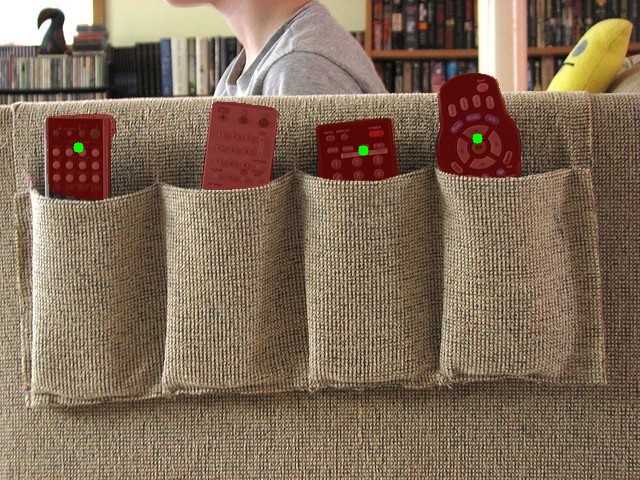}
    \subcaption*{3 clicks 90.4\%}
  \end{subfigure}

  \begin{subfigure}[t]{0.18\linewidth}
    \includegraphics[width=1\linewidth]{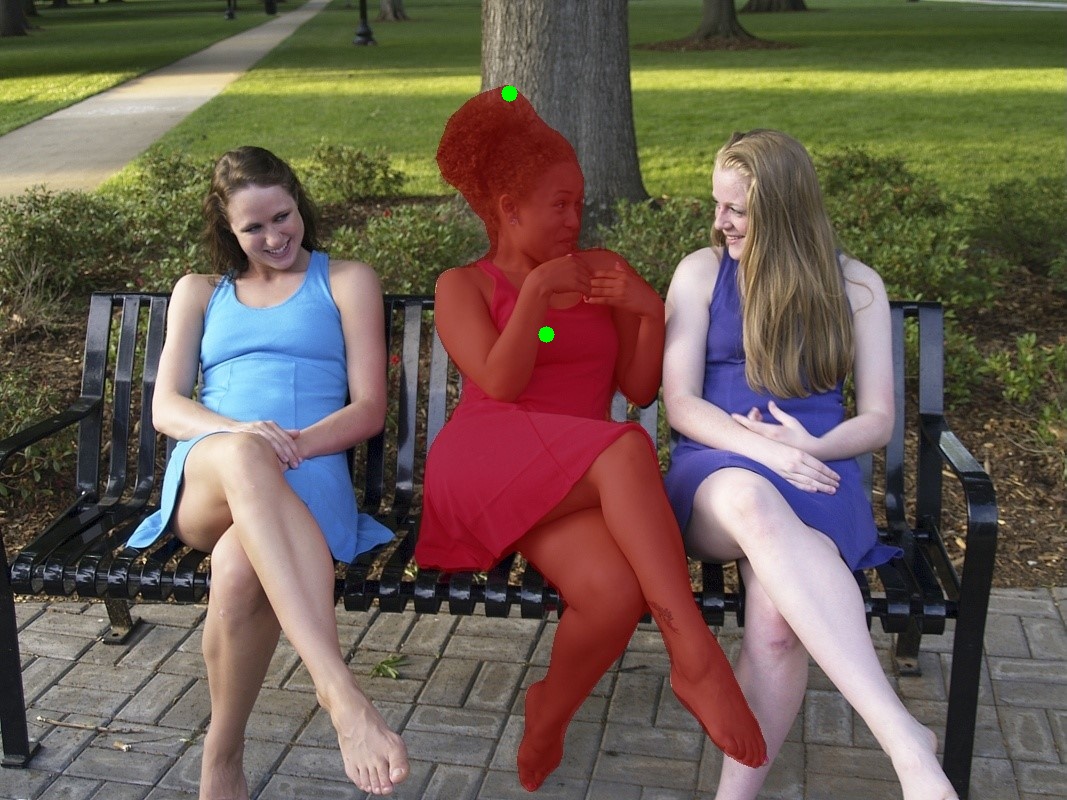}
    \subcaption*{Exemplar 93.4\%}
  \end{subfigure}
  \hspace{2mm}
  \begin{subfigure}[t]{0.18\linewidth}
    \includegraphics[width=1\linewidth]{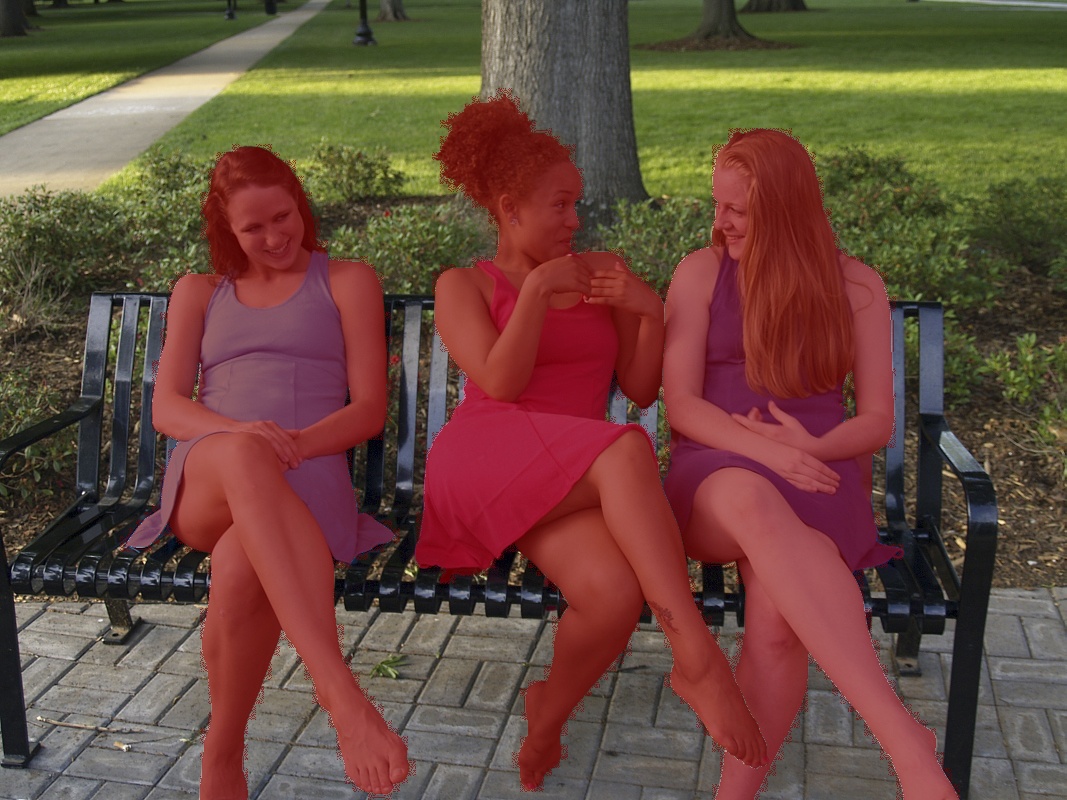}
    \subcaption*{GT}
  \end{subfigure}
  \hspace{2mm}
  \begin{subfigure}[t]{0.18\linewidth}
    \includegraphics[width=1\linewidth]{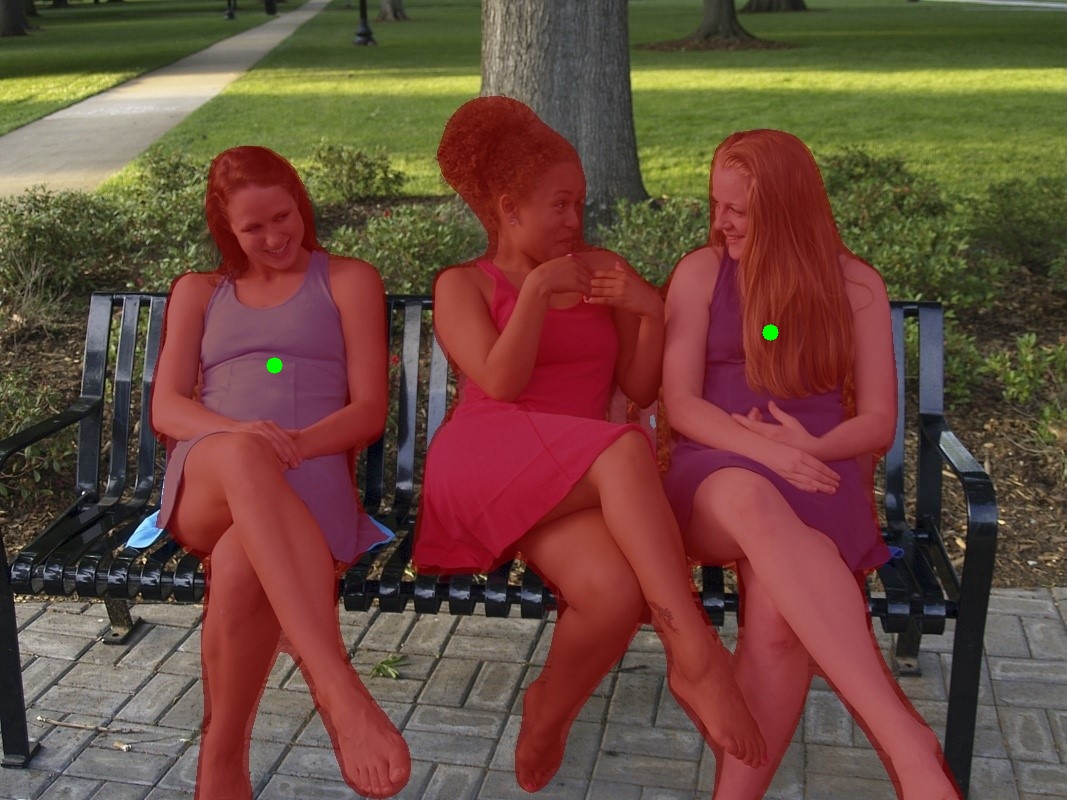}
    \subcaption*{2 clicks 91.8\%}
  \end{subfigure}
  \hspace{2mm}
  \begin{subfigure}[t]{0.18\linewidth}
    \includegraphics[width=1\linewidth]{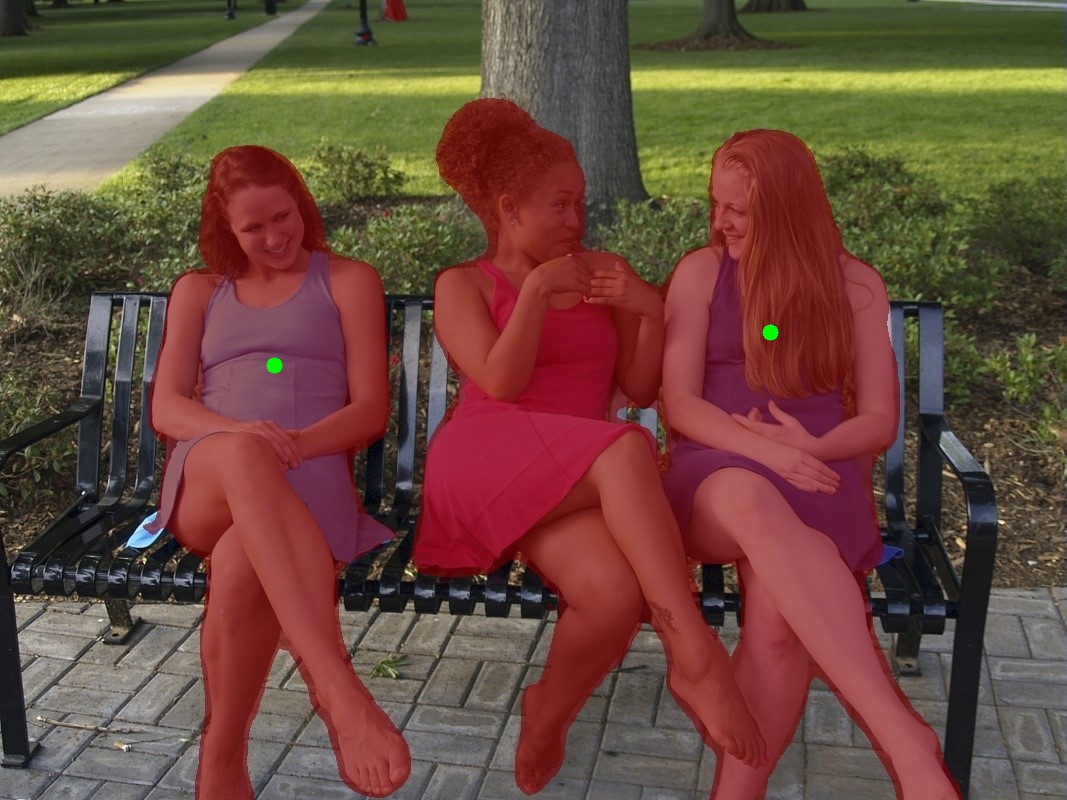}
    \subcaption*{2 clicks 92.1\%}
  \end{subfigure}
  \hspace{2mm}
  \begin{subfigure}[t]{0.18\linewidth}
    \includegraphics[width=1\linewidth]{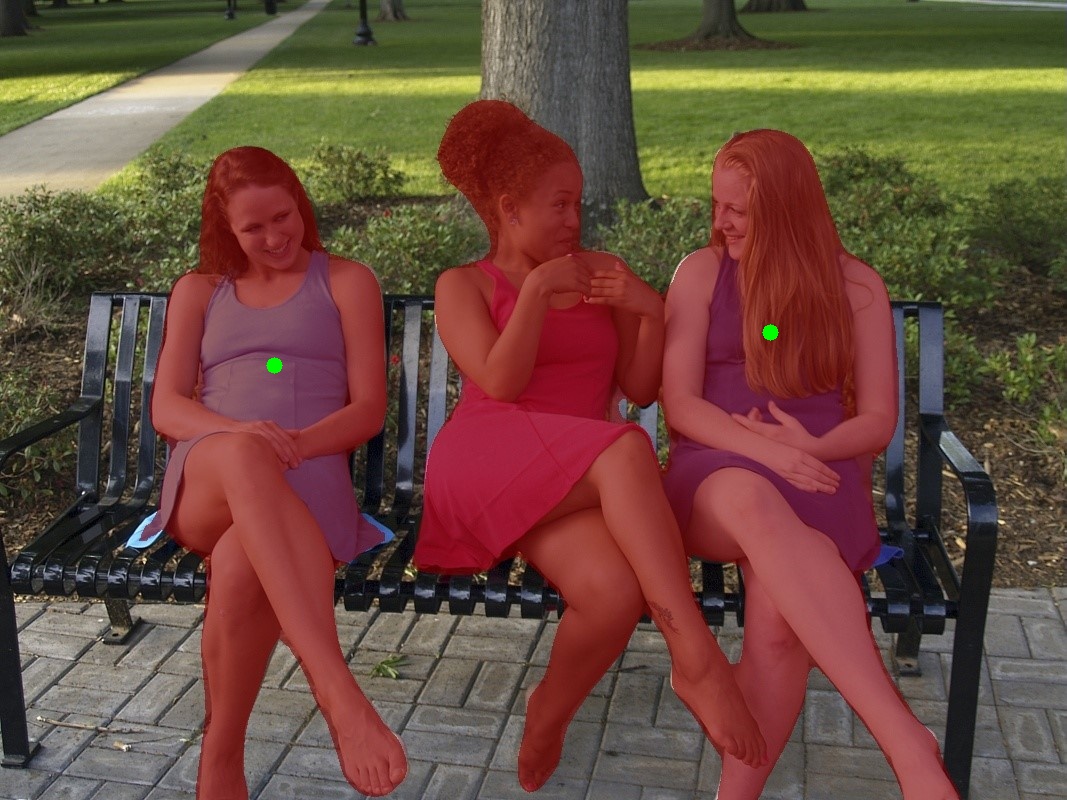}
    \subcaption*{2 clicks 94.1\%}
  \end{subfigure}

    \begin{subfigure}[t]{0.18\linewidth}
    \includegraphics[width=1\linewidth]{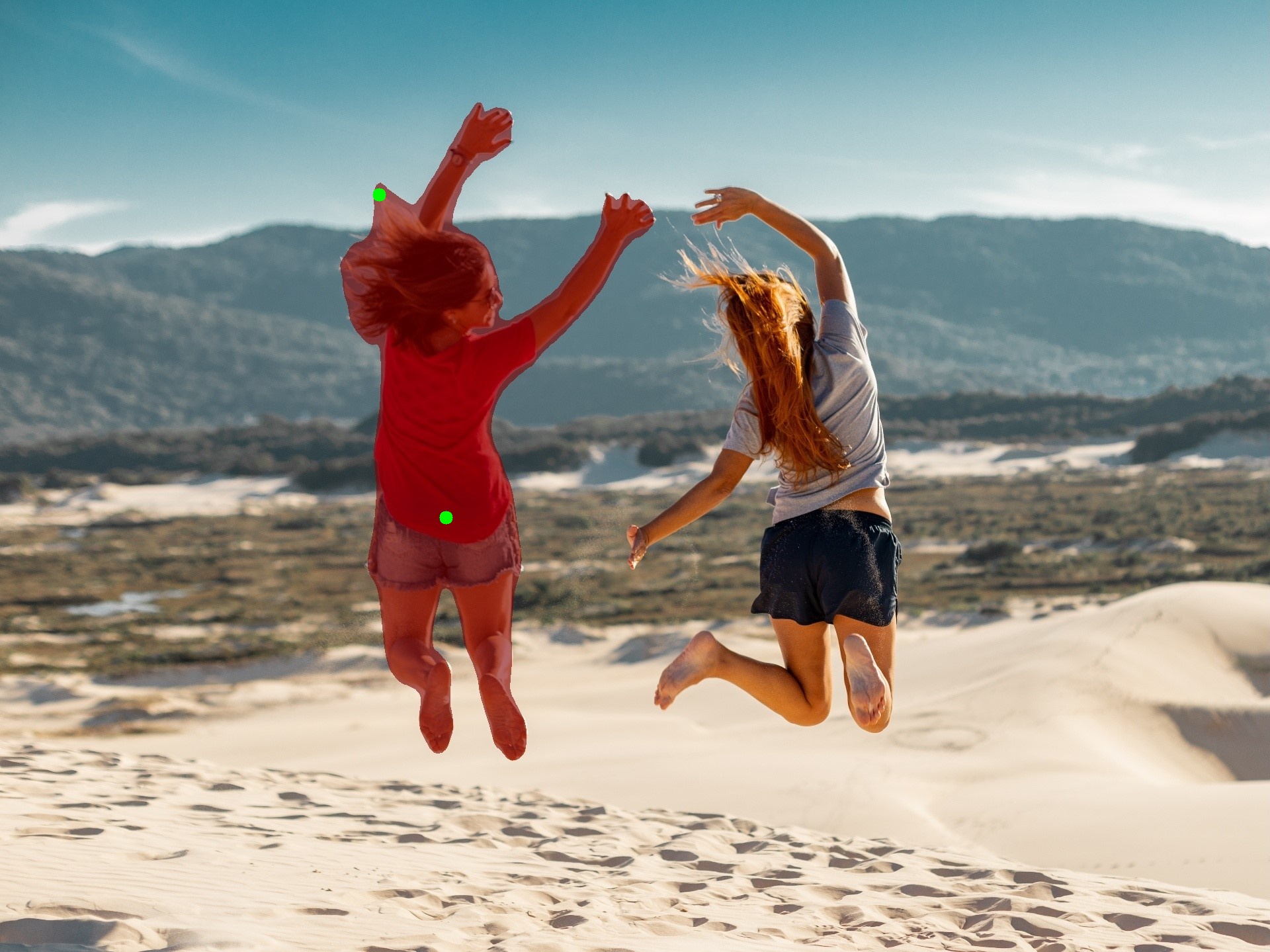}
    \subcaption*{Exemplar 91.6\%}
  \end{subfigure}
  \hspace{2mm}
  \begin{subfigure}[t]{0.18\linewidth}
    \includegraphics[width=1\linewidth]{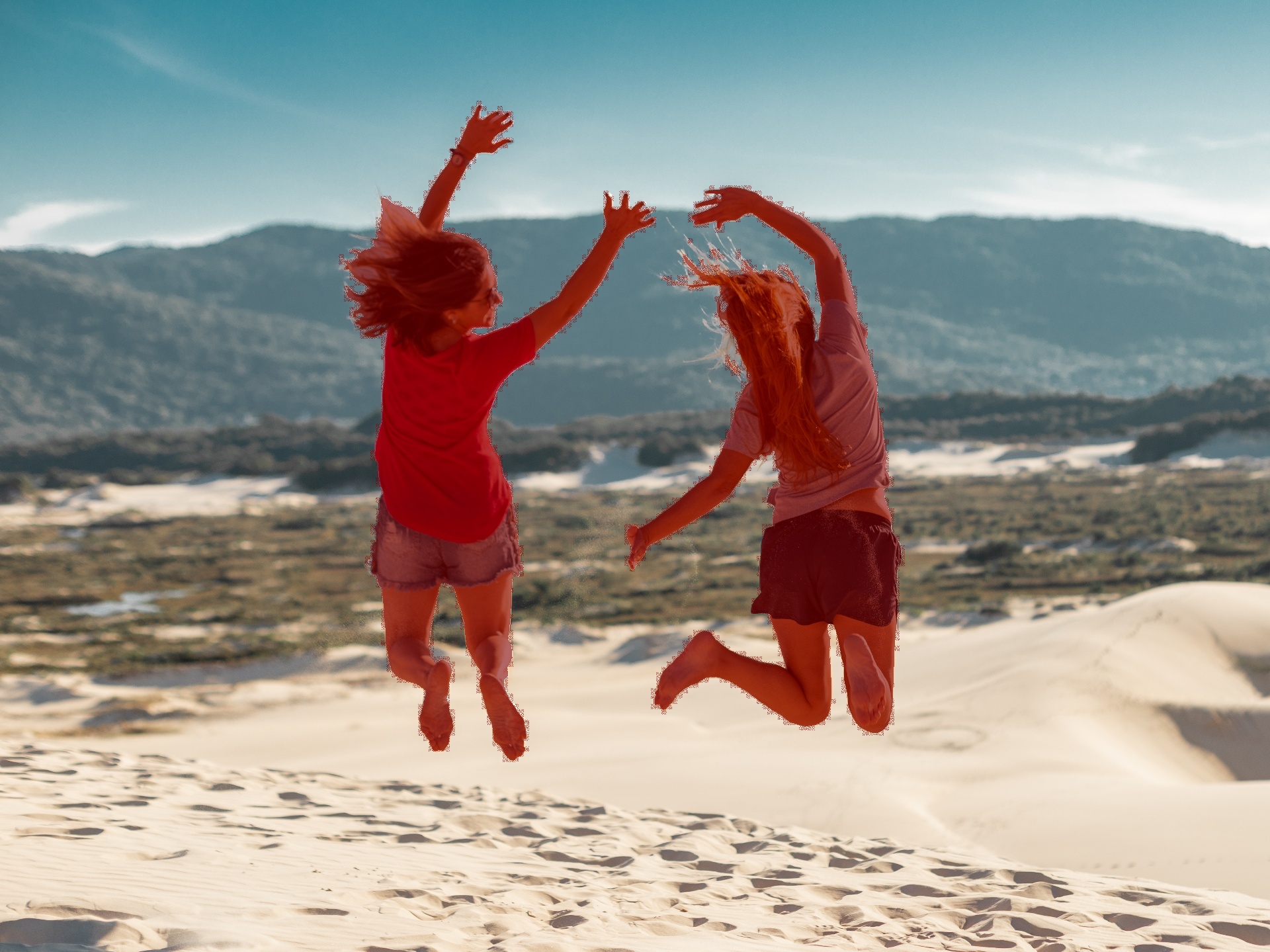}
    \subcaption*{GT}
  \end{subfigure}
  \hspace{2mm}
  \begin{subfigure}[t]{0.18\linewidth}
    \includegraphics[width=1\linewidth]{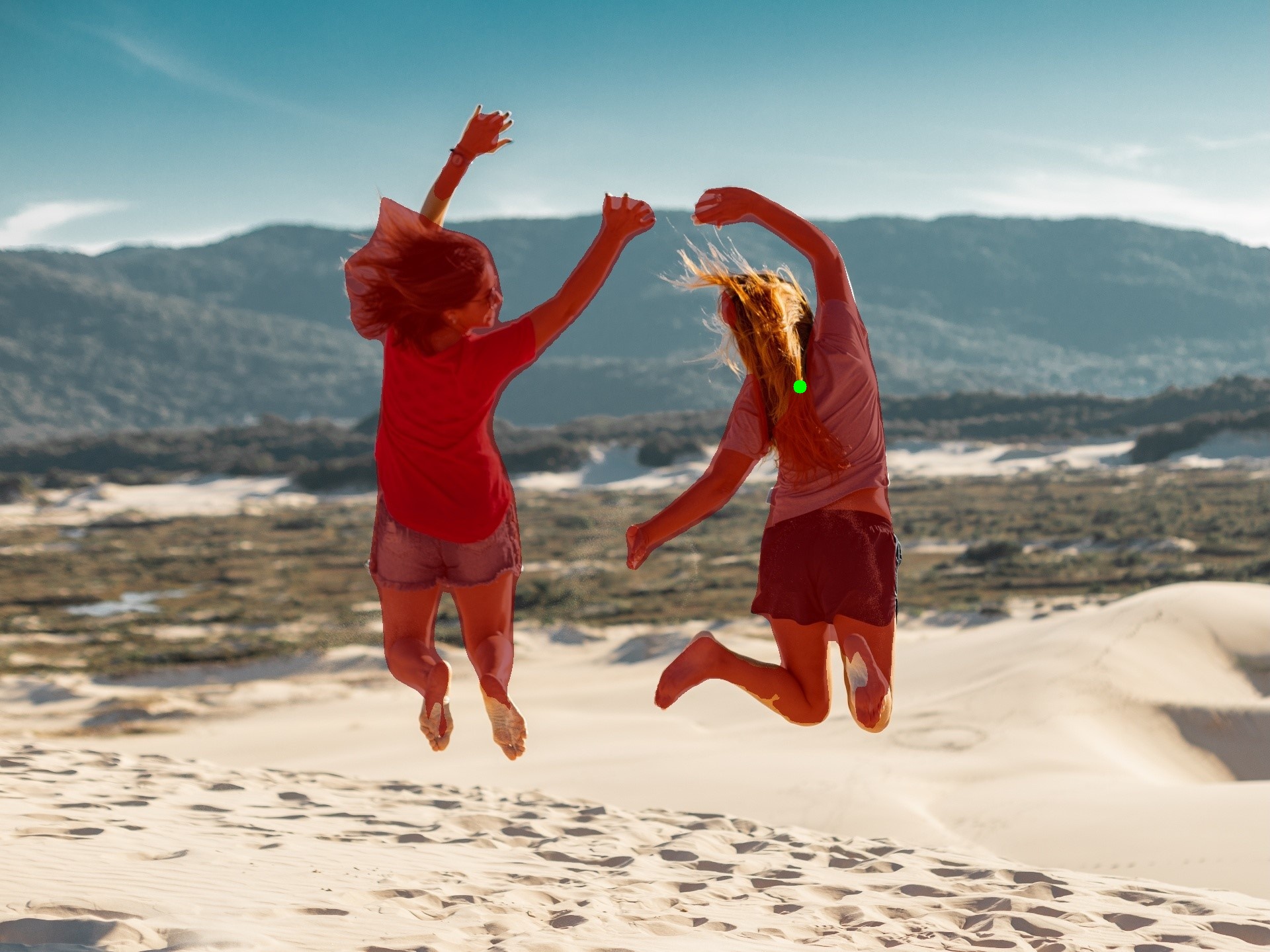}
    \subcaption*{1 click 81.6\%}
  \end{subfigure}
  \hspace{2mm}
  \begin{subfigure}[t]{0.18\linewidth}
    \includegraphics[width=1\linewidth]{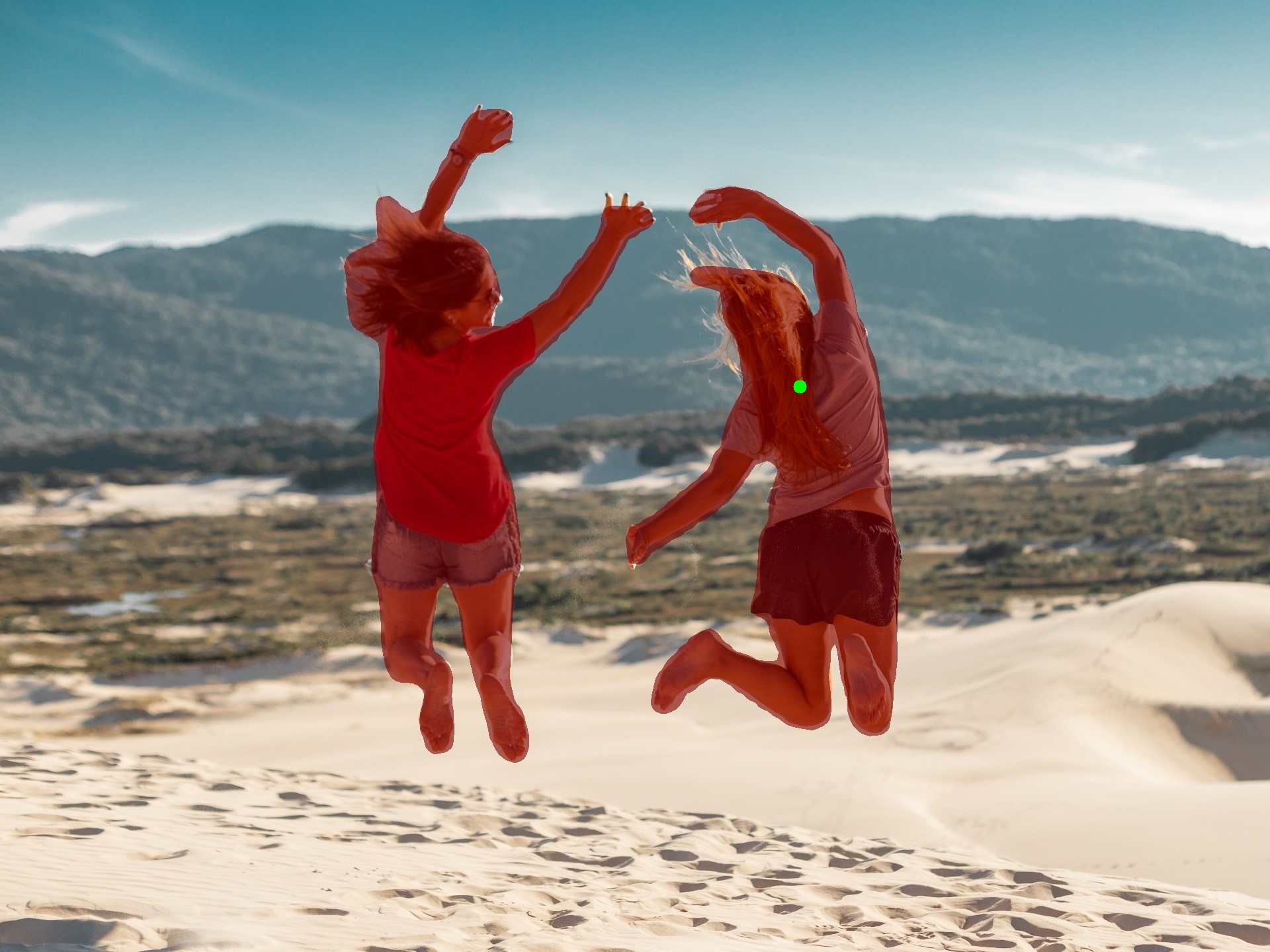}
    \subcaption*{1 click 90.0\%}
  \end{subfigure}
  \hspace{2mm}
  \begin{subfigure}[t]{0.18\linewidth}
    \includegraphics[width=1\linewidth]{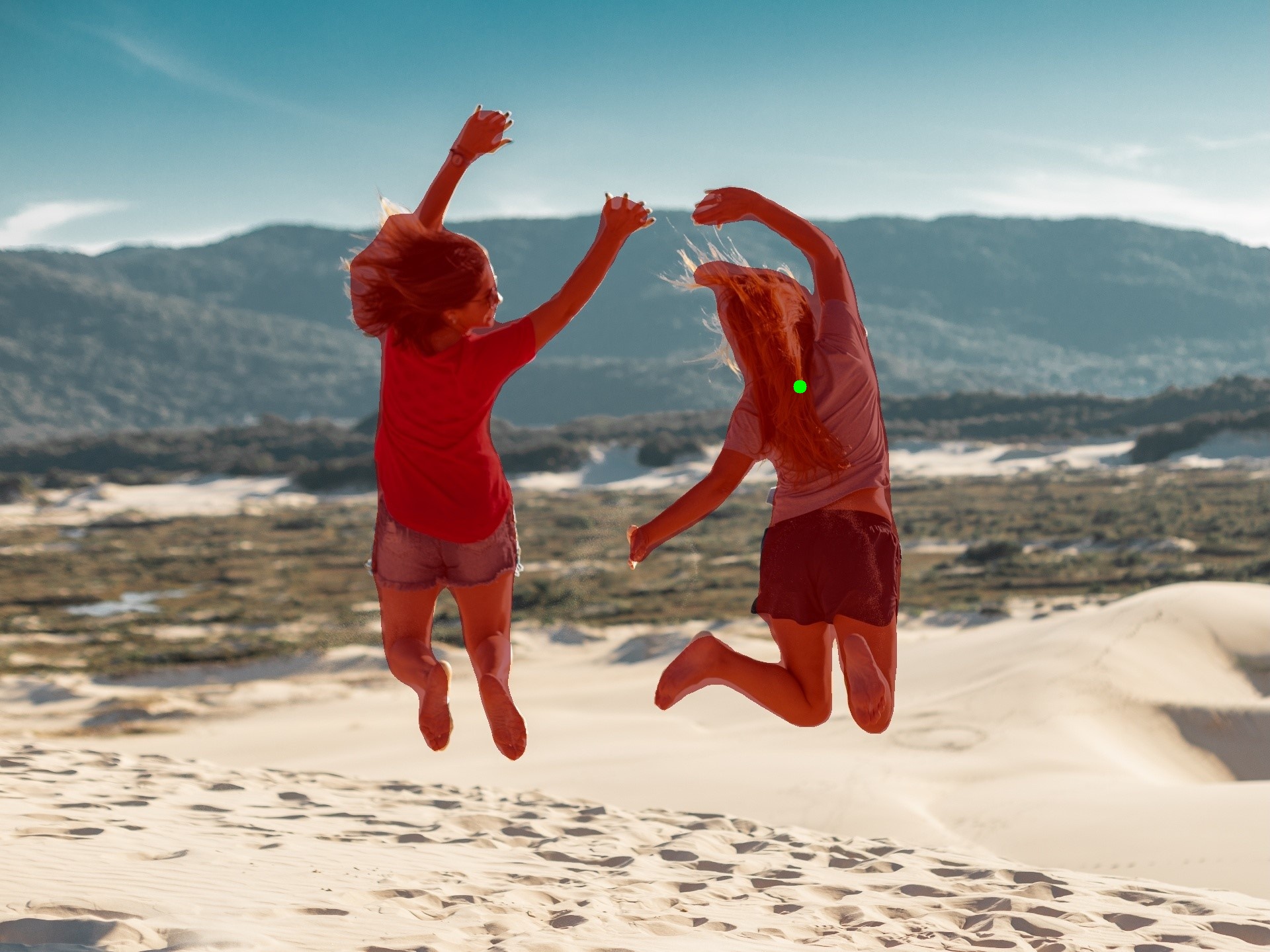}
    \subcaption*{1 click 90.9\%}
  \end{subfigure}

  \begin{subfigure}[t]{0.18\linewidth}
  \vspace{-2mm}
    \subcaption*{Exemplar with mask}
  \end{subfigure}
  \hspace{2mm}
  \begin{subfigure}[t]{0.18\linewidth}
    \vspace{-2mm}
    \subcaption*{GT with mask}
  \end{subfigure}
  \hspace{2mm}
  \begin{subfigure}[t]{0.18\linewidth}
  \vspace{-2mm}
    \subcaption*{RITM}
  \end{subfigure}
  \hspace{2mm}
  \begin{subfigure}[t]{0.18\linewidth}
  \vspace{-2mm}
    \subcaption*{AdaptiveClick}
  \end{subfigure}
  \hspace{2mm}
  \begin{subfigure}[t]{0.18\linewidth}
  \vspace{-2mm}
    \subcaption*{iCMFormer++ (Ours)}
  \end{subfigure}

   \vspace{-2mm}
  \caption{Qualitative results compared to RITM \citep{sofiiuk2022ritm}, FocalClick \citep{chen2022focalclick}, and AdaptiveClick \citep{lin2024adaptiveclick}, within one positive click for each object in the same category from the MOIS evaluation datasets.}
  \vspace{-2mm}
  \label{fig:comparison with multiple}
\end{figure*}

\subsubsection{Model Complexity Analysis}
\label{sec: ab complexity}
We performed the computation analysis concerning parameters, FLOPs, and inference speed.
In Table~\ref{tab:computation comparison}, we report the corresponding models to represent various methods under the same computing environment (NVIDIA A40 GPU and Intel Silver 4216 CPU).
However, some methods process input images with different sizes. For example, FocalClick \citep{chen2022focalclick} uses a smaller size of 256, while most other methods use sizes around 448.
Hence, we also conducted our experiments with an image size of 448 for comparison.
The numbers of parameters are collected from the original works \citep{sofiiuk2022ritm, lin2022focuscut, chen2022focalclick, liu2023simpleclick, lin2024adaptiveclick}.
The inference speed was evaluated by the average time per click for the previous methods as in SimpleClick.
Although both our proposed models require more parameters compared to the previous methods, their inference speeds, \eg, \SI{78}{ms} and \SI{92}{ms}, are on the same scale as the other models and meet real-time demands for IS.
Hence, we believe that iCMFormer and iCMFormer++ have a high potential for real-world applications.

\begin{table}
  \caption{Computation comparison with different models concerning parameters, FLOPs, and inference speed.}
  \centering
  \setlength\tabcolsep{3pt}
  \begin{tabular}{p{2.15cm}ccccc}
    \toprule
    Model & Backbone &Size & \# Params & \# FLOPs & SPC\\
    \midrule
    RITM & HRNet-32 & 400 & 30.95M & 82.84G & \SI{84}{ms}\\
    FocusCut & ResNet-101 & 384 & 59.35M & 102.02G & \SI{384}{ms}\\
    FocalClick & SegF-B0-S2 & 256 & 3.72M & 3.54G & \SI{42}{ms}\\
    SimpleClick & ViT-B & 448 & 96.46M & 169.78G & \SI{67}{ms}\\
    AdaptiveClick & ViT-B & 448 & 116.41M & 269.81G & \SI{84}{ms}\\
    iCMFormer(Ours) & ViT-B & 448 & 124.81M & 297.54G & \SI{78}{ms}\\
    iCMFormer++(Ours) & ViT-B & 448 & 178.73M &  496.32G & \SI{92}{ms}\\
    \bottomrule
  \end{tabular}
  \label{tab:computation comparison}
  \vspace{-4mm}
\end{table}

\subsection{Qualitative Results}
To qualitatively evaluate the performance of our IS models, we show some predicted results for all evaluation datasets in this section.

We provide some representative segmentation results of our iCMFormer on the four datasets for SOIS in Fig.~\ref{fig:segmentations single}. The first two rows display the examples from GrabCut and Berkeley, respectively.
They show that our model can obtain high-quality segmentation masks within two clicks.
The last three rows display the results from SBD and COCO MVal, showing that the segmentation results get better with progressively added clicks on the incorrect prediction regions.

We show some segmentation results on COCO MOIS and HIM2K for MOIS using our iCMFormer++ model in Fig.~\ref{fig:segmentations multiple}.
The challenge of predicting high-quality masks for multiple objects varies in different scenarios, especially with the increase in the number of target objects.
We present two common cases and one difficult case where achieving target IoUs requires more than the average number of clicks.
The results show that our iCMFormer++ can obtain good segmentation masks of remaining objects with the help of satisfactory \textit{exemplars}.

Furthermore, we compare the qualitative results generated by iCMFormer++ with that of other models, \ie, RITM \citep{sofiiuk2022ritm}, FocalClick \citep{chen2022focalclick}, SimpleClick \citep{liu2023simpleclick}, and AdaptiveClick \citep{lin2024adaptiveclick}.
Normally, the qualitative results are collected from subjective human clicks, which is different for every inference.
In other words, the qualitative comparison with other methods is unfair considering the judgment of the results and added clicks could be strongly user-dependent.
To mitigate this issue, we randomly picked several examples from the SOIS datasets and put the first positive click in the same place for the tested images, as shown in Fig.~\ref{fig:comparison with single}.
These examples both visually and quantitatively verify the superiority of our proposed iCMFormer model.
Similarly, we added only one positive click for each object in the MOIS task and show the compared segmentation results in Fig.~\ref{fig:comparison with multiple}. 
Our method, iCMFormer++, outperforms the other state-of-the-art models by achieving a higher mIoU.

\begin{figure}[t!]
  \centering
  \begin{subfigure}[t]{0.53\linewidth}
  \includegraphics[height=1.308cm]{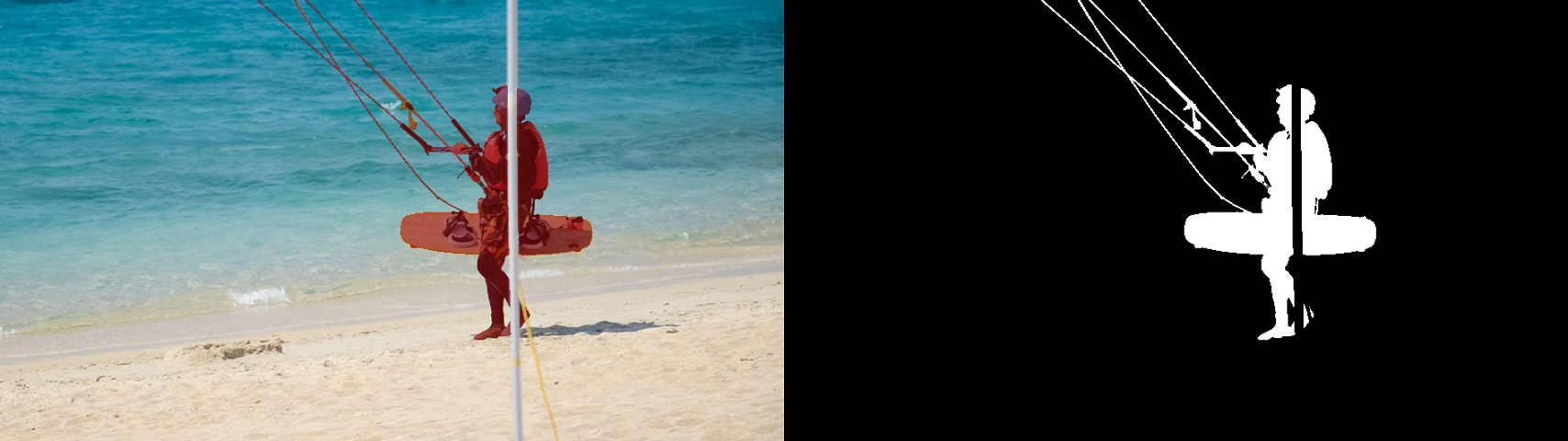}
    \subcaption*{Input \& GT}
  \end{subfigure}
  \hspace{1mm}
  \begin{subfigure}[t]{0.44\linewidth}
    \includegraphics[height=1.308cm]{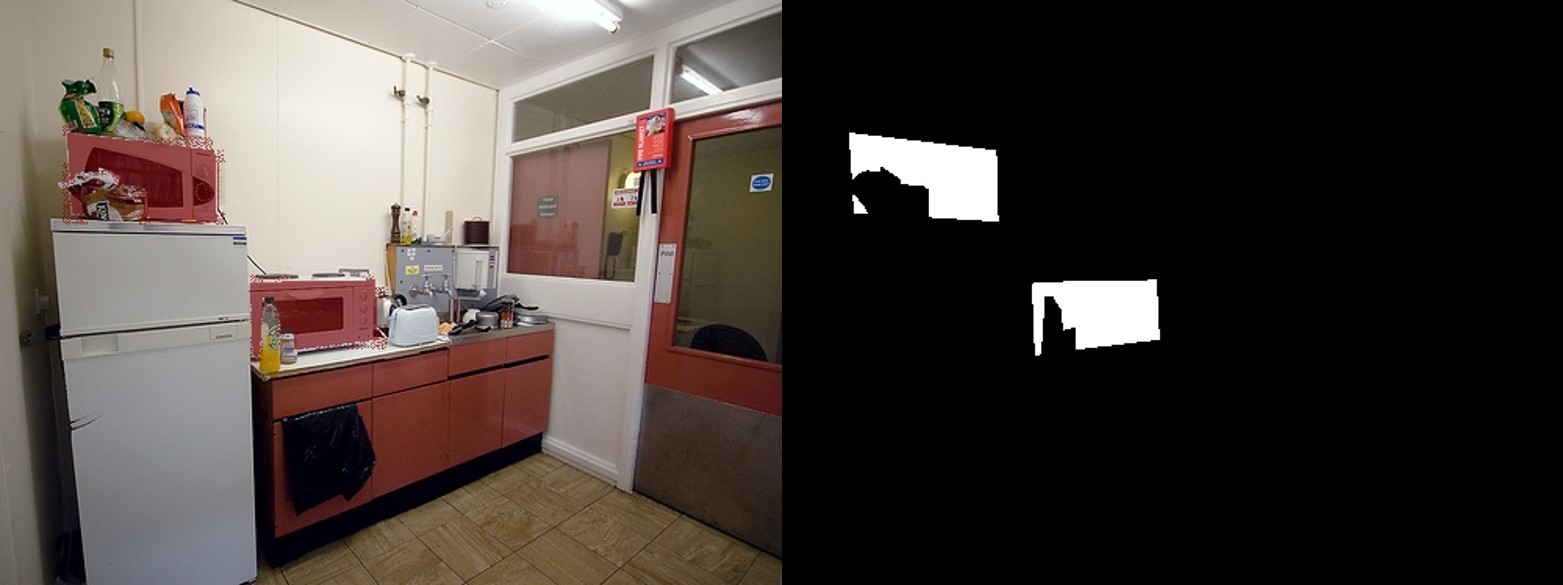}
    \subcaption*{Input \& GT}
  \end{subfigure}

  \begin{subfigure}[t]{0.53\linewidth}
        \includegraphics[height=1.308cm]{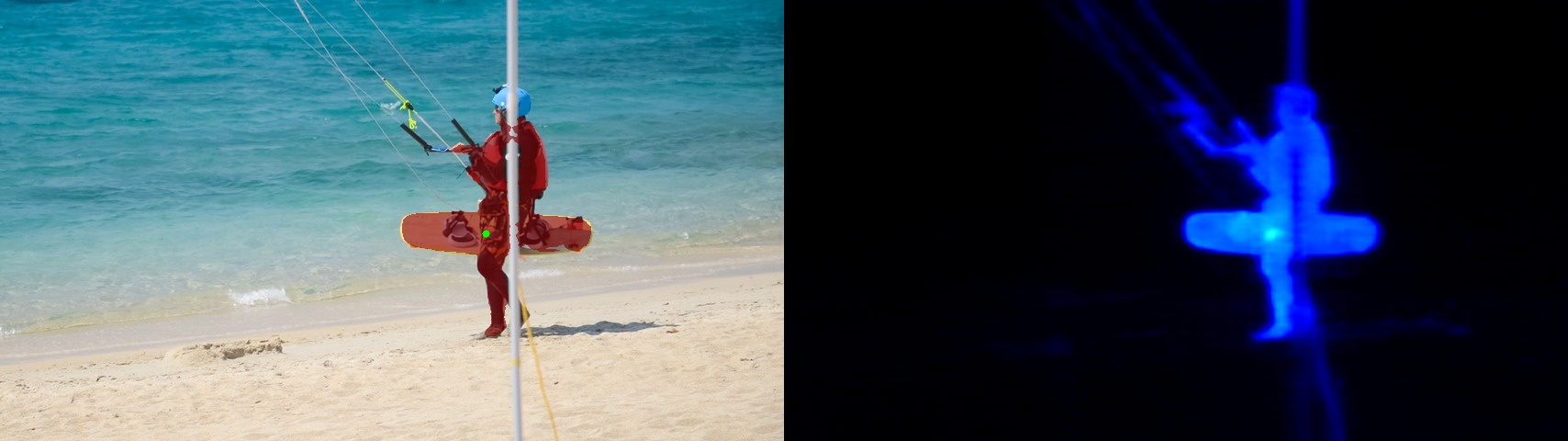}
        \subcaption*{1 click 70.6\% \& PM}
      \end{subfigure}
  \hspace{1mm}
  \begin{subfigure}[t]{0.44\linewidth}
    \includegraphics[height=1.308cm]{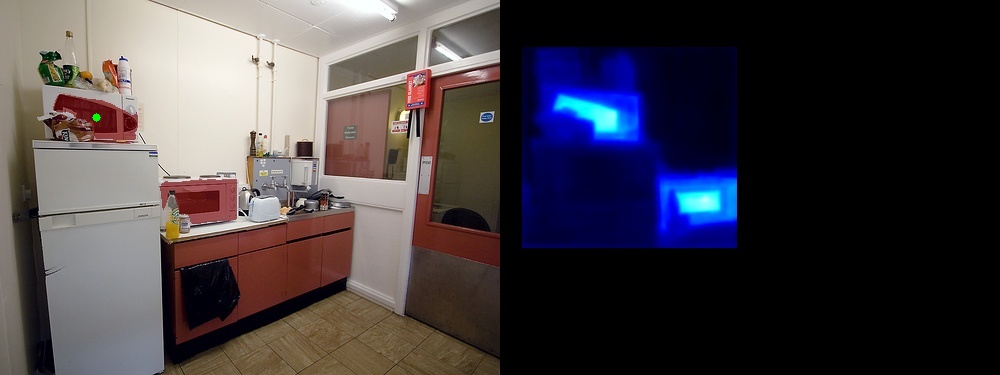}
    \subcaption*{1 click 66.1\% \& PM}
  \end{subfigure}

  \begin{subfigure}[t]{0.53\linewidth}
    \includegraphics[height=1.308cm]{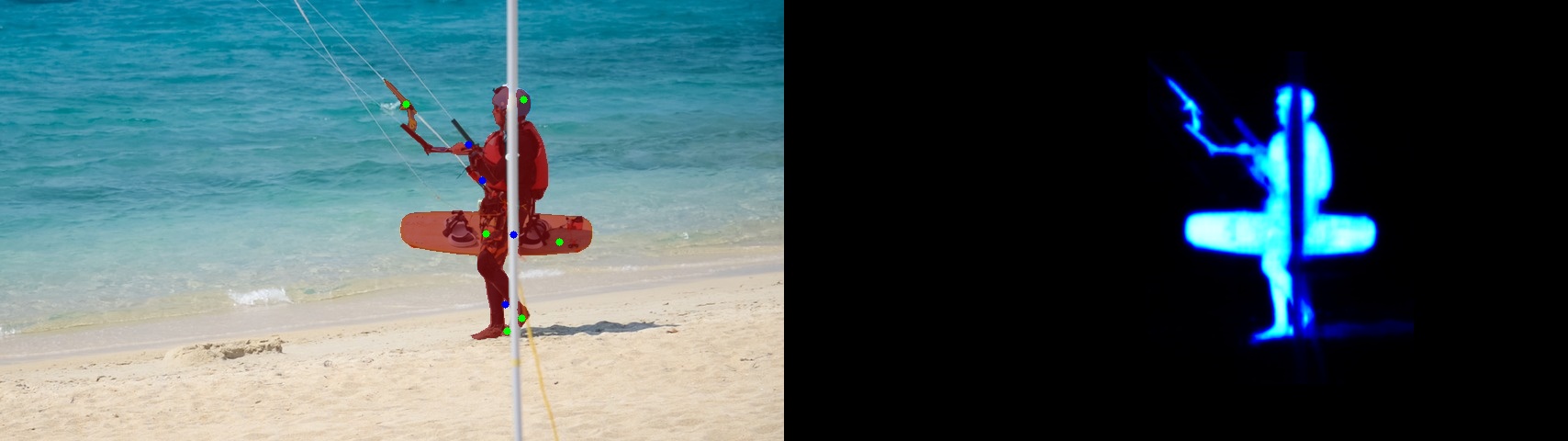}
    \subcaption*{10 clicks 82.2\% \& PM}
  \end{subfigure}
  \hspace{1mm}
  \begin{subfigure}[t]{0.44\linewidth}
    \includegraphics[height=1.308cm]{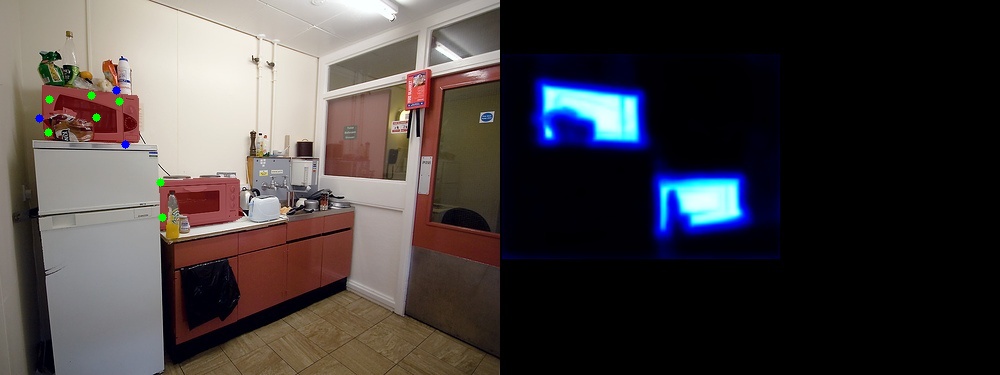}
    \subcaption*{10 clicks 83.5\% \& PM}
  \end{subfigure}

    \begin{subfigure}[t]{0.53\linewidth}
    \includegraphics[height=1.308cm]{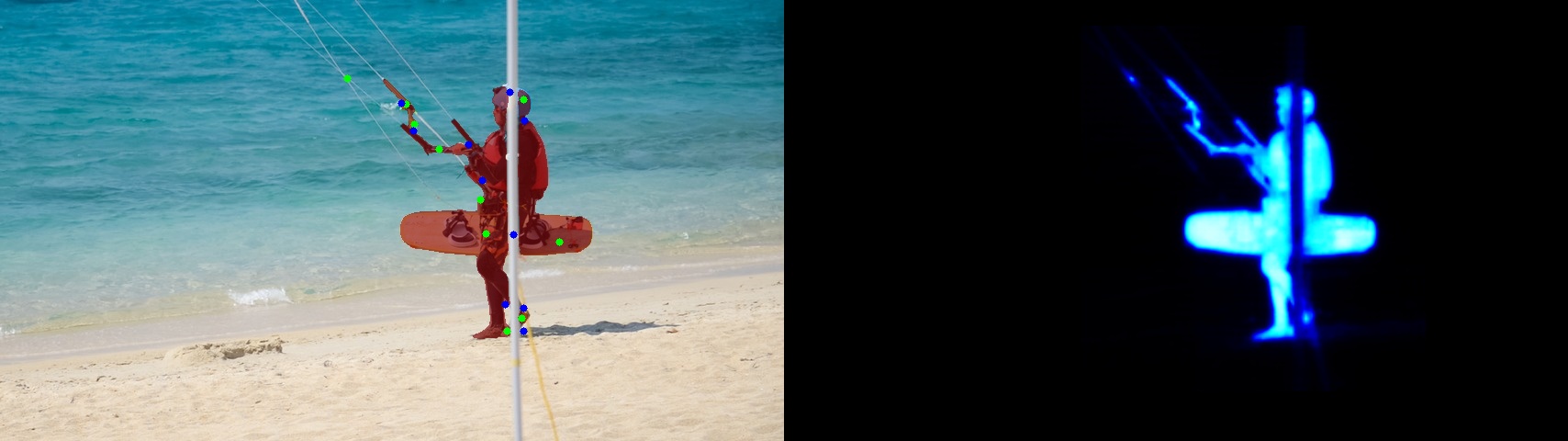}
    \subcaption*{20 clicks 83.1\% \& PM}
  \end{subfigure}
  \hspace{1mm}
  \begin{subfigure}[t]{0.44\linewidth}
    \includegraphics[height=1.308cm]{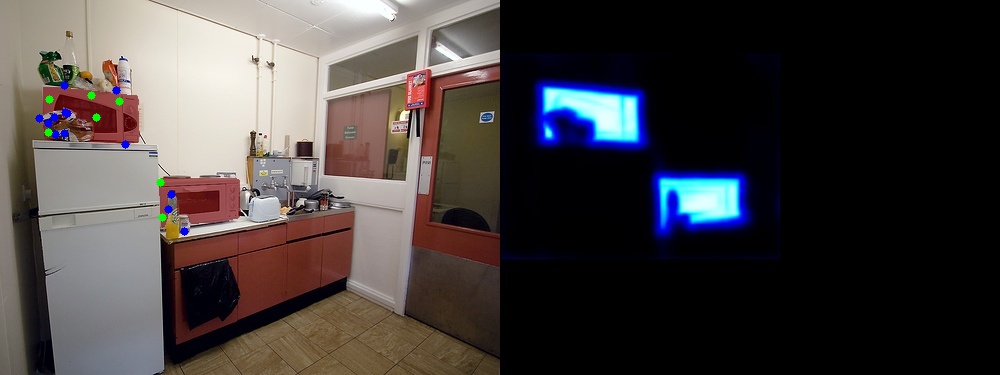}
    \subcaption*{20 clicks 84.8\% \& PM}
  \end{subfigure}
  
   \vspace{-2mm}
  \caption{Two illustrations of failure cases. The PM next to an image refers to the probability map of a prediction.}
  \vspace{-2mm}
  \label{fig:failure_cases}
\end{figure}

\section{Discussions}
\label{sec: limitation}
We introduce a new framework iCMFormer++ based on our previous iCMFormer model for MOIS.
There are two limitations to the proposed methods.
First, the predicted segmentation masks might be unsatisfactory when dealing with cluttered targets in complex scenarios.
We show some failure cases in Fig.~\ref{fig:failure_cases}.
For instance, the presence of the bottles and bags in front of the microwaves limits the improvement in segmentation performance even with additional clicks.
Fortunately, local refinements \citep{lin2022focuscut, chen2022focalclick} coupled with post-processing optimizations will enhance the segmentation accuracy.
Second, our method is not totally labor-free as the prediction of multiple objects requires additional clicks to improve the results.
The recent advancements in prompt learning \citep{zhu2023segprompt, kirillov2023sam} might provide a solution based on large-scale pretraining.
We defer these improvements to future work.

\section{Conclusions}
\label{sec: conclusion}
To solve the problem of interaction redundancy for multiple objects within the same category, this paper presents iCMFormer++, a novel approach for interactive image segmentation.
Specifically, in this work, we introduce a new concept, termed \textit{exemplar}, with a satisfactory mask to leverage the category information for segmenting the remaining objects. 
iCMFormer++ adopts iCMFormer, our model designed for single object interactive segmentation, to predict the mask of a randomly selected object as the \textit{exemplar} for each image.
The \textit{exemplar} guides the prediction of remaining objects sharing the same category with the \textit{exemplar object}.
To learn the category information passed by the \textit{exemplar}, we design an \textit{exemplar} informed module by comparing the similarity between the \textit{exemplar object} and the image regions. A simple channel fusion module is proposed to selectively and effectively fuse the attended features from different modules.
Comprehensive experimental evaluations conducted on various benchmarks demonstrate the superiority of the proposed approach.
Our models support interactive image segmentation from a single object to multiple objects in the same category, validating the benefit of using the category information to speed up the segmentation of remaining objects in the same category and highlighting its potential as an annotation tool in practical scenarios.

\section*{Conflict of Interest}
The authors declare that they have no conflict of interest.

\bibliographystyle{spbasic}
\bibliography{main}

\end{document}